\documentclass[acmsmall,screen,authorversion,nonacm]{acmart}
\AtBeginDocument{%
  }

\def\eg{e.g.}
\def\ie{i.e.}

\def\secref#1{Sec.~\ref{#1}}
\def\figref#1{Fig.~\ref{#1}}
\def\tabref#1{Tab.~\ref{#1}}
\usepackage{enumitem}
\usepackage[normalem]{ulem}
\useunder{\uline}{\ul}{}
\usepackage{longtable}
\usepackage{tabulary}
\usepackage{makecell}
\usepackage{pdflscape}
\usepackage{caption}
\usepackage{afterpage}
\usepackage{rotating}

\usepackage[utf8]{inputenc}
\usepackage{xcolor}
\usepackage[nolist,printonlyused]{acronym}
\usepackage{pgfgantt}
\usepackage{booktabs}
\usepackage{tabularx}
\usepackage{todonotes}
\usepackage{multirow}
\usepackage{tabularray}
\usepackage{wrapfig}
\usepackage{tikz}
\usepackage{amsfonts}
\usetikzlibrary{matrix,fit,calc}
\usepackage{array}
\usepackage{pdflscape}
\usepackage[nolist,printonlyused]{acronym}

\usepackage{hyperref}
\usepackage{pifont}
\newcommand{\cmark}{\ding{51}}%

\begin{document}

\title[Survey on Multi-label Hate Speech Classification in English]{A Survey of Machine Learning Models and Datasets for the Multi-label Classification of Textual Hate Speech in English}

\author{Julian Bäumler}
\orcid{0000-0002-4535-8036}

\author{Louis Blöcher}
\orcid{0009-0001-4284-3941}

\author{Lars-Joel Frey}
\orcid{0009-0001-2039-1315}

\author{Xian Chen}
\orcid{0009-0008-9603-2715}

\author{Markus Bayer}
\orcid{0000-0002-2040-5609}

\author{\allowbreak Christian Reuter}
\orcid{0000-0003-1920-038X}

\affiliation{%
  \institution{Technical University of Darmstadt, Science and Technology for Peace and Security (PEASEC)}
  \streetaddress{Pankratiusstraße 2}
  \city{Darmstadt}
  \country{Germany}
  \postcode{64289}
}

\email{ { baeumler,bayer,reuter } @peasec.tu-darmstadt.de}

\email{ { loius.bloecher,lars-joel.frey,xian.chen } @stud.tu-darmstadt.de}

\renewcommand{\shortauthors}{Bäumler et al.}

\begin{abstract}
The dissemination of online hate speech can have serious negative consequences for individuals, online communities, and entire societies. This and the large volume of hateful online content prompted both practitioners’, \ie, in content moderation or law enforcement, and researchers’ interest in machine learning models to automatically classify instances of hate speech. Whereas most scientific works address hate speech classification as a binary task, practice often requires a differentiation into sub-types, \eg, according to target, severity, or legality, which may overlap for individual content. Hence, researchers created datasets and machine learning models that approach hate speech classification in textual data as a multi-label problem. This work presents the first systematic and comprehensive survey of scientific literature on this emerging research landscape in English (N=46). We contribute with a concise overview of 28 datasets suited for training multi-label classification models that reveals significant heterogeneity regarding label-set, size, meta-concept, annotation process, and inter-annotator agreement. Our analysis of 24 publications proposing suitable classification models further establishes inconsistency in evaluation and a preference for architectures based on Bidirectional Encoder Representation from Transformers (BERT) and Recurrent Neural Networks (RNNs). We identify imbalanced training data, reliance on crowdsourcing platforms, small and sparse datasets, and missing methodological alignment as critical open issues and formulate ten recommendations for research. While transparency, data source diversity, conceptual rigor, label set consolidation, inter-annotator agreement, and model validation on existing datasets warrant more attention, multi-label tasks related to criminal relevance and the utility of large foundation models represent promising avenues for future work.
\end{abstract}



\begin{acronym}
\acro{ai}[AI]{artificial intelligence}
\acro{xai}[XAI]{explainable artificial intelligence}
\acro{iaa}[IAA]{inter-annotator agreement}
\acro{nlp}[NLP]{natural language processing}
\acro{ml}[ML]{machine learning}
\acro{llms}[LLMs]{large language models}
\acro{ngos}[NGOs]{non-governmental organizations}
\acro{api}[API]{Application programming interface}
\acro{br}[BR]{binary relevance}
\acro{lp}[LP]{label power-set}
\acro{dml}[DML]{direct multi-label classification}
\acro{mt}[MT]{Multi-Task}
\acro{tccc}[TCCC]{Toxic Comment Classification Challenge}
\acro{k}[k]{Thousand}
\acro{m}[m]{Million}

\acro{bert}[BERT]{Bidirectional Encoder Representation from Transformers}
\acro{bilstm}[Bi-LSTM]{Bidirectional Long Short Term Memory}
\acro{bigrus}[Bi-GRU]{Bidirectional Gated Recurrent Units}
\acro{cnn}[CNN]{Convolutional Neural Network}
\acro{capsnet}[CapsNet]{Capsule Neural Network}
\acro{lstm}[LSTM]{Long Short Term Memory}
\acro{lr}[LR]{Logistic Regression}
\acro{lgbm}[LGBM]{Light Gradient-Boosting Machine}
\acro{cart}[CART]{Classification and Regression Trees}
\acro{xlm}[XLM]{Cross-lingual Language Model}
\acro{nn}[NN]{neural network}
\acro{knn}[kNN]{k-nearest neighbors}
\acro{rnn}[RNN]{Recurrent Neural Network}
\acro{roberta}[roBERTa]{Robustly Optimized BERT Pretraining Approach}

\acro{pabak}[PABAK]{Prevalance and Bias Adjusted Kappa}
\acro{pabakos}[PABAK-OS]{Prevalance and Bias Adjusted Kappa-Ordinal Scale}
\acro{sa}[SA]{Subset Accuracy}
\acro{hl}[HL]{Hamming Loss}
\acro{er}[ER]{Error Rate}
\acro{p}[P]{Precision}
\acro{r}[R]{Recall}
\acro{f1}[F1]{F1 Score}
\acro{acc}[Acc]{Accuracy}
\acro{roc-auc}[ROC-AUC]{Area Under the Receiver Operating Characteristic Curve}
\acro{tp}[TP]{True Positives}
\acro{tn}[TN]{True Negatives}
\acro{fp}[FP]{False Positives}
\acro{fn}[FN]{False Negatives}
\acro{mcc}[MCC]{Matthews Correlation Coefficient}
\acro{ap}[AP]{Average Precision}
\acro{ir}[IR]{Irrelevance Rate}
\acro{auc}[AUC]{Area Under the Curve}
\acro{bpsn-auc}[BPSN-AUC]{Background Positive Subgroup Negative Area Under the Curve}
\acro{gmb-auc}[GMB-AUC]{Generalized Mean of Bias Area Under the Curve}

\acro{en}[EN]{English}
\acro{es}[ES]{Spanish}
\acro{it}[IT]{Italian}
\acro{de}[DE]{German}
\acro{hi}[HI]{Hindi}
\acro{fr}[FR]{French}
\acro{ar}[AR]{Arabic}
\acro{bn}[BN]{Bengali}

\acro{im}[IM]{Implicit}
\acro{ex}[EX]{Explicit}
\acro{ag}[AG]{Aggressive}
\acro{tg}[TG]{Targeted}
\acro{hs}[HS]{Hate speech}
\acro{det}[det]{Detection}
\acro{class}[class]{Classification}
\acro{hd}[HD]{Assault on human dignity}
\acro{cv}[CV]{Call for violence}
\acro{nh}[NH]{Non hateful}
\acro{vo}[VO]{Vulgarity/offensive}
\acro{rae}[RAE]{Race or ethnicity}
\acro{nat}[NAT]{Nationalism/regionalism}
\acro{gen}[GEN]{Gender}
\acro{rel}[REL]{Religion/spiritual belief}
\acro{sxo}[SXO]{Sexual orientation}
\acro{mph}[MPH]{Mental/physical health status}
\acro{pol}[POL]{Political identification}
\acro{idl}[IDL]{Ideology}
\end{acronym}


\maketitle

\section{Introduction}
\label{sec:review:intro}
Over the past decades, social media has become integral to many people's everyday social lives. However, harmful content, such as hate speech, is also disseminated on these platforms. Hate speech can have serious negative consequences. For victims, it can induce emotional distress and psychological trauma \cite{keipi_online_2016,leets_experiencing_2002}. Both among victims and bystanders, it can further encourage self-censorship and a reduction in online activities \cite{lenhart_online_2016}. In response, many social media platforms prohibit hate speech and sanction its dissemination \cite{schaffner_community_2024}. The substantial increase in users and engagement has prompted many platforms to (partially) automate content moderation tasks, including the identification of hateful content \cite{singhal_sok_2023,scheuerman_framework_2021}. Moreover, the dissemination of extreme forms of hate speech constitutes a criminal offense in some countries, and thus, law enforcement agencies are tasked with its prosecution \cite{chetty_hate_2018}. Some of them cooperate with non-profit hate speech reporting centers, which support victims in content deletion, finding psycho-social support, and the initiation of criminal proceedings \cite{baumler_harnessing_2025}. As these organizations are also confronted with large incident volumes, potential is seen in the partial automation of classification \cite{baumler_cyber_2025,kaufhold_cylence_2023}. Against this background, hate speech classification received significant attention in the scientific fields of \ac{ml} and \ac{nlp} \cite{fortuna_survey_2019,jahan_systematic_2023,albladi_hate_2025}.

However, the majority of works on models for textual data address this as a binary task, \ie, classifying whether a text message includes hate speech or not \cite{jahan_systematic_2023,alkomah_literature_2022,van_aken_challenges_2018}. Yet, this does not entirely meet the requirements of most application domains. In content moderation, the urgency and extent of sanctioning harmful content, ranging from warnings and content deletion to the suspension of the responsible user profile, often depends on the severity of the specific content \cite{scheuerman_framework_2021}. A fine-grained classification of hate speech subtypes can therefore provide an extended basis for decision-making and subsequent justification \cite{agarwal_toxic_2020}. 
In the law enforcement and reporting center domain, there is particular interest in determining which social groups are most frequently targeted by hate speech in order to generate nuanced situational pictures for decision-makers, derive trends, and adapt counter-strategies accordingly \cite{baumler_cyber_2025}. In addition, it is particularly relevant to differentiate content relevant under criminal law or content that could be indicative of an immediate threat, such as hate speech inciting violence \cite{baumler_towards_2024}.
As soon as there is the need to classify the type, target, severity, or legality of hate, binary classification is no longer sufficient. In recent literature, hate speech classification has therefore often been addressed as a multi-class problem by assigning a sample to one of the classes that represent types or targets of hate \cite{van_aken_challenges_2018}. Yet, dealing with hate speech classification on this level as a multi-class task is problematic because it assumes labels to be mutually exclusive. This contradicts the reality that a text can include multiple accounts of hate speech with varying severity and hostility patterns that target different group affiliations of the same individual \cite{burnap_us_2016,salminen_anatomy_2018,van_aken_challenges_2018,liu_fuzzy_2019}. Similarly, this applies to criminally relevant instances of hate speech. While not all jurisdictions criminalize drastic instances of it \cite{chetty_hate_2018,zufall_legal_2022}, some criminal codes, \eg, the German, may penalize it under various criminal norms, some of which may apply simultaneously \cite{demus_comprehensive_2022,baumler_towards_2024}.

Thus, in this survey, we focus on multi-label hate speech classification, a setting in which multiple labels can be associated with a single sample. Hate speech classification has already been the subject of several literature surveys. While classification in multilingual textual \cite{narula_comprehensive_2025} and multi-modal \cite{hermida_detecting_2023,hee_recent_2024} data has attracted growing interest in recent years, there are, to the best of our knowledge, no surveys focusing on multi-label hate speech classification. Surveys of models are primarily concerned with binary detection, whereby multi-label tasks are either not mentioned \cite{ayo_machine_2020,fortuna_survey_2019,ramos_comprehensive_2024,schmidt_survey_2017} or only peripherally addressed \cite{albladi_hate_2025,alkomah_literature_2022,mansur_twitter_2023,rawat_hate_2024}. A similar picture emerges with regard to an overview of suitable training data. Datasets with multi-label annotations are occasionally included in surveys with a broader scope \cite{poletto_resources_2021,jahan_systematic_2023,hee_recent_2024,albladi_hate_2025,alkomah_literature_2022}, but are not systematically analyzed and compared with a focus on their multi-label character. 
Therefore, we present the first systematic literature survey on multi-label classification of textual hate speech in English language. Thereby, we will provide a comprehensive and thorough assessment of the state-of-the-art for both available datasets and \ac{ml} models. By presenting existing resources and best practices, identifying deficiencies, and providing an outlook on open research opportunities, we want to facilitate and motivate further research. Following the transparent and reproducible approach of \citet{vom_brocke_standing_2015}, we analyze 46 relevant publications and provide four contributions (C1-C4) to the state of research:
\begin{itemize}
    \item A concise overview of available textual datasets (N=28) that can be leveraged for the training of \ac{ml} models for multi-label hate speech classification tasks in English. This includes datasets focusing on hate speech sub-types and related concepts. Their systematic comparison by dimensions like size, data source, meta-concept, or label-set structure can support the identification of suitable datasets for varying objectives (C1).
    \item A comparative overview of works presenting \ac{ml} models (N=24) for these tasks, with particular consideration of employed evaluation metrics. This illustrates the state of research and allows for an initial assessment of advantages and deficits of different approaches (C2).
    \item An identification and discussion of open issues in research on multi-label hate speech classification. Thereby we indicate how choices regarding labeling, model architecture, or training may provide mitigation and explore major methodological incoherencies (C3).
    \item Ten recommendations for the advancement of research on multi-label hate speech classification in textual data. They indicate both potentials for methodological improvement and directions for future research (C4).
\end{itemize}

Our work is structured as follows. We will first provide some background on both the concept of hate speech and multi-label classification (\secref{sec:review:background}). Then, we outline our methodology for literature search, selection, and analysis (\secref{sec:review:method}). Based on our analysis, we describe the state of the art regarding datasets and models for textual multi-label hate speech classification in English (\secref{sec:review:results}). On that basis, we identify open issues in research, derive recommendations, and outline our limitations (\secref{sec:review:discussion}). We close the paper with a brief conclusion (\secref{sec:review:conclusion}).
\section{Background}
\label{sec:review:background}
Since this paper surveys the current state of research on multi-label hate speech classification in English language textual data, it seems reasonable to first provide some background on hate speech as a concept (\secref{sec:review:background:hatespeech}) and on multi-label classification as a machine learning task (\secref{sec:review:background:multilabel}).

\subsection{Hate Speech}
\label{sec:review:background:hatespeech}
Hate speech is a contested concept \cite{siegel_online_2020}. On the one hand, this results in different legal definitions that vary by jurisdiction \cite{chetty_hate_2018}, often reflecting local cultural perceptions and legal traditions \cite{gagliardone_countering_2015}. On the other hand, social media platforms also formulate varying definitions that form the basis of their content moderation practices \cite{gagliardone_countering_2015,sellars_defining_2016}. These are often vague, allowing flexibility in dealing with unwanted speech and user behavior \cite{sellars_defining_2016}. Finally, hate speech definitions in academia are often shaped by the respective research motivation \cite{sellars_defining_2016}. This can be problematic, as underlying definitions shape multiple aspects of \ac{ml}-based hate speech classification research.
First, as \citet[p. 302]{laaksonen_datafication_2023} puts it: ``To build an automated system to identify hate speech, hate needs to be datafied''. The systems need to operate on data and hate speech must be defined in correspondence with this data to allow for its identification by any \ac{ml} system. 
Second, a reasonably precise definition improves the labeling of the training data because it can potentially reduce annotator differences by decreasing the need for individual interpretation \cite{macavaney_hate_2019}. 
Third, definitions used in this research context should be as consistent as possible. Currently, datasets are often underpinned by different understandings of the concept \cite{banko_unified_2020}. Varying definitions can result in different labeling approaches, implying that the same labels can have different underlying meanings. This impairs the comparative evaluation of \ac{ml} classification models \cite{macavaney_hate_2019} and limits their generalizability and applicability across different data \cite{yin_towards_2021}.

Considering the scope of this work, developing a novel definition is unattainable. Instead, selecting a suitable definition appears to be sufficient. There are, however, several challenges. First, even though in research there is tentative consensus that hate speech attacks targets based on group membership or identity \cite{macavaney_hate_2019,sellars_defining_2016}, there is disagreement whether only pre-determined 'protected' groups, \ie, minorities, or any kind of groups can be targeted \cite{paasch-colberg_insult_2021}. Second, the term is used to refer to a variety of negative or aggressive statements \cite{paasch-colberg_insult_2021}. Third, hate speech is frequently conflated with broader concepts, such as toxicity, abusive and offensive language, or specific forms of group-related hostility, \eg, racism \cite{poletto_resources_2021}. Fourth, from a theoretical perspective, hate speech definitions typically focus on either the articulator’s intention, the content of the speech and its form, or the effect on the victim \cite{paasch-colberg_insult_2021,poletto_resources_2021}.
To inform a systematic survey, we argue that a definition should account for the following:
\begin{enumerate}
    \item It should define hate speech with reference to its content and form. \ac{ml} models can only classify hate speech based on content and linguistic characteristics of the respective textual samples. In many instances, this does not permit an assessment of articulators' intentions and effects on victims \cite{poletto_resources_2021}.
    \item It should explicitly cover more subtle instances of hate speech. Hateful expressions can justify or reinforce discrimination against groups, even if they do not include evidently offensive language or are working with negative stereotypes, metaphors, irony, or humor \cite{nielsen_subtle_2002,vidgen_challenges_2019}.
    \item It should not narrow down societal groups that can be targeted by hate speech, as such delimitations are often culture-specific and novel targets can emerge over time \cite{fortuna_survey_2019}.
    \item It should not be limited to individual national contexts and application domains, \eg, law enforcement, content moderation, and should account for the perspectives of various relevant stakeholders.
\end{enumerate}

In light of these considerations, we adopt the definition of \citet[p. 5]{fortuna_survey_2019}, which, in context of a survey on textual hate speech detection, was derived from an analysis of common elements among definitions from scientific sources, social networks, minority organizations, and supranational political entities:
\begin{quote}
``Hate speech is language that attacks or diminishes, that incites violence or hate against groups, based on specific characteristics such as physical appearance, religion, descent, national or ethnic origin, sexual orientation, gender identity or other, and it can occur with different linguistic styles, even in subtle forms or when humour is used.''
\end{quote}
This definition recognizes the ambiguity of hate speech, allows for various nuances, and spans a broad spectrum of online content. It reflects that hate speech can have different manifestations, especially regarding the targeted social groups or group identities, its severity, and its criminalization. The significance of such differentiations for many potential application domains, such as content moderation, law enforcement, or scientific research \cite{agarwal_toxic_2020,demus_comprehensive_2022,baumler_towards_2024}, underscores the relevance of multi-label classification approaches and motivates this survey.

\subsection{Multi-label Classification}
\label{sec:review:background:multilabel}
In contrast to single-label learning like in binary and multi-class classification, where each example is associated with a single label, examples in multi-label classification are associated with a set of labels \cite{tsoumakas_multi-label_2007}. 
\ac{ml} research approached multi-label classification in many different ways. \citet{tsoumakas_multi-label_2007} have grouped the applied methods into algorithm adaptation methods and transformation methods. \textit{Algorithm adaptation methods} are straightforward, as they address the multi-label classification problem directly. For example, \citet{abburi_fine-grained_2021} trained a sexism classifier using a sigmoid activation in their output layer of size $|L|$ where L is the set of labels. In this case, a single model outputs a value between 0 and 1 for each label, which can be interpreted as the extent to which the respective form of sexism is present in the provided example.
\textit{Transformation methods} instead rely on casting the multi-label classification problem into one or more single-label classification or regression problems \cite{tsoumakas_multi-label_2007}. The most common transformation methods used for multi-label hate speech classification transform the problem into multiple binary problems or a single multi-class problem.

Transforming a multi-label classification problem with N labels into N binary classification problems is called \textit{\ac{br}}. The idea is to train one binary classifier per label that decides if the respective label should be associated with a given example or not. For unseen examples, \ac{br} queries all binary classifiers and predicts the corresponding label set as the combination of all labels that have been associated with the example by the label-specific binary classifiers \cite{zhang_review_2014}. To train the binary classifiers, one dataset $D_l$ per label $l$ is created on the basis of a multi-label dataset by assigning all examples having $l$ in their label-set to the positive and all others to the negative class \cite{zhang_review_2014}. While \ac{br} is popular due to its simplicity, it ignores potential correlations among labels \cite{zhang_review_2014} and is computationally expensive \cite{abburi_fine-grained_2021}.
    
A straightforward transformation method to cast a multi-label classification problem into a multi-class one is called \textit{\ac{lp}}. \ac{lp} treats every distinct label set occurring in a multi-label dataset as a new class. For a multi-label dataset with $N$ binary labels, \ac{lp} therefore divides the data according to their label sets into at most $2^N$ classes, which represent all possible label combinations. One downside of \ac{lp} is that it can result in an exponential increase in the number of classes, which could result in a dataset that only contains a few examples per class \cite{tsoumakas_multi-label_2007}.

In this work, we aim to survey and further research on hate speech classification that acknowledges potential co-occurrences of different subtypes of hate speech. Given the previously outlined relation between binary, multi-class, and multi-label problems, we will also consider (besides datasets with 'genuine' multi-label annotations) multi-class datasets that constitute a \ac{lp} as well as binary datasets that focus on specific forms of hate speech and could be used for a multi-domain multi-label classification by utilizing \ac{br}. This inclusion of other datasets is especially relevant as most of the available datasets for hate speech classification have been designed to tackle binary or multi-class classification problems \cite{poletto_resources_2021}.
In the context of this paper, we refer to \textit{multi-domain} learning for scenarios in which a model is trained on the same task (hate speech classification) within multiple domains, \eg, datasets that differ in their underlying distribution \cite{pan_survey_2010}. This could be the case when training a general hate speech classification model on both general hate speech datasets and datasets on more specific sub-categories of hate speech, \eg, misogyny.
\section{Method}
\label{sec:review:method}
In the following, we outline the procedure that was employed when conducting this systematic survey. To satisfy the quality dimensions of rigor, relevance, and internal consistency \cite{pare_synthesizing_2015}, we justify and reflect on important decisions. We employed an iterative process based on scientific databases that used a combination of keyword-based search and forward- and backward-search to cover a representative set of papers. By doing this, we follow the systematic approach described by \citet{vom_brocke_standing_2015}. The literature search was conducted between 2024-04-26 and 2024-05-04 and followed a two-stage approach. In the \textit{first stage}, we executed a keyword-based search in a selection of relevant scientific databases (ACM Digital Library, IEEE-Xplore, ScienceDirect, and Springer Link). We selected these databases for two reasons: (a) they encompass most of the prominent \ac{ml} journals and conference proceedings, and (b) databases from other relevant publishers, such as ACL or AAAI, provided limited search functionality, lacking features like boolean operators and advanced keyword search capabilities. The full-text and metadata search in the selected databases was not limited to a specific time frame of publication and employed the following boolean search string: \textit{(``hate speech'' OR ``hateful speech'' OR ``racism'' OR ``sexism'' OR ``offensive language'' OR ``abusive language'' OR ``cyberbullying'') AND (``detection'') AND (``classification'' OR ``machine learning'' OR ``machine-learning'' OR ``dataset'') AND (``multi-label'' OR ``multilabel'')}\footnote{The string was syntactically adapted according to the requirements of the respective database. Furthermore, ScienceDirect only allows a limited number of boolean-clauses to be used, so the string was split and queried multiple times with variations.}. The procedure of publication search and selection is illustrated in \tabref{tab:review:method}.

\captionsetup[table]{position=top}
\begin{table}[ht]
\caption{Procedure of publication selection for the systematic literature review, differentiated by database. AAAI Digital Library, ACL Anthology, and other databases were not queried in the initial search, as they had limited or no available search functionality.}
\label{tab:review:method}
\centering
\small
\begin{tabular}{l c c c c c}
\hline
\textbf{Database} & \textbf{Initial Results} & \textbf{Abstract Screening} & \textbf{Relevant} & \textbf{Backward} & \textbf{Forward}\\
\hline
 ACM Digital Library & 127 & 13 & 6 & 1 & 2 \\
 IEEE Xplore         & 295 & 22 & 4 & 3 & - \\
 ScienceDirect       &  94 & 12 & 1 & - & - \\
 Springer Link       & 187 & 24 & 6 & 2 & - \\
\hline
 AAAI Digital Library & - & - & - &  3 & - \\
 ACL Anthology       & - & - & - & 11 &  3 \\
 Other               & - & - & - &  4 & - \\
 \hline
 \textbf{\(\sum\)}   & 703 & 71 & 17 & 24 & 5 \\
\hline
\end{tabular}
\end{table}

The initial search resulted in 703 preliminary results. In a next step, the articles' abstracts were screened by one researcher to identify and remove out-of-scope publications. The remaining 71 papers were then read and the authors decided collectively on their final inclusion. Publications were included if (1) they are peer-reviewed and published\footnote{Datasets published in pre-prints or repositories were also included.}, (2) a \ac{ml} model or a novel training dataset for hate speech classification is presented, (3) at least one of the used labels refers to our conception of hate speech\footnote{Hatred related to targets' group affiliation or identity (see \secref{sec:review:background:hatespeech}).} or subtypes of it (\eg, sexism, racism), (4) multi-label or multi-label equivalent (\eg, powersets) annotations are used, (5) only textual data is used in at least one of the datasets/models presented, and (6) both the publication and a significant part ($\geqq$ 25\%) of the used or introduced dataset(s) is in English. We only considered datasets and papers that use English to ensure that we possess the linguistic competence and a baseline knowledge of the cultural context. This is necessary for being able to evaluate the employed classes/labels with regard to their significance to our conception of hate speech. Furthermore, we explicitly excluded works if (1) they are inaccessible, (2) they are exclusively focused on cyberbullying, offensive language, or aggressive language, (3) their models or datasets are specifically for code-mixed language (\eg, ``Hinglish''), (4) multi-label datasets are used for the training of multi-class or binary models, (5) their documentation is not sufficient to determine whether models or datasets are multi-label. If there were disagreements regarding the final inclusion of a publication, a majority vote was taken among the authors.

The first stage resulted in 17 relevant publications. This set of papers was then enriched in a \textit{second stage} through both forward and backward searches \cite{vom_brocke_standing_2015}. For backward search, we did one step (search-depth of one) starting with our preliminary publication set from the first stage. We screened papers that were either cited by them within the context of multi-label classification or generally referenced by them and contained ``multi-label'' in their titles. For the forward search, we also did one step (search-depth of one) from the same starting point and screened papers that cited initially selected papers and had more than 50 citations, according to Google Scholar. We applied the same criteria for screening these works as for the first stage. The second stage resulted in the addition of 29 relevant publications. Altogether, 46 publications were surveyed for this work, which present 28 datasets and 24 models.

For systematically \textit{analyzing} the literature, we created two separate spreadsheets in Excel to collect data for both datasets and models on key dimensions of interest to our study, which were initially generated by screening related survey papers (\eg, \cite{jahan_systematic_2023,poletto_resources_2021,fortuna_survey_2019,zhang_review_2014}), and then revised inductively during engagement with the publications. Besides meta-information like publication year or dataset language, our initial analytical framework included common analytical categories like data source, dataset size, model type, or multi-label approach. We further adopted the dimension meta-concept from \citet{kumar_multilingual_2023}. During the analysis, we expanded the framework in three regards. First, we introduced dimensions particularly relevant in multi-label contexts, \eg, label-set structure or \ac{mt} learning. Second, for datasets we introduced additional dimensions to increase analytical depth beyond their meta-concept, \ie, definitions and labels. Third, we recognized significant heterogeneity in annotation processes, \ac{iaa} calculation, and evaluation metrics, thus complementing corresponding dimensions. The framework provided the foundation for our presentation and contextualization of the state of research in \secref{sec:review:results}. Finally, to identify open issues and research recommendations for \secref{sec:review:discussion}, we continuously reflected on reported challenges, methodological inconsistencies, and research gaps.
\section{Results}
\label{sec:review:results}
In the following, we present our results for the datasets (\secref{sec:review:results:datasets}) and \ac{ml} models (\secref{sec:review:results:models}) introduced in the surveyed works. The description of the state of research on both is supported by comprehensive tabular comparisons.

\subsection{Datasets}
\label{sec:review:results:datasets}
The publications relevant to this survey introduced 28 different datasets that can be leveraged to train multi-label hate speech classification models. While only one paper in the initial set introduced a dataset, we discovered 23 through the backward search, and four through the forward search. Ten datasets were associated with a shared task. In \tabref{tab:review:datasets}, we give an overview. In the following subsections, we compare the datasets across ten dimensions.

\subsubsection{Language(s), Year, and Size}
Whereas 20 datasets only contain textual content in English \textit{language}, eight were multi-lingual with an English subset. The multi-lingual datasets also included subsets in Spanish (2), French (2), German (2), Hindi (2), Italian (2), Arabic (1), and Bengali (1).

\begin{figure}[ht]
    \centering
    \begin{minipage}{0.45\textwidth}
        \centering
        \includegraphics[scale=0.4]{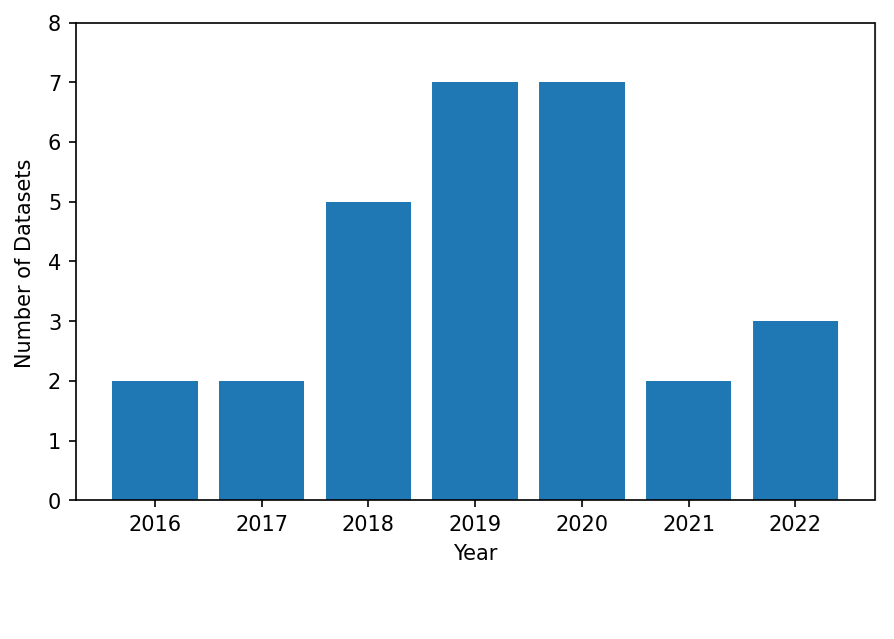}
        \caption{Number of datasets published per year.}
        \label{fig:review:dataset_year}
    \end{minipage}
    \begin{minipage}{0.45\textwidth}
        \centering
        \includegraphics[scale=0.4]{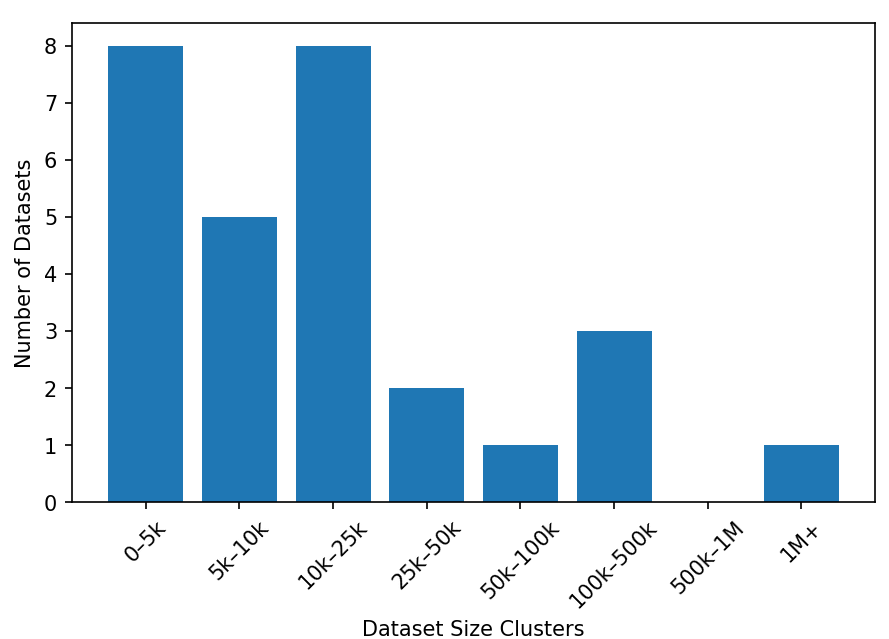}
        \caption{Frequency of dataset sizes (clustered).}
        \label{fig:review:dataset_sizes}
    \end{minipage}
\end{figure}

As regards the publication \textit{year}, the datasets covered in our survey were published in scientific papers, repositories, or as part of a shared task between 2016 and 2023 (see \figref{fig:review:dataset_year}). While the lower end of this time range may indicate the emergence of research on multi-label hate speech classification, the upper end can be attributed to limitations of our search methodology. One caveat regarding the datasets' publication time is that it differs from the time of data collection and the time of the actual content creation. In addition, the respective times are not always documented. This means that the content may have been created significantly before the publication of a dataset (\eg, \citet{alshamrani_hate_2021} consider YouTube comments from 2008 and earlier).

The \textit{size} of the analyzed datasets varies considerably (see \figref{fig:review:dataset_sizes}). While the majority (14/28) is within the 5,000 to 15,000 data point range (for the English subset, if multi-lingual), the ``Unintended Bias in Toxicity Classification'' dataset, which was part of the Kaggle 2019 shared task \cite{cjadams_jigsaw_2019}, is a noteworthy outlier with more than 1.8 Million data points \cite{abbasi_deep_2022}. A common approach for creating datasets is to compile an initial collection of samples that might be suitable for labeling \cite{kennedy_introducing_2022}. These collections can be considerably more extensive than the final datasets. They are then often reduced to get a more ``manageable'' dataset. This manageability is primarily a matter of available resources and the capacity to label the data manually. Further details on data annotation are given below.

\subsubsection{Data Source(s)}
\label{sec:review:results:datasets:source}
By far the most popular data source is Twitter/X (12/28). This may have two reasons. First, it formerly had a comparatively open and powerful \ac{api}, which made data access for researchers relatively easy and inexpensive \cite{pfeffer_this_2023}. 
Second, the formerly prescribed format limitations, especially regarding tweet length (first 128 and then 256 characters), maintained a useful level of textual feature for efficient data annotation, since they are more verbose and natural than purely considering n-grams but still relatively short.
Other datasets encompass comments from other websites, social media, and web services, including Wikipedia \cite{cjadams_jigsaw_2019,ibrahim_imbalanced_2018,cjadams_toxic_2017,wulczyn_ex_2017}, YouTube \cite{gupta_hateful_2023,bhattacharya_developing_2020}, Gab \cite{kennedy_introducing_2022,mathew_hatexplain_2021}, Facebook \cite{mandl_overview_2019}, BitChute \cite{gupta_hateful_2023}, Github \cite{cheriyan_towards_2021}, Gitter \cite{cheriyan_towards_2021}, Stack Overflow \cite{cheriyan_towards_2021}, and Slack \cite{cheriyan_towards_2021}. Further data sources are crowd-based aggregation platforms like ``everydaysexism.com'' \cite{parikh_multi-label_2019} or ``Hatebusters'' \cite{mollas_ethos_2022}, which collect stories or social media posts regarding a specific topic or type of hate speech and can thus be used as a second-hand data repository.
Finally, some datasets do not use content from real-world sources. Instead, their authors opted for an artificial generation of hateful and not-hateful samples \cite{vidgen_learning_2021,rottger_hatecheck_2021,fanton_human---loop_2021,chung_conan_2019}.
Further details regarding their procedure will be discussed below.
    
\subsubsection{Meta-Concept}
The creation of a tag/label set for data annotation can be seen as the operationalization of a meta-concept that should be represented through the data in a way that renders this concept ``learnable'' for a model \cite{kumar_multilingual_2023}.
As discussed in \secref{sec:review:background:hatespeech}, various available hate speech definitions try to capture the same meta-concept. Among the 28 analyzed datasets, only eleven\footnote{For some papers, the stated meta-concept and the actually used label set are not completely congruent. Therefore some papers are assigned to multiple groups regarding their meta-concept.}
try to capture hate speech as a meta-concept. This is a consequence of our selection criteria for publications.

A second group of nine datasets addresses meta-concepts like misogyny \cite{fersini_overview_2018, silberztein_automatic_2018, waseem_are_2016, waseem_hateful_2016, parikh_multi-label_2019, basile_semeval-2019_2019, bhattacharya_developing_2020, arampatzis_overview_2023}, racism \cite{waseem_are_2016, waseem_hateful_2016, basile_semeval-2019_2019}, or islamophobia \cite{chung_conan_2019}, therefore capturing specific sub-types of hate speech. Even though they are not particularly useful for training generalized hate speech classification models, they nevertheless merit attention as they could be beneficial in a multi-domain approach and, at least partially, capture far more nuances related to the specific sub-concept than more generic hate speech datasets. These nuances might be instrumental in a multi-label scenario, where they could be used to specifically capture one hate speech sub-type or, depending on the label-set used, one label (\eg, a sexism label as a type of hate speech).\footnote{For instance, one could train a hate speech classification model using a one-dimensional, multi-label label-set with different subtypes of hate being the labels (\eg, sexism, homophobia, racism). In this context, a specific ``sexism'' dataset could be used in a multi-domain training setting to improve model performance, especially on the ``sexism''-label.}

\afterpage{%
\clearpage
{\small
\begin{landscape}
  \setlength\LTleft{0pt} 
  \setlength\LTright{0pt} 
  \captionof{table}{Surveyed datasets that can be used for multi-label hate speech classification in English.}
  \addtocounter{table}{-1}  
\label{tab:review:datasets}
  \begin{longtable}{%
    p{2.5cm}p{0.5cm}p{0.55cm}p{0.8cm}p{1.6cm}p{0.7cm}p{1.6cm}p{1.5cm}p{4.8cm}p{1.0cm}p{0.8cm}%
  }
  \\[0.5ex]
  \hline
  \textbf{Name} & \textbf{Ref.} & \textbf{Year} & \textbf{Size} & \textbf{Source} & \textbf{Lang.} & \textbf{IAA} & \textbf{IAA Method} & \textbf{Labels} & \textbf{Online} & \textbf{Link} \\
  \hline
  \endfirsthead
  
  \multicolumn{11}{c}{{\tablename\ \thetable{} -- \textit{continued from previous page}}} \\
  \hline
  \textbf{Name} & \textbf{Ref.} & \textbf{Year} & \textbf{Size} & \textbf{Source} & \textbf{Lang.} & \textbf{IAA} & \textbf{IAA Method} & \textbf{Labels} & \textbf{Online} & \textbf{Link} \\
  \hline
  \endhead
  
  \hline \multicolumn{11}{r}{{\textit{Continued on next page}}} \\ \hline
  \endfoot
  
  \hline
  \endlastfoot
  Wasseem and Hovy (2016) &
  \cite{waseem_hateful_2016} &
  2016 &
  16,914 &
  Twitter &
  EN &
  $\kappa$=0.84 &
  not given &
  (Racism, Sexism, Neither)\textsubscript{multi-class} &
  Yes & 
  \href{https://github.com/zeeraktalat/hatespeech}{\textbf{[link]}} \\
  \hline
  
  Wasseem (2016) &
  \cite{waseem_are_2016} &
  2016 &
  6,909 &
  Twitter &
  EN &
  $\kappa$=0.57 &
  not given &
  (Racism, Sexism, Neither, Both)\textsubscript{multi-class} &
  Yes & 
  \href{https://github.com/zeeraktalat/hatespeech}{\textbf{[link]}} \\
  \hline
  
  Davidson et al. (2017) &
  \cite{davidson_automated_2017} &
  2017 &
  24,783/ \newline 24,802\footnote{Some samples have been retroactively discarded.} &
  Twitter &
  EN &
  92\% &
  Given by CrowdFlower &
  (Hatespeech, Offensive, Neither)\textsubscript{multi-class} &
  Yes &
  \href{https://github.com/t-davidson/hate-speech-and-offensive-language/tree/master/data}{\textbf{[link]}} \\
  \hline
  
  Toxic Comment Classification Challenge (TCCC) &
  \cite{cjadams_toxic_2017} &
  2018 &
  312,737 &
  Wikipedia Comments &
  EN &
  - &
  - &
  (Toxic, Severe Toxic, Obscene, Threat, Insult, Identity Hate)\textsubscript{multi-label} &
  behind login &
  \href{https://www.kaggle.com/c/jigsaw-toxic-comment-classification-challenge/data}{\textbf{[link]}} \\
  \hline
  
  Kaggle+ &
  \cite{ibrahim_imbalanced_2018} &
  2018 &
  >215k\footnote{The number of augmented samples that have been added to the original dataset is not given.} &
  Wikipedia Comments &
  EN &
  no annotation done &
  - &
  Toxic\textsubscript{binary} \newline
  (Toxic, Severe Toxic, Obscene, Threat, Insult, Identity Hate)\textsubscript{multi-label} &
  No &
  - \\
  \hline
  
  Anzovino et al. (2018) &
  \cite{silberztein_automatic_2018} &
  2018 &
  4,454 &
  Twitter &
  EN &
  \makecell[tl]{det=0.487\\class=0.373} &
  {PABAK\-OS} (for detection and classification) &
  Misogyny\textsubscript{binary} \newline
  (Discredit, Sexual Harassment \& Threats of Violence, Stereotype \& Objectification, Dominance, Derailing)\textsubscript{multi-class} &
  behind login &
  \href{https://amiibereval2018.wordpress.com/important-dates/data/}{\textbf{[link]}} \\
  \hline
  
  EVALITA 2018 &
  \cite{fersini_overview_2018} &
  2018 &
  4k per lang. &
  Twitter &
  EN, IT &
  \makecell[tl]{misogyny\\
  =0.81,\\
  misogyny\\
  \_category \\
  =0.45, \\ 
  target=0.49} &
  not given &
  Misogyny\textsubscript{binary} \newline
  (Discredit, Sexual Harassment \& Threats of Violence, Stereotype \& Objectification, Dominance, Derailing)\textsubscript{multi-class} \newline
  Targeted\textsubscript{binary} &
  behind login &
  \href{https://groups.google.com/g/amievalita2018}{\textbf{[link]}}  \\
  \hline
  
  Founta et al. (2018) &
  \cite{founta_large_2018} &
  2018 &
  80k &
  Twitter &
  EN &
  \makecell[tl]{
  55.9\% over-\\whelming,\\
  36.6\% nor-\\mal,\\
  7.5\% small\footnote{The authors use a simple majority vote on five crowdsourced judgments and give the IAA as ``[...]($\sim$ 55.9\%) achieve an overwhelming agreement, which means that at least 4 out of 5 annotators agreed on the label. The remaining $\sim$36.6\% of tweets reach an agreement of more than 3 out of 5 votes and only very few ($\sim$7.5\%) achieve a majority with only two annotators.''}} &
  not given &
  (Abusive, Hateful, Normal, Spam)\textsubscript{multi-class} &
  outdated &
  \href{http://ow.ly/BqCf30jqffN}{\textbf{[link]}} \\
  \hline
  
  HASOC2019 &
  \cite{mandl_overview_2019} &
  2019 &
  $\sim$8k per lang. &
  Facebook, Twitter &
  EN, DE, HI &	
  \makecell[tl]{EN=72\%,\\
  DE=96\%,\\
  HI=83\%\footnote{Only over at least twice annotated samples.}} &
  not given &
  Hate/Offensive\textsubscript{binary}\newline
  Targettd\textsubscript{binary}\newline
  (Hate, Offensive, Profane)\textsubscript{multi-class} &
  behind login &
  \href{https://hasocfire.github.io/hasoc/2019/dataset.html}{\textbf{[link]}} \\
  \hline
  
  Unintended Bias in Toxicity Classification &
  \cite{cjadams_jigsaw_2019} &
  2019 &
  $\sim$1.8m &
  Civil Comments Platform (Wikipedia Comments) &
  EN &
  - &
  - &
  Toxic\textsubscript{[0,1]} \newline
  Severe Toxic\textsubscript{[0,1]} \newline
  (Severe Toxic, Obscene, Threat, Insult, Identity Attack, Sexually Explicit)\textsubscript{each [0,1]} \newline
  Identity: (24 classes)\textsubscript{[0,1]} &
  behind login &
  \href{https://www.kaggle.com/c/jigsaw-unintended-bias-in-toxicity-classification/data?select=all_data.csv}{\textbf{[link]}} \\
  \hline
  CONAN &
  \cite{chung_conan_2019} &
  2019 &
  14,988/ 4,078/ 4,078 &
  artificially generated &
  EN, FR, IT &
  $\kappa$=0.92\footnote{Over all three languages.} &
  Cohen's Kappa &
  (Crimes, Culture, Economics, Islamization, Rapism, Terrorism, Women Opression, Generic)\textsubscript{multi-class} &
  Yes &
  \href{https://github.com/marcoguerini/CONAN}{\textbf{[link]}} \\
  \hline
  
  Parikh et al. (2019) &
  \cite{parikh_multi-label_2019} &
  2019 &
  13,023 &
  everyday sexism.com &
  EN &
  $\kappa$=0.584\footnote{Average of the Cohen’s Kappa scores for the per-category pairs of binary label vectors.} &
  Cohen’s Kappa &
  Misogyny Type: (23 classes)\textsubscript{multi-label} &
  outdated &
  \href{https://irel.iiit.ac.in/sexism-classification}{\textbf{[link]}} \\
  \hline
  
  HatEval &
  \cite{basile_semeval-2019_2019} &
  2019 &
  13k/ 6.6k &
  Twitter &
  EN, ES &
  HS=0.83, TR=0.70, AG=0.73 &
  F8 reported average confidence\footnote{``(i.e., a measure combining inter-rater agreement and reliability of the contributor).'' \cite{basile_semeval-2019_2019}} &
  Hate Speech\textsubscript{binary} \newline
  Targeted\textsubscript{binary} \newline
  Aggressive\textsubscript{binary} &
  Yes &
  \href{https://github.com/msang/hateval/tree/master/SemEval2019-Task5/datasets}{\textbf{[link]}} \\
  \hline
  
  Ousidhoum et al. (2019) &
  \cite{ousidhoum_multilingual_2019} &
  2019 &
  5,647/ 4,014/ 3,353 &
  Twitter &
  EN, FR, AR &
  EN=0.153, FR=0.244, AR=0.202 &
  Krippen-dorff's alpha &
  Directness\textsubscript{binary} \newline
  (Abusive, Hateful, Offensive, Disrespectful, Fearful, Normal)\textsubscript{multi-label} \newline
  Target attribute: (8 classes)\textsubscript{multi-class} \newline
  Target group: (13 classes)\textsubscript{multi-class} \newline
  Annotator Sentiment: (7 classes)\textsubscript{multi-label} &
  Yes &
  \href{https://github.com/HKUST-KnowComp/MLMA_hate_speech}{\textbf{[link]}} \\
  \hline
  
  Offensive Language Identification Dataset (OLID) &
  \cite{zampieri_predicting_2019} &
  2019 &
  14.2k &
  Twitter &
  EN &
  \textit{``Two annotators agreed for $\sim$60\%''} &
  - &
  Offensive\textsubscript{binary} \newline
  Targeted\textsubscript{binary} \newline
  (Individual, Group, Other)\textsubscript{multi-class} &
  Yes &
  \href{https://drive.google.com/file/d/1Tksi8UyzW-drFWd7maGr7MoHVa-VHQCO/view}{\textbf{[link]}} \\
  \hline
  
  HASOC2020 &
  \cite{mandl_overview_2020} &
  2020 &
  4,522/ 2,899/ 3,626 &
  Twitter &
  EN, DE, HI &
  ``Disagree-ment $\in[0.16, 0.34]$'' &
  - &
  Hate/Offensive\textsubscript{binary} \newline
  (Hate, Offensive, Profane)\textsubscript{multi-class} &
  behind login &
  \href{https://hasocfire.github.io/hasoc/2020/dataset.html}{\textbf{[link]}} \\
  \hline
  
  TRAC-2 &
  \cite{bhattacharya_developing_2020} &
  2020 &
  >9,275/ >5,175/ >9,775\footnote{Authors give a percentage of ''over 25,000 comments''. 25,000 was used as basis for calculating the number of samples. Code-mixed samples not included.} &
  YouTube Comments &
  EN, HI, BN &
  \textit{``unanimous in $\sim$75\% or more cases''} &
  - &
  (Overtly aggressive, Covertly aggressive, Non-aggressive)\textsubscript{multi-class} \newline    
  Gendered/Misogynous\textsubscript{binary} &
  behind login &
  \href{https://sites.google.com/view/trac2/shared-task}{\textbf{[link]}} \\
  \hline
  
  Multitarget CONAN &
  \cite{fanton_human---loop_2021} &
  2021 &
  5,003 &
  - &
  EN &
  - &
  - &
  Target: (9 classes)\textsubscript{multi-class} &
  Yes &
  \href{https://github.com/marcoguerini/CONAN}{\textbf{[link]}} \\
  \hline
  
  HateCheck &
  \cite{rottger_hatecheck_2021} &
  2021 &
  3,901 &
  Generated &
  EN &
  $\kappa$=0.93 &
  Fleiss’ Kappa &
  Hateful\textsubscript{binary} \newline
  Target: (7 classes)\textsubscript{multi-class} \newline
  Targeted\textsubscript{binary} &
  Yes &
  \href{https://github.com/paul-rottger/hatecheck-data}{\textbf{[link]}} \\
  \hline
  
  Dynamically-Generated-Hate-Speech-Dataset &
  \cite{vidgen_learning_2021} &
  2021 &
  41,255 &
  Generated by annotated HitL RoBERTa &
  EN &
  $\alpha$=0.52 &
  Krippen-dorff’s alpha &
  Hateful\textsubscript{binary} \newline
  (Derogation, Animosity, Threatening language, Support for hateful entities, Dehumanization)\textsubscript{multi-class} \newline
  Target of Hate: (34 classes)\textsubscript{multi-class} &
  Yes &
  \href{https://github.com/bvidgen/Dynamically-Generated-Hate-Speech-Dataset}{\textbf{[link]}} \\
  \hline
  
  ETHOS &
  \cite{mollas_ethos_2022} &
  2021 &
  433/ 998\footnote{433 samples with multi-label labels, 998 with only a binary isHate label.} &
  YouTube (Hatebusters), Reddit &
  EN &
  $\in[0.84, 0.98]$ depending on label &
  Fleiss’ kappa &
  Hateful\textsubscript{binary} \newline
  (Violence, Directed vs Generalised, Gender, Race, National origin, disability, Sexual Orientation, Religion)\textsubscript{multi-label} &
  Yes &
  \href{https://github.com/intelligence-csd-auth-gr/Ethos-Hate-Speech-Dataset}{\textbf{[link]}} \\
  \hline
  Gab Hate Corpus &
  \cite{kennedy_introducing_2022} &
  2022 &
  27,665 &
  Gab &
  EN &
  HD: $\kappa$=0.23,\newline PABAK=0.67,\newline CV: $\kappa$=0.28,\newline PABAK=0.97,\newline VO: $\kappa$=0.30,\newline PABAK=0.79 &
  Fleiss’ kappa und PABAK &
  Hate-based rhetoric: (HD, CV, NH)\textsubscript{multi-label}\newline Vulgarity/offensive language (VO)\textsubscript{binary}\newline Targeted population (RAE, NAT, GEN, REL, SXO, IDL, POL, MPH)\textsubscript{multi-label}\newline Framing (EX, IM)\textsubscript{binary} &
  Yes &
  \href{https://osf.io/edua3/}{\textbf{[link]}} \\
  \hline
  
  HateXplain &
  \cite{mathew_hatexplain_2021} &
  2021 &
  20,148\newline (9,055 Twitter, 11,093 Gab) &
  Twitter, Gab &
  EN &
  $\alpha$=0.46 &
  Krippen-dorf's alpha &
  (Hatespeech, Offensive, normal)\textsubscript{multi-class}\newline Target: (10 classes)\textsubscript{multi-label}\newline Rationales (spans) &
  Yes &
  \href{https://github.com/hate-alert/HateXplain?tab=readme-ov-file}{\textbf{[link]}} \\
  \hline
  
  EXist 2023 &
  \cite{arampatzis_overview_2023} &
  2023 &
  $\sim$4.7k per lang. &
  Twitter &
  EN, ES &
  -\footnote{No label aggregate, multiple annotations per example.} &
  - &
  Sexist\textsubscript{binary}\newline (Direct, Reported, Judgmental)\textsubscript{multi-class}\newline (Ideological and inequality, Stereotyping and dominance, Objectification, Sexual violence, Misogyny and non-sexual violence)\textsubscript{multi-label} &
  behind login &
  \href{https://nlp.uned.es/exist2023/}{\textbf{[link]}} \\
  \hline
  
  HateComment &
  \cite{gupta_hateful_2023} &
  2023 &
  2,071 &
  Youtube, BitChute &
  EN &
  -\footnote{Two annotators per sample, samples with disagreement were removed.} &
  - &
  Hate\textsubscript{binary}\newline (Community, Location, Individual, Organization)\textsubscript{multi-label} &
  Yes &
  \href{https://drive.google.com/file/d/1EUbWDUokv1CYkWKlwByUC6yIuBGUw2MN/edit}{\textbf{[link]}} \\
  \hline
  
  Cheriyan et al. (2021) &
  \cite{cheriyan_towards_2021} &
  2021 &
  2,308\footnote{Github 155/ Gitter 991/ Stack Overflow 403/Slack 759.} &
  Github, Gitter, Stack Overflow, Slack &
  EN &
  $\kappa$=0.79 &
  Cohen's Kappa on 100 samples for each source &
  Perspective API Score [0,1]\newline Personal\textsubscript{binary}\newline Racial\textsubscript{binary}\newline Swearing\textsubscript{binary} &
  Yes &
  \href{https://github.com/jithincheriyan/Offense-detection-in-SE-communities}{\textbf{[link]}} \\
  \hline
  
  Alshamrani et al. (2021) &
  \cite{alshamrani_hate_2021} &
  2021 &
  5,958 &
  Youtube &
  EN &
  -\footnote{Only one annotator.} &
  - &
  toxic\textsubscript{binary}\newline (Obscene, Threat, Insult, Identity hate)\textsubscript{multi-label} &
  No &
  - \\
  \hline
  
  Wikipedia Comment Corpus\footnote{To ensure brevity, only the personal attack part of the dataset is presented here. For toxicity and aggression refer to the original paper.} &
  \cite{wulczyn_ex_2017} &
  2017 &
  100k &
  Wikipedia &
  EN &
  $\alpha$=0.45 &
  Krippen-dorf's alpha &
  Quoting\textsubscript{binary}\newline Recipient\textsubscript{binary}\newline Third party\textsubscript{binary}\newline 
  Other\textsubscript{binary}\newline Attack\textsubscript{binary} &
  Yes &
  \href{https://figshare.com/articles/dataset/Wikipedia_Talk_Labels_Personal_Attacks/4054689}{\textbf{[link]}} \\
  \hline
  
  \end{longtable}

\raggedright
\footnotesize Abbreviations: \acs{ag} = \acl{ag}; \acs{api} = \acl{api}; \acs{ar} = \acl{ar}; \acs{bn} = \acl{bn}; \acs{class} = \acl{class}; \acs{cv} = \acl{cv}; \acs{de} = \acl{de}; \acs{det} = \acl{det}; \acs{en} = \acl{en}; \acs{es} = \acl{es}; \acs{ex} = \acl{ex}; \acs{fr} = \acl{fr}; \acs{gen} = \acl{gen}; \acs{hd} = \acl{hd}; \acs{hi} = \acl{hi}; \acs{hs} = \acl{hs}; \acs{iaa} = \acl{iaa}; \acs{idl} = \acl{idl}; \acs{im} = \acl{im}; \acs{it} = \acl{it}; \acs{k} = \acl{k}; \acs{m} = \acl{m}; \acs{mph} = \acl{mph}; \acs{nat} = \acl{nat}; \acs{nh} = \acl{nh}; \acs{pabak} = \acl{pabak}; \acs{pabakos} = \acl{pabakos}; \acs{pol} = \acl{pol}; \acs{rae} = \acl{rae}; \acs{rel} = \acl{rel}; \acs{sxo} = \acl{sxo}; \acs{tg} = \acl{tg}; \acs{vo} = \acl{vo}.
\end{landscape}
}
}

A third group of ten datasets could be used to train models for the delineation between hate speech and other concepts, \eg, toxic, abusive, or offensive content. The respective datasets covered in our work follow two different conceptual logics. On the one hand, they explicitly include hate speech and other content types and discriminate the concepts from each other (see, \eg, \cite{mandl_overview_2019,mandl_overview_2020,davidson_automated_2017,founta_large_2018}). On the other hand, they nominally aim to capture a different meta-concept, but their tag set for the respective concept includes one or more labels that could be used for classifying hate speech due their definition \cite{cjadams_toxic_2017,ibrahim_imbalanced_2018,cjadams_jigsaw_2019,zampieri_predicting_2019,bhattacharya_developing_2020,cheriyan_towards_2021,alshamrani_hate_2021,wulczyn_ex_2017}. 
For instance, the ``Unintended Bias in Toxicity Classification'' dataset \cite{cjadams_jigsaw_2019} nominally captures toxicity, but its tag set consists of labels for \textit{toxic} (binary), \textit{severe toxicity} (binary), \textit{toxicity subtype} (\textit{obscene, threat, insult, identity\_attack, sexual\_explicit}; multi-label) and \textit{identity} (\textit{male, female, transgender, other\_gender, heterosexual, homosexual\_gay\_or\_lesbian, bisexual, other\_sexual\_orientation, christian, jewish, muslim, hindu, buddhist, atheist, other\_religion, black, white, asian, latino, other\_race\_ethnicity, physical\_disability, intellectual\_or\_mental\_illness, other\_disability}; multi-label).
Given our adopted definition (see \secref{sec:review:background:hatespeech}), samples that are labeled as identity attack would fit within the conceptual boundaries of hate speech.

\subsubsection{Definitions and Labels}
\label{sec:review:results:datasets:definitions}
Inconsistencies in the definition of hate speech, as described in \secref{sec:review:background:hatespeech}, can also be found for the datasets and their associated publications. While some scholars adopt hate speech definitions from other works \cite{vidgen_learning_2021,mathew_hatexplain_2021,arampatzis_overview_2023}, others formulate their own definitions \cite{davidson_automated_2017,founta_large_2018,rottger_hatecheck_2021,kennedy_introducing_2022}. For instance, \citet{davidson_automated_2017} define hate speech as ``language that is used to expresses [sic!] hatred towards a targeted group or is intended to be derogatory, to humiliate, or to insult the members of the group.''
A notable exception are \citet{kennedy_introducing_2022}, who derive their own definition from sociological research and legal texts. Instead of providing concise definitions, others create a list of criteria for the presence of concepts that are partly derived from related literature \cite{waseem_are_2016,waseem_hateful_2016}.
However, for most datasets the adopted meta-concept and respective definition are not explicitly documented. Instead, only the definitions of used labels are reported. In case of crowdsourced annotations, this is often part of annotator instructions.
Label definitions are either quite nuanced, approaching a definition of the respective term (see, \eg, \cite{parikh_multi-label_2019,silberztein_automatic_2018,fersini_overview_2018,founta_large_2018,mandl_overview_2019,basile_semeval-2019_2019,ousidhoum_multilingual_2019,zampieri_predicting_2019,mandl_overview_2020,vidgen_learning_2021,alshamrani_hate_2021}), or simplified, mainly consisting of examples that are covered by it (see, \eg, \cite{chung_conan_2019,bhattacharya_developing_2020,fanton_human---loop_2021,mollas_ethos_2022,arampatzis_overview_2023,gupta_hateful_2023,cheriyan_towards_2021,wulczyn_ex_2017}).

\subsubsection{Label-Set Structure}
\label{sec:review:results:datasets:structure}
As can be seen in \tabref{tab:review:datasets}, the datasets adopt a broad variety of label sets. First, there are power-set tag sets that consist of binary labels in a multi-class setup \cite{waseem_are_2016,waseem_hateful_2016,davidson_automated_2017,founta_large_2018}.
Second, there are one-dimensional, multi-label tag sets (see, \eg, \cite{cjadams_toxic_2017,parikh_multi-label_2019}). Their labels describe one dimension with different manifestations, \eg, the ``toxicity degree'' \cite{cjadams_toxic_2017}. Each label itself is binary (\ie, it is true for a given sample or not), but multiple labels can apply to one sample.
Third, there are multi-dimensional or multi-level tag sets. These consist of multiple dimensions or groups of labels. Each dimension can either be binary, multi-class, or multi-label in itself. The low-level labels are binary in the sense that they are true or not for a given sample. Such tag sets can be found in the majority of the datasets covered in this survey. An example is the dataset presented in \citet{silberztein_automatic_2018}.
Fourth, there are multi-dimensional tag sets that expand beyond binary labels and instead (partially) utilize a continuous label space (\eg, [0,1]) for each label \cite{mollas_ethos_2022,cjadams_jigsaw_2019} or some of them \cite{cheriyan_towards_2021}. \citet{mathew_hatexplain_2021} take this a step further and additionally label spans \cite{mathew_hatexplain_2021}.

Given these different structures, the ``Label'' column of \tabref{tab:review:datasets} should be read as follows. Each dimension within the label set is allocated its own line. Within a single dimension, the possible values can be one label, a set of multi-class labels, or a set of multi-label labels. To differentiate the types of dimensions, we used the respective index with the set of labels in parentheses ((LABEL1, LABEL2, ...)\textsubscript{DIMENSION TYPE}). For dimensions consisting of a singular, binary label, we added a ``binary'' index to the dimension. As far as the specific values of the labels within a dimension are concerned, we assume binary labels by default and mark different label spaces accordingly.
    
\subsubsection{Annotators and Annotation Process}
\label{sec:review:results:datasets:annotation}
The annotation processes of the datasets were documented with varying degrees of transparency and quality. Generally, there are two primary annotation approaches. The samples are either given to domain experts, or crowdsourcing platforms are leveraged to recruit annotators. The creators of datasets that were annotated through crowdsourcing used the platforms ``CrowdFlower'' \cite{waseem_are_2016,davidson_automated_2017,silberztein_automatic_2018,founta_large_2018,wulczyn_ex_2017}, ``Figure Eight'' \cite{cjadams_jigsaw_2019,basile_semeval-2019_2019,zampieri_predicting_2019,mollas_ethos_2022}, ``Amazon Mechanical Turk'' \cite{ousidhoum_multilingual_2019,mathew_hatexplain_2021}, or ``Prolific'' \cite{arampatzis_overview_2023}. We could not recognize an established approach for the use of crowd workers. The procedure varied regarding the annotators' required language skills, an evaluation of their comprehension of relevant concepts, the extent of explanations and instructions given to them, and their financial compensation. The number of judgments per sample in the crowdsourced datasets is mostly around three, and a majority vote often determines the final label (see, \eg, \cite{founta_large_2018,davidson_automated_2017,silberztein_automatic_2018,mathew_hatexplain_2021}). To ensure the quality of crowdsourced labels, some researchers create a significantly smaller subset of ``gold standard labels'' that are annotated by experts and then used to validate the other annotators, \eg, by ignoring all judgments of a specific annotator if they fall short of a certain accuracy-threshold on these labels \cite{silberztein_automatic_2018,rottger_hatecheck_2021,mathew_hatexplain_2021,wulczyn_ex_2017}.
    
For the datasets with expert annotations, the specific expertise type varies. While some researchers annotated the data themselves \cite{waseem_hateful_2016,wiegand_overview_2018,cheriyan_towards_2021,alshamrani_hate_2021}, others involved staff of \ac{ngos} that deal with the type of hate speech of interest \cite{chung_conan_2019,waseem_are_2016}. Others relied on individuals with a formal academic education in a thematically relevant field of research \cite{parikh_multi-label_2019,mandl_overview_2019}, or a group of individuals that received specialized training beforehand \cite{vidgen_learning_2021,rottger_hatecheck_2021,kennedy_introducing_2022}. For one dataset, the type of experts was not specified \cite{fersini_overview_2018}. For another, crowdsourced and expert annotations were combined. Specifically, \citet{basile_semeval-2019_2019} had each sample annotated once with an aggregated crowd label and twice by separate experts, with the final label being determined by a majority vote.

A different approach for the creation of annotated data is the generation of artificial samples that fit specific labels. This can be useful to ensure an even distribution of different classes/labels in the training data. The authors of the CONAN dataset \cite{chung_conan_2019}, \eg, asked 111 annotators from three \ac{ngos} to create ``typical'' examples of hate speech they would encounter during their work. \citet{fanton_human---loop_2021} expanded this approach by adding an augmentation step in which they generated instances of hate speech with an LLM (GPT-2) that were then validated or corrected by human annotators. \citet{vidgen_learning_2021} applied another approach to generate artificial samples. First, they trained a hate speech classification model on previously published datasets. Then they tasked human annotators to create new samples that ``trick'' the model using ``adversarial pivots'', \ie, minimal adjustments of a given sample that should change the assigned label. The annotators, therefore, simultaneously created edge cases and imitated detection-avoidant behavior of internet users trying to circumvent an automated detection system.

\subsubsection{Inter-Annotator Agreement}
\label{sec:review:results:datasets:iaa}
Information on concrete \ac{iaa} metrics and their respective values is only provided for 13 of the 28 analyzed datasets. Another five only indicated \ac{iaa} values without specifying the corresponding metric. This lack of transparency is problematic since it complicates the assessment of annotation quality. For instance, \citet{fersini_overview_2018} report an \ac{iaa} of 0.81 for their ``misogynous'' label, but the underlying evaluation metric is not given. For other datasets, only a ``disagreement percentage'' is reported \cite{founta_large_2018,bhattacharya_developing_2020,zampieri_predicting_2019,mandl_overview_2020}, which refers to the share of instances with self-defined disagreement among all samples. Finally, one dataset was only annotated by one person \cite{alshamrani_hate_2021}, and for another, samples with disagreement between its two annotators were dismissed \cite{gupta_hateful_2023}. Thus, no \ac{iaa} was calculated for them. The metrics used to calculate the \ac{iaa} were \textit{Krippendorff's alpha} \cite{ousidhoum_multilingual_2019,vidgen_learning_2021,mathew_hatexplain_2021,wulczyn_ex_2017}, \textit{Cohen's Kappa} \cite{chung_conan_2019,parikh_multi-label_2019,cheriyan_towards_2021}, \textit{Fleiss' Kappa} \cite{rottger_hatecheck_2021,mollas_ethos_2022,kennedy_introducing_2022}, \textit{F8 reported average confidence} \cite{basile_semeval-2019_2019}, \textit{\ac{pabak}} \cite{kennedy_introducing_2022}, and \textit{\ac{pabakos}} \cite{silberztein_automatic_2018}. We cannot name the most common metric because several were used for three or four datasets.
    
\subsubsection{Availability}
\label{sec:review:results:datasets:availability}
16 of the 28 datasets were publicly accessible online at the time of our analysis. Three were, as far as we can establish, never published while another is either no longer online or subject to link-rot. The remaining eight were all part of a shared task. Access to them is locked behind some type of registration, application, or authentication process. While access might still be possible, we did not investigate their availability further.
    
{\small
\begin{longtable}{p{0.5cm}p{2.2cm}p{1.5cm}p{0.5cm}p{2.5cm}p{3.5cm}p{0.6cm}}
\caption{Surveyed machine learning models for multi-label hate speech classification with information on their multi-label approach, training dataset(s), evaluation metrics, and relation to \ac{mt} Learning.}  \label{tab:review:model_types}  \\
\hline

\textbf{Ref.} & \textbf{Model Type} & \textbf{Approach} & \textbf{MT} & \textbf{Dataset} & \textbf{Evaluation Metrics} & \textbf{Year} \\
\hline
\endfirsthead
\multicolumn{7}{c}{\tablename\ \thetable\ -- \textit{continued from previous page}} \\
\hline
\textbf{Ref.} & \textbf{Model Type} & \textbf{Approach} & \textbf{MT} & \textbf{Dataset} & \textbf{Evaluation Metrics} & \textbf{Year} \\
\hline
\endhead
\hline \multicolumn{7}{r}{\textit{Continued on next page}} \\
\endfoot
\hline
\endlastfoot

\cite{saeed_overlapping_2018} & 
CNN, Bi-LSTM, Bi-GRU & 
DML & & 
TCCC & 
(P, R, F1, ROC-AUC)\textsubscript{macro}, SA & 
2018 \\
\hline
\cite{ibrahim_imbalanced_2018} &
Ensemble: CNN + LSTM + Bi-GRU &
DML & &
Kaggle+ &
F1\textsubscript{micro}, F1\textsubscript{exam} &
2018 \\
\hline
\cite{van_aken_challenges_2018} &
Bi-LSTM, Bi-GRU, LR, CNN, Ensemble (of all 4) &
DML & &
TCCC; Davidson et al. (2017) &
(P, R, F1, ROC-AUC)\textsubscript{macro} &
2018 \\
\hline
\cite{gunasekara_review_2018} &
LGBM (Ensemble), Bi-LSTM &
DML & &
Wikipedia Comment Corpus &
ROC-AUC\textsubscript{?} &
2018 \\
\hline
\cite{del_ser_design_2018} &
CapsNet, Bi-GRU, LR &
DML & &
TCCC &
(ER, F1, ROC-AUC)\textsubscript{label}, ROC-AUC\textsubscript{micro}, ROC-AUC\textsubscript{macro} &
2018 \\
\hline
\cite{ratadiya_attention_2019} &
Multi-headed Self Attention &
DML & &
TCCC &
(P, R, F1, ROC-AUC)\textsubscript{macro} &
2019 \\
\hline
\cite{liu_fuzzy_2019} &
Semi-Supervised Multi-Task Fuzzy Ensemble &
BR & \hspace{0.07cm}\cmark &
Burnap and Williams (2015) &
(R, IR)\textsubscript{label} &
2019 \\
\hline
\cite{parikh_multi-label_2019} &
CNN, Bi-LSTM, Attention &
DML & &
Parikh et al. (2019) &
(F1, Acc)\textsubscript{exam}, F1\textsubscript{micro}, F1\textsubscript{macro} &
2019 \\
\hline
\cite{ousidhoum_multilingual_2019} &
LR, Sluice Network + Bi-LSTM &
DML & \hspace{0.07cm}\cmark &
Ousidhoum et al. (2019) &
F1\textsubscript{micro}, F1\textsubscript{macro} &
2019 \\
\hline
\cite{ozler_fine-tuning_2020} &
BERT &
DML, BR & &
Local News Comments; Local politics Tweets; TCCC &
F1\textsubscript{label}, ROC-AUC\textsubscript{?} &
2020 \\
\hline
\cite{faal_domain_2021} &
BERT &
DML & \hspace{0.07cm}\cmark &
Unintended Bias in Toxicity Classification &
Subgroup AUC, BPSN-AUC, GMB-AUC &
2021 \\
\hline
\cite{husnain_novel_2021} &
LR, Naive Bayes, CART, Decision Tree, Multinomial Naive Bayes &
BR, LP & &
TCCC &
Acc &
2021 \\
\hline
\cite{zhao_comparative_2021} &
BERT, RoBERTa, XLM &
DML & &
TCCC &
F1\textsubscript{micro}, F1\textsubscript{macro} &
2021 \\
\hline
\cite{alshamrani_hate_2021} &
Fully-connected NN &
BR & &
Alshamrani et al. (2021) &
(P, R, F1)\textsubscript{label} &
2021 \\
\hline
\cite{parikh_categorizing_2021} &
Bi-LSTM w/ attention &
DML, BR, LP & &
Parikh et al. (2019) &
Acc, F1\textsubscript{w}, (P, R, F1, Acc)\textsubscript{exam}, F1\textsubscript{micro}, F1\textsubscript{macro} &
2021 \\
\hline
\cite{abburi_fine-grained_2021} &
Bi-LSTM w/ attention on top of finetuned BERT &
DML & &
Parikh et al. (2019) &
(Acc, F1)\textsubscript{exam}, F1\textsubscript{micro}, F1\textsubscript{macro}, SA &
2021 \\
\hline
\cite{mishra_exploring_2021} &
BERT &
LP & \hspace{0.07cm}\cmark &
HASOC2019 &
F1\textsubscript{w}, F1\textsubscript{macro} &
2021 \\
\hline
\cite{dirting_multi-label_2022} &
BERT &
DML & &
TCCC &
(P, R, F1)\textsubscript{?} &
2022 \\
\hline
\cite{aditya_classifying_2022} &
LR, Multi-Label kNN &
BR, LP & &
HateXplain &
(R, Acc)\textsubscript{micro}, R\textsubscript{macro}, HL &
2022 \\
\hline
\cite{mollas_ethos_2022} &
Bi-LSTMs &
BR & &
ETHOS &
(P, R, F1)\textsubscript{exam}, (P, R, F1, AP)\textsubscript{micro}, (P, R, F1, AP)\textsubscript{macro}, HL, SA &
2022 \\
\hline
\cite{bhattacharjee_sentiment_2023} &
LSTM, CNN, LSTM + CNN &
DML & &
UC Berkeley HS corpus &
(Acc, P, R, F1)\textsubscript{?} &
2023 \\
\hline
\cite{gupta_hateful_2023} &
BERT &
DML & &
HateComments &
(P, R, F1, ROC-AUC)\textsubscript{label} &
2023 \\
\hline
\cite{mazari_bert-based_2024} &
Ensemble: BERT + Bi-LSTM + Bi-GRU &
DML & &
TCCC &
(P, R, F1, MCC)\textsubscript{label}, ROC-AUC\textsubscript{?} &
2024 \\
\hline
\cite{abburi_multi-task_2024} &
Bi-LSTM + Attention &
DML & \hspace{0.07cm}\cmark &
Parikh et al. (2019) &
(Acc, F1)\textsubscript{exam}, F1\textsubscript{micro}, F1\textsubscript{macro}, SA &
2024 \\
\hline

\end{longtable}
\raggedright
\footnotesize Abbreviations: \acs{acc} = \acl{acc}; \acs{ap} = \acl{ap}; \acs{auc} = \acl{auc}; \acs{bert} = \acl{bert}; \acs{bigrus} = \acl{bigrus}; \acs{bilstm} = \acl{bilstm}; \acs{br} = \acl{br}; \acs{bpsn-auc} = \acl{bpsn-auc}; \acs{cart} = \acl{cart}; \acs{cnn} = \acl{cnn}; \acs{capsnet} = \acl{capsnet}; \acs{dml} = \acl{dml}; \acs{er} = \acl{er}; \acs{f1} = \acl{f1}; \acs{gmb-auc} = \acl{gmb-auc}; \acs{hl} = \acl{hl}; \acs{ir} = \acl{ir}; \acs{knn} = \acl{knn}; \acs{lgbm} = \acl{lgbm}; \acs{lp} = \acl{lp}; \acs{lr} = \acl{lr}; \acs{lstm} = \acl{lstm}; \acs{mcc} = \acl{mcc}; \acs{mt} = \acl{mt}; \acs{nn} = \acl{nn}; \acs{p} = \acl{p}; \acs{r} = \acl{r}; \acs{roberta} = \acl{roberta}; \acs{roc-auc} = \acl{roc-auc}; \acs{sa} = \acl{sa}; \acs{tccc} = \acl{tccc}; \acs{xlm} = \acl{xlm}.
}

\subsection{Machine Learning Models}
\label{sec:review:results:models}
In total, we identified 24 papers that present models for multi-label hate speech classification. \tabref{tab:review:model_types} provides a concise overview. In the following, we compare them with regard to the adopted model types and multi-label approaches, the implementation of multi-task learning, and applied evaluation metrics.

\subsubsection{Model Type(s)}
\label{sec:review:results:models:types}
Various approaches have been employed for the multi-label classification of textual hate speech in English. Despite this variance, we identified some similarities between the works. For instance, nine works \citep{ratadiya_attention_2019, ozler_fine-tuning_2020, faal_domain_2021, zhao_comparative_2021, abburi_fine-grained_2021, mishra_exploring_2021, dirting_multi-label_2022, mazari_bert-based_2024, gupta_hateful_2023} leverage transformer models like \textit{\ac{bert}} as the foundation of their models. The widespread use of \ac{bert} might be attributed to its performance in \ac{nlp} tasks, which stems from its implementation of the attention mechanism \citep{vaswani_attention_2017} for the creation of sophisticated text representations (embeddings). This allows the network to consider contextual clues in the text on a global scope, which is particularly beneficial in the context of hate speech classification. Furthermore, the model is pre-trained on a massive amount of data. Therefore, a lot of general knowledge about patterns and concepts in natural language is already encoded in \ac{bert}'s base model. When facing a specific \ac{nlp} task, such as hate speech classification, a small number of examples is typically sufficient to fine-tune the model to the target domain and achieve a good performance. However, as \ac{bert} has millions of parameters, one disadvantage is that it requires some resources to run~\cite{devlin_bert_2019}.
    
In our survey we identified ten works \citep{saeed_overlapping_2018, van_aken_challenges_2018, gunasekara_review_2018, parikh_multi-label_2019, ousidhoum_multilingual_2019, parikh_categorizing_2021, abburi_fine-grained_2021, mollas_ethos_2022, mazari_bert-based_2024, abburi_multi-task_2024} using \textit{\ac{bilstm}} \cite{zhang_bidirectional_2015}. The \ac{bilstm} architecture is designed to capture sequential dependencies, which can also be beneficial in capturing contextual data. The bi-directionality enables the network to consider information not only from the past, \ie, preceding words or tokens, but also information from the future, \ie, subsequent words or tokens. Even though the \ac{bilstm} is capable of capturing both directions, it only does so in its local context. The global context cannot be captured by the network to the same extent as with \ac{bert} \cite{vaswani_attention_2017,hochreiter_vanishing_1998}. This is especially problematic for longer documents, where the local context does not match the global context. Furthermore, due to its sequential nature, there is only a limited capability to process the data in parallel. This causes the training to be rather exhaustive and time-consuming for large data volumes \cite{vaswani_attention_2017}.
    
In five works \cite{saeed_overlapping_2018, ibrahim_imbalanced_2018, van_aken_challenges_2018, del_ser_design_2018, mazari_bert-based_2024}, \textit{\ac{bigrus}} were used for classification. \ac{bigrus} are similar to \ac{bilstm}s but occupy a simpler architecture. 
Due to the more simplistic gating mechanism, they can be trained substantially faster. While the representations may not be as rich as \ac{bilstm}s, they often achieve similar performance \cite{chung_empirical_2014}. These networks are primarily designed to capture short-term dependencies. In the context of our survey, several researchers used a combination of \ac{bigrus} and \acs{lstm}s.
    
Several researchers have not used individual models but ensemble methods for classification \citep{ibrahim_imbalanced_2018, van_aken_challenges_2018, liu_fuzzy_2019, mazari_bert-based_2024,gunasekara_review_2018}. For instance, \citet{gunasekara_review_2018} used a \textit{\ac{lgbm}}, a type of gradient boosting framework, in which several weak learners are trained sequentially and the subsequent learner makes up for the mistakes of its previous ones \cite{ke_lightgbm_2017}. This approach is rather efficient as it focuses on scalability and speed. 
Beyond that, other researchers have chosen different ensemble methods \cite{ibrahim_imbalanced_2018,liu_fuzzy_2019,mazari_bert-based_2024,van_aken_challenges_2018}, each with a distinct approach to model combination (see \tabref{tab:review:model_types}).

\begin{figure}
    \centering
    \includegraphics[scale=0.55]{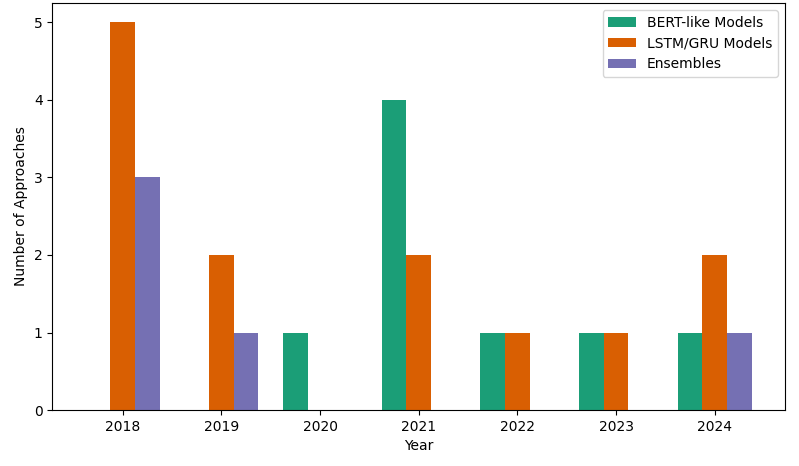}
        \caption{Distribution of model types by year.}
\label{fig:review:model_types_year}
\end{figure}

A temporal perspective on model type selection, as shown in \figref{fig:review:model_types_year}, reveals a common pattern in \ac{nlp} research. Even though overall more researchers opted for \ac{bilstm}s, many researchers have instead chosen transformers since their release. In the surveyed works, \ac{bilstm}s were often combined with other model types to reach a good performance \citep{ibrahim_imbalanced_2018, van_aken_challenges_2018, ousidhoum_multilingual_2019,parikh_categorizing_2021, abburi_fine-grained_2021, mazari_bert-based_2024, abburi_multi-task_2024, bhattacharjee_sentiment_2023}. By contrast, \ac{bert} achieved a strong performance in multi-label hate speech classification as a standalone model, which further substantiates its outstanding position in the \ac{nlp} field. Unfortunately, no surveyed works have used large foundation models such as GPT-4 or Llama 3 for multi-label hate speech classification. As these models constitute a new state of the art for many \ac{nlp} tasks and may be particularly useful in multi-label scenarios \cite{brown_language_2020, peskine_definitions_2023}, we discuss them in \secref{sec:review:discussion:recommendations}.
    
\subsubsection{Multi-Label Approach}
In \tabref{tab:review:model_types} we show how the multi-label classification task was approached in the different works. \textit{Direct multi-label classification (\acs{dml})} refers to addressing the multi-label classification problem directly, without converting it into one or more single label classification tasks (18 works).
In contrast, \textit{\acl{br} (\acs{br})} approaches (seven works) transform the multi-label problem into multiple binary problems, while \textit{\acl{lp} (\acs{lp})} classification converts it into a single multi-class task (four works). We provide detailed descriptions of \acs{br} and \acs{lp} for multi-label classification in \secref{sec:review:background:multilabel}. 
A commonality across all \acs{dml} approaches is that some base model, \eg, \ac{bilstm}, \ac{bigrus}, \ac{bert}, gets expanded by an output layer whose dimension equals the number of target labels in the dataset. The most prominent activation function for this output layer is the sigmoid function, as it outputs a value between 0 and 1 for each label that can be interpreted as the probability of the respective label given the input sample. It should be noted that softmax in particular is not being used, as softmax normalizes the probability across all labels to sum to one so that one label at most can have a high probability. This renders it ideal for multi-class classification but inherently unsuitable for multi-label tasks.

\subsubsection{Multi-Task Learning}
In five of the 24 surveyed articles that proposed model architectures, multi-label hate speech classification has been explored in the context of \textit{\acl{mt} (\acs{mt}) Learning}. \citet[p. 31]{zhang_overview_2018} have defined the \acs{mt} learning paradigm as follows: ``Given $T_{1...m}$ related but not identical learning tasks, multi-task learning aims to improve the learning of the model for a task $T_i$ by using the knowledge contained in all of the $m$ tasks.''

While a variety of techniques has been used in the literature to address the challenges faced in hate speech classification (see \secref{sec:review:discussion:open_issues}), we put special emphasis on multi-task learning as it has not only been applied to mitigate general limitations in hate speech classification but also to enhance hate speech classification in multi-label settings specifically. One advantage of \acs{mt} Learning is that it allows us to leverage multiple binary datasets, each tailored to a specific type of hate speech (\eg, sexism, racism) for multi-label classification similar to the \acs{br} approach. However, unlike for \acs{br}, where a separate binary classifier is trained for each label, in \acs{mt} Learning, a single multi-label model is trained on multiple datasets collaboratively, leveraging shared information across tasks. Following this concept, \citet{liu_fuzzy_2019} propose a \acs{mt} framework for training a multi-label classifier using multiple single-label datasets, while \citet{ousidhoum_multilingual_2019} train a unified model across multiple multi-label tasks, leveraging sluice networks \cite{ruder_latent_2019} to learn shared and task-specific representations.

\citet{abburi_multi-task_2024} use \acs{mt} Learning to address the issue of data sparsity by providing complementary signals during learning that extend beyond the information available in a single task. In particular, they train a model not only on the hate speech classification task at hand but also on additional homogeneous (intra-domain) and heterogeneous (cross-domain) auxiliary tasks to enhance its multi-label classification performance. They create their homogeneous auxiliary tasks through techniques that do not require manual labeling, such as unsupervised learning and weak labeling, so that they can utilize unlabeled data during training. Their heterogeneous tasks include the identification of sarcasm and emotional cues, thereby enhancing the classifier's capacity to make better-informed decisions.

In a similar fashion, \citet{faal_domain_2021} used multi-label identity group classifications as auxiliary tasks to improve their binary classifier. This was done by using the identity clues to mitigate unintended bias in their model's prediction and resulted in an improvement in the model's generalization capabilities. Additionally, they trained a model consisting of domain adaptive \ac{bert} layers that are shared by all tasks, followed by task-specific layers that produce outputs for each task separately.
Finally, \citet{mishra_exploring_2021} transformed the HASOC 2019 challenge, which consists of three binary and multi-class sub-tasks, into a single multi-label task by training on all three tasks simultaneously. They train a shared model with task-specific output heads and report that their \acs{mt} model is computationally efficient. 
In summary, rationales for the use of \acs{mt} Learning are related to issues of data sparsity, unintended bias, computational costs, and poor generalization across related tasks.
    
\subsubsection{Evaluation Metrics}
\label{sec:review:results:models:evaluation}
\begin{figure}
    \centering
    \includegraphics[scale=0.8]{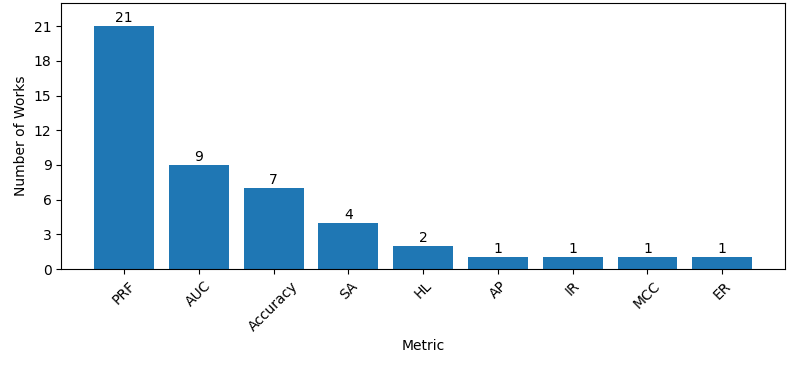}
        \caption{Frequency of metric types used for evaluating the surveyed multi-label hate speech classification models. ``PRF'' includes any form of \acs{p}, \acs{r}, or \acs{f1}. ``AUC'' includes all mentioned variants (\eg, \acs{roc-auc}, Subgroup \acs{auc}, \acs{bpsn-auc}).}
\label{fig:review:metrics}
\end{figure}

As can be seen in \figref{fig:review:metrics}, the metrics used for the evaluation of hate speech models vary considerably.
For our analysis, we follow the topology for multi-label evaluation metrics by \citet{zhang_review_2014}. According to them, multi-label evaluation metrics can generally be seen as either example-based or label-based.

\textit{Example-based:} 
Example-based metrics evaluate the model's performance on each sample (\ie, each labeled text fragment) separately and then take the mean value to weight all examples equally. One popular example-based metric is the \textit{\acl{sa} (\acs{sa})}. \acs{sa} is similar to the binary \acl{acc} measure in the sense that it measures the proportion of correctly classified examples. However, when evaluating \acs{sa}, an example only counts as correctly classified when the model predicted exactly the label-subset associated with the example. Hence, \acs{sa} is quite a strict metric, especially for large label spaces, since predicting all of an example's correct labels except one is still counted as a misclassification \cite{zhang_review_2014}.
Another example-based metric, which is used by \citet{mollas_ethos_2022}, is the \textit{\acl{hl} (\acs{hl})}. For each example, it evaluates the fraction of misclassified labels, \ie, missing relevant labels or predicted irrelevant labels, and averages this value over all instances \cite{zhang_review_2014}. \citet{del_ser_design_2018} used the \acs{hl} metric to evaluate a binary classifier. Here the \acs{hl} simplifies to the \textit{\acl{er} (\acs{er})}, which represents the fraction of misclassified samples.
The most frequently used metrics for multi-label hate speech classification, however, are adaptations of common binary evaluation metrics, such as \textit{\acl{p} (\acs{p})}, \textit{\acl{r} (\acs{r})}, \textit{\acl{f1} (\acs{f1})}, and \textit{\acl{acc} (\acs{acc})}. There are multiple ways to render them applicable to multi-label tasks. The example-based approach is to first evaluate these metrics per example and then calculate the mean value of all instance scores. For instance, example-wise \acs{p} for $p$ examples with corresponding label-sets $Y_p$ and a classifier model $h(x)$ becomes:
    \begin{align}
        P_{exam}(h) = \frac{1}{p} \sum_{i=1}^{p} \frac{|Y_i \cap h(x_i)|}{|h(x_i)|}
    \end{align}
\acs{r}, \acs{f1}, and \acs{acc} can be defined analogously \cite{zhang_review_2014}. Whenever binary evaluation metrics have been used example-wise for multi-label classification, we indexed them with \textit{exam} in \tabref{tab:review:model_types}.
    
\textit{Label-based:}
In contrast to example-based metrics, label-based metrics first evaluate the model on each label separately and then average the label-wise scores across all class labels \cite{zhang_review_2014}. As a model's predictions with regard to a single label can be seen as a binary classification task, binary evaluation metrics are applied. In addition to \acs{p}, \acs{r}, and \acs{f1}, the \textit{\acl{roc-auc} (\acs{roc-auc})} metric is reported in several works. While applying these metrics label-wise is straightforward, there are different approaches to average them across all class labels.
The first approach is micro-averaging. For this, the label-wise numbers of \ac{tp}, \ac{tn}, \ac{fp}, and \ac{fn} are determined first. Then, they are added up for all labels before the respective binary evaluation metric is calculated. With $B$ denoting the respective metric and $q$ denoting the number of labels, \citet{zhang_review_2014} have formulated this as follows:
    \begin{align}
        B_{micro} = B(\sum_{j=1}^q TP_j, \sum_{j=1}^q TN_j, \sum_{j=1}^q FP_j, \sum_{j=1}^q FN_j)
    \end{align}
Macro-averaging is the other popular, and most trivial, averaging method. Here, the label-wise metric scores are evaluated first, and then the mean value across all labels is calculated \cite{zhang_overview_2018}. For a metric $B$ and $q$ labels, macro-averaging can be formulated as follows:
    \begin{align}
        B_{macro} = \frac{1}{q}\sum_{j=1}^q B_j
    \end{align}
The main difference between both methods is that macro averaging weighs all labels equally, while micro averaging weighs examples equally \cite{zhang_review_2014}. This results in different scores for both metrics when classes are unbalanced. In \tabref{tab:review:model_types}, we indexed label-wise binary metrics with either \textit{micro} or \textit{macro}, according to the chosen method.
If scholars have not aggregated the label-wise binary evaluation metrics but instead reported them separately for each label, we indexed with \textit{label}. In addition to binary metrics discussed above, \citet{mazari_bert-based_2024} reported the label-wise \textit{\acl{mcc} (\acs{mcc})}, stating that it can be used for accurate evaluation even if there is a significant class imbalance in the test set.

\textit{Multi-class Evaluation:}
In two works that utilized \ac{lp} to transform the multi-label into a multi-class task, we also observed the use of metrics for multi-class classification \cite{mishra_exploring_2021, parikh_categorizing_2021}. Both works calculated the \textit{\acs{f1}\textsubscript{macro} score}, which can also be used for multi-label evaluation, and the \textit{support-weighted \acs{f1} score (\acs{f1}\textsubscript{w})}. While the \acs{f1}\textsubscript{macro} score assigns equal weights to all classes when averaging the class-wise \acs{f1} scores, \acs{f1}\textsubscript{w} weights each class by the proportion of instances belonging to it.

\textit{Uncommon metrics:}
Some metrics were only employed in individual works. \citet{mollas_ethos_2022} reported the micro and macro aggregated \textit{\acl{ap} (\acs{ap})}, which is the label-wise area under the Precision-Recall curve. Furthermore, \citet{liu_fuzzy_2019} report a label-wise Detection Rate and a label-wise \textit{\acl{ir} (\acs{ir})}. They characterize both metrics as novel. However, following their definitions, the Detection Rate is calculated analogously to label-wise \acl{r}. Thus, in \tabref{tab:review:model_types} we indicated it as \acs{r}\textsubscript{label}. Label-wise \acs{ir} does not correspond to any common evaluation metric. They define it as the number of instances to which the classifier assigned a label l (\acs{tp} + \acs{fp}), divided by the number of negative instances not associated with the label l (\acs{tn} + \acs{fp}). Unfortunately, it remains unclear how this metric conveys a quantification of a model's prediction quality.
The metrics used by \citet{faal_domain_2021}, \ie, \textit{Subgroup \ac{auc}}, \textit{\ac{bpsn-auc}}, and \textit{\ac{gmb-auc}} are distinctive, since they do not measure a model's classification performance but its ability to mitigate unintended bias. The researchers did not evaluate their multi-label classifier with other performance metrics, since in their \ac{mt} setting, the multi-label task's purpose was to reduce bias in their binary classification task.

Whenever authors reported binary evaluation metrics for multi-label classifiers without any specification of how they were made applicable to such a setting, we indexed the respective metrics with a question mark (?) in \tabref{tab:review:model_types}. For \acs{roc-auc} it can be assumed that authors refer to macro-averaged \acs{roc-auc}, since it was the official metric used for Kaggle's \ac{tccc} \cite{cjadams_toxic_2017}. Nevertheless, the ambiguous documentation of evaluation metrics represents a deficit in parts of the research landscape. In addition, it should be emphasized that using (macro-)accuracy as a performance measure can lead to false inferences because it favors the majority class \cite{ratadiya_attention_2019}. Since it can also be biased in case of unbalanced data \cite{ibrahim_imbalanced_2018}, we recommend the use of other metrics.

Altogether, three conclusions can be drawn on basis of our examination of the evaluation metrics used for multi-label hate speech classification. 
\textit{First}, some authors use metrics that by themselves are not well-defined in the multi-label setting while not elaborating on them in their works (see, \eg, \cite{liu_fuzzy_2019,gunasekara_review_2018,ozler_fine-tuning_2020,dirting_multi-label_2022,bhattacharjee_sentiment_2023,mazari_bert-based_2024}). This complicates the comprehension of their metrics' calculation and the interpretation of their results.
\textit{Second}, there are no universally recognized evaluation metrics for multi-label hate speech classification yet. This makes the comparison of available models more difficult.
Whereas \acs{roc-auc} and \acs{f1} score are most frequently used (see \figref{fig:review:metrics}), there are different methods to apply such binary measures to multi-label tasks. In case of the \acs{f1} score, at least \acs{f1}\textsubscript{micro} and \acs{f1}\textsubscript{macro} should be reported to allow for an easy comparison.
\textit{Third}, given that class imbalance is a prevalent issue in several hate speech datasets \cite{ibrahim_imbalanced_2018,alkomah_literature_2022,mazari_bert-based_2024,del_ser_design_2018}, evaluation metrics should account for this issue. \citet{jeni_facing_2013} found that among the metrics they calculated, only \acs{roc-auc} remained unaffected by imbalanced datasets. \citet{mazari_bert-based_2024} used the \acs{mcc} to address this issue. \acs{mcc} is particularly suited for unbalanced datasets since it is invariant to the proportion of positive and negative samples. As this ensures that neither class is favored, it has been described as more informative, truthful, and reliable than comparable metrics \cite{chicco_matthews_2021}. However, both \acs{roc-auc} and \acs{mcc} are binary metrics that require adaptation for multi-label classification using one of the above-mentioned approaches. We recommend complementing label-wise metrics with holistic metrics that evaluate the model's overall performance, which allows for easier inter-model comparisons.

\subsubsection{Further Resources}
While \ac{bert} is generally one of the most frequently used models for multi-label hate speech classification, some researchers further trained it with domain-specific data. Even though \citet{caselli_hatebert_2021} do not evaluate their HATEBERT model for multi-label tasks, they show that this  \ac{bert} model specifically trained for hate speech classification outperforms regular \ac{bert} in binary and multi-class tasks. HATEBERT's domain-adapted embeddings may also outperform the standard model in multi-label settings. However, this expectation requires empirical validation.
\section{Discussion}
\label{sec:review:discussion}
In this first systematic literature survey on textual multi-label hate speech classification in English, we contributed with a comprehensive and comparative overview of both available datasets (C1) and \ac{ml} models (C2). During our analysis, we observed several challenges faced by the research community. However, since we only implicitly engaged with them in \secref{sec:review:results}, we will present and discuss the four most prevalent open issues (I1-I4) in research as our third contribution in \secref{sec:review:discussion:open_issues} (C3). We will first engage with dataset-related issues, discussing how choices regarding labeling process, model architecture, or training procedure may provide mitigation, and then examine several methodological aspects that are indicative of a missing alignment of scholars in the surveyed field of research. In response to these issues and as our last contribution, we formulate ten recommendations (R1-R10) for future research in \secref{sec:review:discussion:recommendations} (C4). We conclude the discussion by reflecting on this work's limitations and avenues for future research in \secref{sec:review:discussion:limitations}.

\subsection{Open Issues in Multi-label Hate Speech Classification}
\label{sec:review:discussion:open_issues}
As regards training datasets for textual multi-label hate speech classification, one central issue is related to \textit{class imbalance (I1)} \cite{ibrahim_imbalanced_2018,alkomah_literature_2022,mazari_bert-based_2024,del_ser_design_2018}. Given that most social media content is non-hateful, datasets constructed through content sampling typically exhibit a disproportionate level of representation. Benign samples significantly outnumber instances of hate speech, even when sampling techniques were deliberately designed to capture hateful content, \eg, by searching tweets that use certain hashtags \cite{waseem_hateful_2016,kennedy_introducing_2022}. While such biased sampling strategies may reduce class imbalances, they can introduce other distortions to the data. For instance, when using keyword-lists to pre-select hateful samples, the resulting sample set is destined to over-represent these keywords. In the surveyed works, little emphasis is put on mitigating such biases and only few researchers discuss this issue explicitly (see \eg, \cite{kennedy_introducing_2022}). Datasets with multi-label annotations often exhibit not only imbalances regarding the distribution of non-hateful and hateful samples but also the different hateful classes \cite{parikh_multi-label_2019}. In response, data augmentation techniques have been used to generate additional samples for the underrepresented classes. This can either be achieved through the algorithmic alteration of available samples \cite{ibrahim_imbalanced_2018} or the generation of entirely synthetic samples \cite{rottger_hatecheck_2021}, allowing for an explicit control of class proportions. As proposed by \citet{vidgen_challenges_2019}, both methods can also be combined in a human-in-the-loop setup.

The common \textit{reliance on crowdsourcing platforms (I2)} for data annotation, as described in \secref{sec:review:results:datasets:annotation}, is another issue. While these platforms allow for enhanced scalability and a (potentially) more diverse annotator pool \cite{founta_large_2018}, researchers transfer significant control over annotation quality to crowd workers \cite{kumar_multilingual_2023}. Given this trade-off, researchers must take a case-by-case decision in consideration of available resources and the specific use case of datasets. When relying on crowdsourcing, a lack of access and control as well as enhanced annotator diversity increase the possibility that the label set is either incoherently understood by all annotators or inconsistently applied \cite{price_six_2020}. This complicates the assurance of a high annotation quality. However, researchers can implement measures to enhance quality at three stages of a crowdsourced annotation process.
First, they can focus on ensuring consensus on and comprehension of label sets before initiating the process. This may be done by limiting the pool of viable annotators, \eg, by only involving individuals with specific language skills, expertise, or cultural background. This may be checked by some kind of ``pre-test'' \cite{ousidhoum_multilingual_2019}. Beyond that, the provision of detailed and easily comprehensible annotation guidelines and label definitions can be beneficial.
Second, researchers can filter annotators during the annotation process by their performance. This is typically realized by creating a subset of gold standard samples that is then used to check individual annotation quality \cite{silberztein_automatic_2018,rottger_hatecheck_2021,mathew_hatexplain_2021,wulczyn_ex_2017}. Some crowdsourcing platforms even provide features to support this \cite{basile_semeval-2019_2019}.
Finally, researchers can improve annotation quality after the process by combining multiple individual judgments (see, \eg, \cite{founta_large_2018}). In the examined datasets, authors mostly applied simple majority voting among three to five annotators.

The availability of mostly \textit{small and/or sparse datasets (I3)}, which already limits the performance of binary hate speech classification models \cite{alkomah_literature_2022,subramanian_survey_2023}, constitutes an even more significant issue for multi-label hate speech classification. This is due to the much larger output space of multi-label problems, which increases exponentially with the number of labels \cite{zhang_review_2014}. To handle such an extensive output space, multi-label \ac{ml} models not only have to learn how to discriminate between different labels but also the dependencies among them. In the surveyed works, researchers addressed the issue of small and/or sparse datasets with different design choices regarding model architecture and training procedure.
First, it can be mitigated by leveraging \textit{problem transformation}, using either \ac{br} (see \cite{aditya_classifying_2022, alshamrani_hate_2021, husnain_novel_2021, liu_fuzzy_2019, mollas_ethos_2022, ozler_fine-tuning_2020, parikh_categorizing_2021}) or \ac{lp} (see \cite{parikh_categorizing_2021, mishra_exploring_2021, husnain_novel_2021, aditya_classifying_2022}). Researchers can for example use \ac{br} to decompose the multi-label problem into multiple binary tasks to ensure dense training data classifier-wise. Additionally, one is not restricted to the still limited amount of multi-label datasets but can also utilize binary datasets that match labels of the original multi-label task in a multi-domain fashion.
Second, researchers can follow the \textit{\ac{mt} Learning} paradigm by training models on more than one dataset (see, \eg, \cite{liu_fuzzy_2019,ousidhoum_multilingual_2019,faal_domain_2021,mishra_exploring_2021,abburi_multi-task_2024}).
Third, researchers can \textit{finetune pre-trained \ac{llms}}. As described in \secref{sec:review:results:models}, many multi-label hate speech classifiers are built upon pre-trained \ac{llms}, such as \ac{bert}. To make such models, which have been pre-trained on large corpora of natural language to capture its patterns and concepts, applicable to hate speech classification, they must be fine-tuned with suitable training data. \citet{devlin_bert_2019} have shown that even if the amount of data used for fine-tuning is small, \ac{llms} with proper pre-training can perform well. By contrast, models trained from scratch must rely on limited training data to learn the specific task and develop a general understanding of natural language, which is essential for achieving high performance and generalization capabilities on \ac{nlp} tasks like hate speech classification.
Fourth, \textit{data augmentation and synthetic data generation} can not only be leveraged to mitigate the issue of class imbalance \cite{ibrahim_imbalanced_2018,rottger_hatecheck_2021}, but also to increase the size of datasets. For instance, \citet{chung_conan_2019} used such techniques explicitly to satisfy the data demands of more sophisticated \ac{ml} models.

Beyond these dataset-related issues, we found that the multi-label hate speech classification research landscape is highly fragmented, with individual studies pursuing different research objectives and employing various datasets, labels, methods, and experimental setups. Hence, we argue that that one of the most critical issues in the surveyed field of research is the \textit{missing alignment of scholars (I4)} on various methodological aspects:
\begin{itemize}
    \item First, in \secref{sec:review:results:datasets:definitions} we described a concerning \textit{lack of definitions} of hate speech or related concepts in the literature. Moreover, we found that given definitions \textit{varied significantly} in their scope. As argued above, knowledge of datasets' meta-concepts and their definitions is a prerequisite for any comparative assessment.
    \item Second, \textit{dataset creation is often intransparent}. While some datasets are documented in great detail, several publications describe the annotation process only peripherally or insufficiently define the used labels (see \secref{sec:review:results:datasets:annotation}). As \citet{kumar_multilingual_2023} argue, this not only limits the comprehensibility and reproducibility of works, but also seriously obstructs possibilities to extend, update, or combine datasets.
    \item Third, in \secref{sec:review:results:datasets:structure} we showed that there is a \textit{wide array of label sets}, which differ by structure, label type, incorporated dimension, encompassed meta-concept, and definition. Since thus far most researchers have opted to create novel labels from scratch, label sets are typically only used for individual datasets. This renders their comparison or combination challenging.
    \item Fourth, as described in \secref{sec:review:results:datasets:iaa}, there is \textit{no consensus on \ac{iaa} metrics and their reporting}. The differing use of various metrics reduces the comparability of dataset and annotation quality. Even more concerning, however, is the fact that for a substantial number of datasets, either no \ac{iaa} value is given or the underlying calculation method is not specified.
    \item Finally, as can be seen in \figref{fig:review:metrics}, there is also an \textit{incoherent use of evaluation metrics}. This prevents a meaningful comparison of different models' performance. In this context, some authors did not explain adaptations of binary metrics for multi-label tasks (see \eg, \cite{gunasekara_review_2018,ozler_fine-tuning_2020,mazari_bert-based_2024}) or proposed novel metrics (see \cite{liu_fuzzy_2019}).
\end{itemize}

On the side of practitioners from various domains, these inconsistencies complicate decision-making on effective design choices, whereas for scholars, the establishment of a state-of-the-art and of future research potentials is rendered more difficult. The latter is also valid for this survey. While we could not establish one state-of-the-art model (see \secref{sec:review:discussion:limitations}), we hold that utilizing large foundation models might be a promising direction for future research (see \secref{sec:review:discussion:recommendations}).

 \subsection{Recommendations for Future Research on Multi-Label Hate Speech Classification}
 \label{sec:review:discussion:recommendations}
To gradually overcome these issues, improve the methodological alignment within the field of multi-label hate speech classification, and address major gaps in the literature, we formulate ten recommendations for research.

\textit{Report data collection and annotation processes precisely and transparently (R1).} Our analysis revealed significant transparency deficits for several datasets, particularly regarding label definitions (see \secref{sec:review:results:datasets:definitions}) and annotation procedure documentation (see \secref{sec:review:results:datasets:annotation}). Hence, we recommend creators of future datasets to publish their annotation guidelines and describe their entire methodological approach in a way that allows for its replication.

\textit{Use diverse sources for dataset creation (R2).} Twitter/X emerged as the most popular data source, and most datasets only contained data from one platform (see \secref{sec:review:results:datasets:source}). However, differences among individual platforms, \eg, regarding terms of service, technical restrictions, moderation, and user discourse, can introduce biases to datasets that limit the generalizability of models that are trained on them. We posit that including content from diverse platforms in datasets might offer some mitigation for this.

\textit{Explicitly define underlying meta-concepts and their label-set encodes (R3).} As described in \secref{sec:review:results:datasets:definitions}, the surveyed datasets adopt various meta-concepts, which are often neither defined nor derived from related work in the accompanying publications. In future research, scholars should thus transparently situate definitions within the research landscape and only create novel definitions if absolutely necessary. The application of established definitions consolidates the research field and increases comparability.

\textit{Consolidate label sets, by expanding, combining, or including existing label (sub-)sets (R4).} Since there is also a large variance among the label sets with only a limited amount of overlap (see \tabref{tab:review:datasets}), researchers could strive to consolidate these through careful combinations and extensions. This would, on the one hand, reduce the number of variables that have to be considered when comparing datasets and, on the other hand, increase possibilities regarding the combination and expansion of datasets to enhance the sample number and diversity within datasets.

\textit{Calculate \ac{iaa} values and report them with the respective metrics (R5).} Given that no \ac{iaa} was calculated for a significant proportion of surveyed datasets and the concrete metric remained uncertain for others (see \secref{sec:review:results:datasets:iaa}), we identify a clear improvement potential in this regard. \ac{iaa} metrics represent an abstraction from the concrete number of agreements and disagreements during annotation, which should ideally also be reported, and facilitate the comparability of different annotation processes.\footnote{Differences in \ac{iaa} metrics should not be used as the sole indicator of annotation quality, as the values heavily depend on the samples annotated, the annotators tasked with label creation, the label-set employed, the annotation guidelines provided, and the given socio-political context.}

\textit{Report a broad spectrum of well-defined evaluation metrics for multi-label tasks (R6).} As can be seen in \figref{fig:review:metrics}, scholars report various metrics to evaluate their multi-label models. Furthermore, different methods are used to calculate them, which are not always made transparent (see \secref{sec:review:results:models:evaluation}). Until scholars converge to report a fixed set of metrics, which then constitute a new standard, calculating and communicating a variety of metrics facilitates the comparison of different models' performance.

\textit{Evaluate model performance on existing hate speech datasets (R7).} The models covered in this paper were trained and evaluated on a total of 16 different datasets (see \tabref{tab:review:model_types}). Only the datasets from \citet{cjadams_toxic_2017} (\ac{tccc}) and \cite{parikh_multi-label_2019} were utilized multiple times. We recommend researchers to evaluate novel model architectures not only on newly created datasets with multi-label annotations but also on previously published ones to enable a continuous performance comparison between new and old models and model architectures.

\textit{Organize additional multi-label specific shared tasks (R8).} Shared tasks with a multi-label focus may not only incentivize researchers to address respective hate speech classification problems, but are also often accompanied by novel datasets that serve as common benchmarks for the participants' models and constitute a valuable resource for the entire research community. In our survey, ten of 28 datasets were associated with a shared task. Unfortunately, eight of them were not publicly accessible without registration or authentication (see \secref{sec:review:results:datasets:availability}). Hence, we encourage the research community to organize additional shared tasks and make the associated datasets public.

\textit{Investigate multi-label hate speech classification with regard to criminal relevance (R9).} Our survey revealed no datasets, label sets, or models specifically designed to align with legal frameworks (see \tabref{tab:review:datasets} and \tabref{tab:review:model_types}). While there is some empirical and conceptual literature on it for the German \cite{baumler_towards_2024,baumler_cyber_2025} and Slovenian context \cite{fiser_legal_2017}, and researchers created multi-label datasets \cite{demus_comprehensive_2022,zufall_legal_2022} and multi-class classifiers \cite{schafer_bias_2023} with such a focus for the German language, there appears to be a research gap regarding other languages and jurisdictions.

\textit{Utilize large foundation models for multi-label hate speech classification (R10).} Among the surveyed works proposing novel model architectures, none utilized large foundation models, such as GPT-4 and Llama 3, for multi-label classification tasks (see \figref{fig:review:model_types_year}). However, as highlighted in \secref{sec:review:discussion:open_issues}, available training datasets exhibit several issues that complicate multi-label hate speech classification, including data sparsity, labeling biases, and multiple target sets. Therefore, in line with the general direction of \ac{nlp} research, we identify large foundation models as one of the most promising research directions. Due to their extensive training data, these models can perform zero-shot classification, meaning they require no task-specific data while still outperforming smaller models trained on labeled task data \cite{brown_language_2020}. Moreover, these models are instruction-trained, allowing for the inclusion of hate speech class definitions without explicitly encoding them in the training data. This makes them particularly useful for multi-label scenarios. While foundation models have shown impressive results in binary hate speech classification \cite{albladi_hate_2025}, and research demonstrates their empirical effectiveness for general multi-label problems \cite{peskine_definitions_2023}, we are not aware of works explicitly exploring them for multi-label hate speech classification.

\subsection{Limitations}
\label{sec:review:discussion:limitations}
Our systematic literature survey is subject to several limitations.
\textit{First}, to ensure the practical feasibility of a systematic and reproducible approach, we had to implement strict search parameters (\eg, keywords, databases) as well as inclusion and exclusion criteria (\eg, data modality, language, concept, accessibility). While they inevitably limit the breadth and depth of our work, they point toward opportunities for future literature surveys, \eg, on multi-label hate speech classification in multi-modal or multi-lingual data.

\textit{Second}, we chose to restrict the extent of our forward and backward search, since the first stage of our literature search procedure was already quite extensive. For both, we adopted a search-depth of one, and in the forward search, we only included papers with 50 or more citations on Google Scholar. This may limit the representativeness of our results, particularly regarding newer publications and datasets. Many datasets were only found during the second stage, as they were often published in repositories and cited by authors in context of model training. For future surveys, an increased search depth and a less rigorous filter threshold during the forward and backward search may thus yield more comprehensive and current results.

\textit{Third}, we could not determine a state-of-the-art model for multi-label hate speech classification, since the datasets and metrics used for evaluation varied significantly among the surveyed works (see \secref{sec:review:discussion:open_issues}). However, for another study, we have used the HateCheck dataset \cite{rottger_hatecheck_2021}, to evaluate the performance of the most prominent models for binary hate speech classification (see \secref{sec:review:appendix:SotA} for details). All of the tested models were built upon \ac{bert} or RoBERTa, as transformer models are most commonly used in contemporary binary hate speech classification. We found Facebook’s RoBERTa Hate Speech Model to be the by-far best-performing model across all tested metrics, besides \acl{p}, where Google’s Perspective \ac{api} scores best (see \tabref{tab:review:sota}). Therefore, the approach of \citet{vidgen_learning_2021} to train a base model on a dynamically growing dataset over multiple ``human-in-the-loop'' rounds, where experts tried to trick the model, and to generate real world-like samples which are tailored to specific marginalized groups, seems to be working pretty well. Hence, future work could explore the suitability of similar approaches for multi-label tasks.
\section{Conclusion}
\label{sec:review:conclusion}
Motivated by the relevance of differentiating between often overlapping sub-types of hate speech in domains like content moderation or law enforcement, we conducted the, to our knowledge, first systematic and comprehensive survey of scientific literature on textual multi-label hate speech classification in English. Based on our analysis of 46 publications, we contributed to research with a concise overview of 28 datasets that can be used to train multi-label hate speech classification models. Our comparison of them revealed significant heterogeneity, especially regarding their label-set, label-set structure, size, meta-concept, annotation process, and \ac{iaa}, whereas Twitter/X emerged as the most common data source. Moreover, we comparatively analyzed 24 publications that propose model architectures for multi-label tasks, uncovering an inconsistent use of diverse evaluation metrics and establishing \ac{bert} and popular \ac{rnn} architectures, such \ac{bilstm}s with attention layers, as most common. On that basis, we identified imbalanced training data, common reliance on crowdsourcing platforms in annotation, small and sparse datasets, and missing methodological alignment as critical open issues for the research community. Finally, we formulated ten recommendations for future research, focusing on transparency in dataset creation, diversity of data sources, conceptual rigor, consolidated label sets, \ac{iaa}-value calculation, evaluation metric communication, and model validation on existing datasets. We further encouraged scholars to organize multi-label specific shared tasks, investigate multi-label classification in relation to criminal relevance, and explore the utility of large foundation models.

\begin{acks}
This research work has been co-funded by the German Federal Ministry of Education and Research (BMBF) in the project  CYLENCE (13N16636), as well as by the BMBF and the Hessian Ministry of Higher Education, Research, Science and the Arts (HMWK) within their joint support of the National Research Center for Applied Cybersecurity ATHENE.
\end{acks}

\bibliographystyle{ACM-Reference-Format}
\bibliography{bibliography}


\begin{thebibliography}{117}


\ifx \showCODEN    \undefined \def \showCODEN     #1{\unskip}     \fi
\ifx \showISBNx    \undefined \def \showISBNx     #1{\unskip}     \fi
\ifx \showISBNxiii \undefined \def \showISBNxiii  #1{\unskip}     \fi
\ifx \showISSN     \undefined \def \showISSN      #1{\unskip}     \fi
\ifx \showLCCN     \undefined \def \showLCCN      #1{\unskip}     \fi
\ifx \shownote     \undefined \def \shownote      #1{#1}          \fi
\ifx \showarticletitle \undefined \def \showarticletitle #1{#1}   \fi
\ifx \showURL      \undefined \def \showURL       {\relax}        \fi
\providecommand\bibfield[2]{#2}
\providecommand\bibinfo[2]{#2}
\providecommand\natexlab[1]{#1}
\providecommand\showeprint[2][]{arXiv:#2}

\bibitem[Abbasi et~al\mbox{.}(2022)]%
        {abbasi_deep_2022}
\bibfield{author}{\bibinfo{person}{Ahmed Abbasi}, \bibinfo{person}{Abdul~Rehman
  Javed}, \bibinfo{person}{Farkhund Iqbal}, \bibinfo{person}{Natalia
  Kryvinska}, {and} \bibinfo{person}{Zunera Jalil}.}
  \bibinfo{year}{2022}\natexlab{}.
\newblock \showarticletitle{Deep learning for religious and continent-based
  toxic content detection and classification}.
\newblock \bibinfo{journal}{\emph{Scientific Reports}} \bibinfo{volume}{12},
  \bibinfo{number}{1} (\bibinfo{date}{Oct.} \bibinfo{year}{2022}),
  \bibinfo{pages}{17478}.
\newblock
\showISSN{2045-2322}
\href{https://doi.org/10.1038/s41598-022-22523-3}{doi:\nolinkurl{10.1038/s41598-022-22523-3}}


\bibitem[Abburi et~al\mbox{.}(2021)]%
        {abburi_fine-grained_2021}
\bibfield{author}{\bibinfo{person}{Harika Abburi}, \bibinfo{person}{Pulkit
  Parikh}, \bibinfo{person}{Niyati Chhaya}, {and} \bibinfo{person}{Vasudeva
  Varma}.} \bibinfo{year}{2021}\natexlab{}.
\newblock \showarticletitle{Fine-{Grained} {Multi}-label {Sexism}
  {Classification} {Using} a {Semi}-{Supervised} {Multi}-level {Neural}
  {Approach}}.
\newblock \bibinfo{journal}{\emph{Data Science and Engineering}}
  \bibinfo{volume}{6}, \bibinfo{number}{4} (\bibinfo{date}{Dec.}
  \bibinfo{year}{2021}), \bibinfo{pages}{359--379}.
\newblock
\showISSN{2364-1185, 2364-1541}
\href{https://doi.org/10.1007/s41019-021-00168-y}{doi:\nolinkurl{10.1007/s41019-021-00168-y}}


\bibitem[Abburi et~al\mbox{.}(2024)]%
        {abburi_multi-task_2024}
\bibfield{author}{\bibinfo{person}{Harika Abburi}, \bibinfo{person}{Pulkit
  Parikh}, \bibinfo{person}{Niyati Chhaya}, {and} \bibinfo{person}{Vasudeva
  Varma}.} \bibinfo{year}{2024}\natexlab{}.
\newblock \showarticletitle{Multi-task learning neural framework for
  categorizing sexism}.
\newblock \bibinfo{journal}{\emph{Computer Speech \& Language}}
  \bibinfo{volume}{83} (\bibinfo{date}{Jan.} \bibinfo{year}{2024}),
  \bibinfo{pages}{101535}.
\newblock
\showISSN{08852308}
\href{https://doi.org/10.1016/j.csl.2023.101535}{doi:\nolinkurl{10.1016/j.csl.2023.101535}}


\bibitem[Aditya et~al\mbox{.}(2022)]%
        {aditya_classifying_2022}
\bibfield{author}{\bibinfo{person}{Apoorv Aditya}, \bibinfo{person}{Rithwik
  Vinod}, \bibinfo{person}{Ashish Kumar}, \bibinfo{person}{Ishan Bhowmik},
  {and} \bibinfo{person}{J. Swaminathan}.} \bibinfo{year}{2022}\natexlab{}.
\newblock \showarticletitle{Classifying {Speech} into {Offensive} and {Hate}
  {Categories} along with {Targeted} {Communities} using {Machine} {Learning}}.
  In \bibinfo{booktitle}{\emph{2022 {International} {Conference} on {Inventive}
  {Computation} {Technologies} ({ICICT})}}. \bibinfo{publisher}{IEEE},
  \bibinfo{address}{Nepal}, \bibinfo{pages}{291--295}.
\newblock
\showISBNx{978-1-66540-837-0}
\href{https://doi.org/10.1109/ICICT54344.2022.9850944}{doi:\nolinkurl{10.1109/ICICT54344.2022.9850944}}


\bibitem[Albladi et~al\mbox{.}(2025)]%
        {albladi_hate_2025}
\bibfield{author}{\bibinfo{person}{Aish Albladi}, \bibinfo{person}{Minarul
  Islam}, \bibinfo{person}{Amit Das}, \bibinfo{person}{Maryam Bigonah},
  \bibinfo{person}{Zheng Zhang}, \bibinfo{person}{Fatemeh Jamshidi},
  \bibinfo{person}{Mostafa Rahgouy}, \bibinfo{person}{Nilanjana Raychawdhary},
  \bibinfo{person}{Daniela Marghitu}, {and} \bibinfo{person}{Cheryl Seals}.}
  \bibinfo{year}{2025}\natexlab{}.
\newblock \showarticletitle{Hate {Speech} {Detection} using {Large} {Language}
  {Models}: {A} {Comprehensive} {Review}}.
\newblock \bibinfo{journal}{\emph{IEEE Access}}  \bibinfo{volume}{13}
  (\bibinfo{year}{2025}), \bibinfo{pages}{20871--20892}.
\newblock
\showISSN{2169-3536}
\href{https://doi.org/10.1109/ACCESS.2025.3532397}{doi:\nolinkurl{10.1109/ACCESS.2025.3532397}}


\bibitem[Alkomah and Ma(2022)]%
        {alkomah_literature_2022}
\bibfield{author}{\bibinfo{person}{Fatimah Alkomah} {and}
  \bibinfo{person}{Xiaogang Ma}.} \bibinfo{year}{2022}\natexlab{}.
\newblock \showarticletitle{A {Literature} {Review} of {Textual} {Hate}
  {Speech} {Detection} {Methods} and {Datasets}}.
\newblock \bibinfo{journal}{\emph{Information}} \bibinfo{volume}{13},
  \bibinfo{number}{6} (\bibinfo{date}{May} \bibinfo{year}{2022}),
  \bibinfo{pages}{273}.
\newblock
\showISSN{2078-2489}
\href{https://doi.org/10.3390/info13060273}{doi:\nolinkurl{10.3390/info13060273}}


\bibitem[Alshamrani et~al\mbox{.}(2021)]%
        {alshamrani_hate_2021}
\bibfield{author}{\bibinfo{person}{Sultan Alshamrani}, \bibinfo{person}{Ahmed
  Abusnaina}, \bibinfo{person}{Mohammed Abuhamad}, \bibinfo{person}{Daehun
  Nyang}, {and} \bibinfo{person}{David Mohaisen}.}
  \bibinfo{year}{2021}\natexlab{}.
\newblock \showarticletitle{Hate, {Obscenity}, and {Insults}: {Measuring} the
  {Exposure} of {Children} to {Inappropriate} {Comments} in {YouTube}}. In
  \bibinfo{booktitle}{\emph{Companion {Proceedings} of the {Web} {Conference}
  2021}}. \bibinfo{publisher}{ACM}, \bibinfo{address}{Ljubljana Slovenia},
  \bibinfo{pages}{508--515}.
\newblock
\showISBNx{978-1-4503-8313-4}
\href{https://doi.org/10.1145/3442442.3452314}{doi:\nolinkurl{10.1145/3442442.3452314}}


\bibitem[Aluru et~al\mbox{.}(2020)]%
        {aluru_deep_2020}
\bibfield{author}{\bibinfo{person}{Sai~Saketh Aluru}, \bibinfo{person}{Binny
  Mathew}, \bibinfo{person}{Punyajoy Saha}, {and} \bibinfo{person}{Animesh
  Mukherjee}.} \bibinfo{year}{2020}\natexlab{}.
\newblock \bibinfo{title}{Deep {Learning} {Models} for {Multilingual} {Hate}
  {Speech} {Detection}}.
\newblock
\href{https://doi.org/10.48550/ARXIV.2004.06465}{doi:\nolinkurl{10.48550/ARXIV.2004.06465}}
\newblock
\shownote{Version Number: 3}.


\bibitem[Antypas et~al\mbox{.}(2023)]%
        {antypas_supertweeteval_2023}
\bibfield{author}{\bibinfo{person}{Dimosthenis Antypas}, \bibinfo{person}{Asahi
  Ushio}, \bibinfo{person}{Francesco Barbieri}, \bibinfo{person}{Leonardo
  Neves}, \bibinfo{person}{Kiamehr Rezaee}, \bibinfo{person}{Luis
  Espinosa-Anke}, \bibinfo{person}{Jiaxin Pei}, {and} \bibinfo{person}{Jose
  Camacho-Collados}.} \bibinfo{year}{2023}\natexlab{}.
\newblock \showarticletitle{{SuperTweetEval}: {A} {Challenging}, {Unified} and
  {Heterogeneous} {Benchmark} for {Social} {Media} {NLP} {Research}}. In
  \bibinfo{booktitle}{\emph{Findings of the {Association} for {Computational}
  {Linguistics}: {EMNLP} 2023}}. \bibinfo{publisher}{Association for
  Computational Linguistics}, \bibinfo{address}{Singapore},
  \bibinfo{pages}{12590--12607}.
\newblock
\href{https://doi.org/10.18653/v1/2023.findings-emnlp.838}{doi:\nolinkurl{10.18653/v1/2023.findings-emnlp.838}}


\bibitem[Anzovino et~al\mbox{.}(2018)]%
        {silberztein_automatic_2018}
\bibfield{author}{\bibinfo{person}{Maria Anzovino}, \bibinfo{person}{Elisabetta
  Fersini}, {and} \bibinfo{person}{Paolo Rosso}.}
  \bibinfo{year}{2018}\natexlab{}.
\newblock \showarticletitle{Automatic {Identification} and {Classification} of
  {Misogynistic} {Language} on {Twitter}}.
\newblock In \bibinfo{booktitle}{\emph{Natural {Language} {Processing} and
  {Information} {Systems}}}, \bibfield{editor}{\bibinfo{person}{Max
  Silberztein}, \bibinfo{person}{Faten Atigui}, \bibinfo{person}{Elena
  Kornyshova}, \bibinfo{person}{Elisabeth Métais}, {and}
  \bibinfo{person}{Farid Meziane}} (Eds.). Vol.~\bibinfo{volume}{10859}.
  \bibinfo{publisher}{Springer International Publishing},
  \bibinfo{address}{Cham}, \bibinfo{pages}{57--64}.
\newblock
\showISBNx{978-3-319-91946-1 978-3-319-91947-8}
\href{https://doi.org/10.1007/978-3-319-91947-8_6}{doi:\nolinkurl{10.1007/978-3-319-91947-8_6}}
\newblock
\shownote{Series Title: Lecture Notes in Computer Science}.


\bibitem[Ayo et~al\mbox{.}(2020)]%
        {ayo_machine_2020}
\bibfield{author}{\bibinfo{person}{Femi~Emmanuel Ayo},
  \bibinfo{person}{Olusegun Folorunso}, \bibinfo{person}{Friday~Thomas
  Ibharalu}, {and} \bibinfo{person}{Idowu~Ademola Osinuga}.}
  \bibinfo{year}{2020}\natexlab{}.
\newblock \showarticletitle{Machine learning techniques for hate speech
  classification of twitter data: {State}-of-the-art, future challenges and
  research directions}.
\newblock \bibinfo{journal}{\emph{Computer Science Review}}
  \bibinfo{volume}{38} (\bibinfo{year}{2020}), \bibinfo{pages}{100311}.
\newblock
\showISSN{1574-0137}
\href{https://doi.org/10.1016/j.cosrev.2020.100311}{doi:\nolinkurl{10.1016/j.cosrev.2020.100311}}


\bibitem[Banko et~al\mbox{.}(2020)]%
        {banko_unified_2020}
\bibfield{author}{\bibinfo{person}{Michele Banko}, \bibinfo{person}{Brendon
  MacKeen}, {and} \bibinfo{person}{Laurie Ray}.}
  \bibinfo{year}{2020}\natexlab{}.
\newblock \showarticletitle{A {Unified} {Taxonomy} of {Harmful} {Content}}. In
  \bibinfo{booktitle}{\emph{Proceedings of the {Fourth} {Workshop} on {Online}
  {Abuse} and {Harms}}}. \bibinfo{publisher}{Association for Computational
  Linguistics}, \bibinfo{address}{Online}, \bibinfo{pages}{125--137}.
\newblock
\href{https://doi.org/10.18653/v1/2020.alw-1.16}{doi:\nolinkurl{10.18653/v1/2020.alw-1.16}}


\bibitem[Basile et~al\mbox{.}(2019)]%
        {basile_semeval-2019_2019}
\bibfield{author}{\bibinfo{person}{Valerio Basile}, \bibinfo{person}{Cristina
  Bosco}, \bibinfo{person}{Elisabetta Fersini}, \bibinfo{person}{Debora Nozza},
  \bibinfo{person}{Viviana Patti}, \bibinfo{person}{Francisco~Manuel
  Rangel~Pardo}, \bibinfo{person}{Paolo Rosso}, {and} \bibinfo{person}{Manuela
  Sanguinetti}.} \bibinfo{year}{2019}\natexlab{}.
\newblock \showarticletitle{{SemEval}-2019 {Task} 5: {Multilingual} {Detection}
  of {Hate} {Speech} {Against} {Immigrants} and {Women} in {Twitter}}. In
  \bibinfo{booktitle}{\emph{Proceedings of the 13th {International} {Workshop}
  on {Semantic} {Evaluation}}}. \bibinfo{publisher}{Association for
  Computational Linguistics}, \bibinfo{address}{Minneapolis, Minnesota, USA},
  \bibinfo{pages}{54--63}.
\newblock
\href{https://doi.org/10.18653/v1/S19-2007}{doi:\nolinkurl{10.18653/v1/S19-2007}}


\bibitem[Bhattacharjee et~al\mbox{.}(2023)]%
        {bhattacharjee_sentiment_2023}
\bibfield{author}{\bibinfo{person}{Deeshant Bhattacharjee},
  \bibinfo{person}{Avishek Paul}, {and} \bibinfo{person}{Deepak Kumar}.}
  \bibinfo{year}{2023}\natexlab{}.
\newblock \showarticletitle{Sentiment {Analysis} for {Hateful} {Content} on
  {Social} {Media}}. In \bibinfo{booktitle}{\emph{2023 {International}
  {Conference} on {Network}, {Multimedia} and {Information} {Technology}
  ({NMITCON})}}. \bibinfo{publisher}{IEEE}, \bibinfo{address}{Bengaluru,
  India}, \bibinfo{pages}{1--6}.
\newblock
\showISBNx{9798350300826}
\href{https://doi.org/10.1109/NMITCON58196.2023.10275876}{doi:\nolinkurl{10.1109/NMITCON58196.2023.10275876}}


\bibitem[Bhattacharya et~al\mbox{.}(2020)]%
        {bhattacharya_developing_2020}
\bibfield{author}{\bibinfo{person}{Shiladitya Bhattacharya},
  \bibinfo{person}{Siddharth Singh}, \bibinfo{person}{Ritesh Kumar},
  \bibinfo{person}{Akanksha Bansal}, \bibinfo{person}{Akash Bhagat},
  \bibinfo{person}{Yogesh Dawer}, \bibinfo{person}{Bornini Lahiri}, {and}
  \bibinfo{person}{Atul~Kr. Ojha}.} \bibinfo{year}{2020}\natexlab{}.
\newblock \showarticletitle{Developing a {Multilingual} {Annotated} {Corpus} of
  {Misogyny} and {Aggression}}. In \bibinfo{booktitle}{\emph{Proceedings of the
  {Second} {Workshop} on {Trolling}, {Aggression} and {Cyberbullying}}},
  \bibfield{editor}{\bibinfo{person}{Ritesh Kumar}, \bibinfo{person}{Atul~Kr.
  Ojha}, \bibinfo{person}{Bornini Lahiri}, \bibinfo{person}{Marcos Zampieri},
  \bibinfo{person}{Shervin Malmasi}, \bibinfo{person}{Vanessa Murdock}, {and}
  \bibinfo{person}{Daniel Kadar}} (Eds.). \bibinfo{publisher}{European Language
  Resources Association (ELRA)}, \bibinfo{address}{Marseille, France},
  \bibinfo{pages}{158--168}.
\newblock
\showISBNx{979-10-95546-56-6}
\urldef\tempurl%
\url{https://aclanthology.org/2020.trac-1.25}
\showURL{%
\tempurl}


\bibitem[Brown et~al\mbox{.}(2020)]%
        {brown_language_2020}
\bibfield{author}{\bibinfo{person}{Tom Brown}, \bibinfo{person}{Benjamin Mann},
  \bibinfo{person}{Nick Ryder}, \bibinfo{person}{Melanie Subbiah},
  \bibinfo{person}{Jared~D Kaplan}, \bibinfo{person}{Prafulla Dhariwal},
  \bibinfo{person}{Arvind Neelakantan}, \bibinfo{person}{Pranav Shyam},
  \bibinfo{person}{Girish Sastry}, \bibinfo{person}{Amanda Askell},
  \bibinfo{person}{Sandhini Agarwal}, \bibinfo{person}{Ariel Herbert-Voss},
  \bibinfo{person}{Gretchen Krueger}, \bibinfo{person}{Tom Henighan},
  \bibinfo{person}{Rewon Child}, \bibinfo{person}{Aditya Ramesh},
  \bibinfo{person}{Daniel Ziegler}, \bibinfo{person}{Jeffrey Wu},
  \bibinfo{person}{Clemens Winter}, \bibinfo{person}{Chris Hesse},
  \bibinfo{person}{Mark Chen}, \bibinfo{person}{Eric Sigler},
  \bibinfo{person}{Mateusz Litwin}, \bibinfo{person}{Scott Gray},
  \bibinfo{person}{Benjamin Chess}, \bibinfo{person}{Jack Clark},
  \bibinfo{person}{Christopher Berner}, \bibinfo{person}{Sam McCandlish},
  \bibinfo{person}{Alec Radford}, \bibinfo{person}{Ilya Sutskever}, {and}
  \bibinfo{person}{Dario Amodei}.} \bibinfo{year}{2020}\natexlab{}.
\newblock \showarticletitle{Language {Models} are {Few}-{Shot} {Learners}}. In
  \bibinfo{booktitle}{\emph{34th {Conference} on {Neural} {Information}
  {Processing} {Systems}}}, \bibfield{editor}{\bibinfo{person}{H.~Larochelle},
  \bibinfo{person}{M.~Ranzato}, \bibinfo{person}{R.~Hadsell},
  \bibinfo{person}{M.~F. Balcan}, {and} \bibinfo{person}{H.~Lin}} (Eds.),
  Vol.~\bibinfo{volume}{33}. \bibinfo{publisher}{Curran Associates, Inc.},
  \bibinfo{address}{Vancouver, Canada}, \bibinfo{pages}{1877--1901}.
\newblock
\urldef\tempurl%
\url{https://proceedings.neurips.cc/paper_files/paper/2020/file/1457c0d6bfcb4967418bfb8ac142f64a-Paper.pdf}
\showURL{%
\tempurl}


\bibitem[Burnap and Williams(2016)]%
        {burnap_us_2016}
\bibfield{author}{\bibinfo{person}{Pete Burnap} {and}
  \bibinfo{person}{Matthew~L Williams}.} \bibinfo{year}{2016}\natexlab{}.
\newblock \showarticletitle{Us and them: identifying cyber hate on {Twitter}
  across multiple protected characteristics}.
\newblock \bibinfo{journal}{\emph{EPJ Data Science}} \bibinfo{volume}{5},
  \bibinfo{number}{1} (\bibinfo{date}{Dec.} \bibinfo{year}{2016}),
  \bibinfo{pages}{11}.
\newblock
\showISSN{2193-1127}
\href{https://doi.org/10.1140/epjds/s13688-016-0072-6}{doi:\nolinkurl{10.1140/epjds/s13688-016-0072-6}}


\bibitem[Bäumler et~al\mbox{.}(2024)]%
        {baumler_towards_2024}
\bibfield{author}{\bibinfo{person}{Julian Bäumler},
  \bibinfo{person}{Marc-André Kaufhold}, \bibinfo{person}{Georg Voronin},
  {and} \bibinfo{person}{Christian Reuter}.} \bibinfo{year}{2024}\natexlab{}.
\newblock \showarticletitle{Towards an {Online} {Hate} {Speech}
  {Classification} {Scheme} for {German} {Law} {Enforcement} and {Reporting}
  {Centers}: {Insights} from {Research} and {Practice}}. In
  \bibinfo{booktitle}{\emph{Mensch und {Computer} 2024 – {Workshopband}}}.
  \bibinfo{publisher}{Gesellschaft für Informatik e.V.},
  \bibinfo{address}{Karlsruhe, Germany}, \bibinfo{pages}{1--11}.
\newblock
\href{https://doi.org/10.18420/muc2024-mci-ws13-124}{doi:\nolinkurl{10.18420/muc2024-mci-ws13-124}}


\bibitem[Bäumler et~al\mbox{.}(2025a)]%
        {baumler_harnessing_2025}
\bibfield{author}{\bibinfo{person}{Julian Bäumler}, \bibinfo{person}{Thea
  Riebe}, \bibinfo{person}{Marc-André Kaufhold}, {and}
  \bibinfo{person}{Christian Reuter}.} \bibinfo{year}{2025}\natexlab{a}.
\newblock \showarticletitle{Harnessing {Inter}-{Organizational} {Collaboration}
  and {Automation} to {Combat} {Online} {Hate} {Speech}: {A} {Qualitative}
  {Study} with {German} {Reporting} {Centers}}.
\newblock \bibinfo{journal}{\emph{Proc. ACM Hum.-Comput. Interact.}}
  \bibinfo{volume}{9}, \bibinfo{number}{2} (\bibinfo{year}{2025}),
  \bibinfo{pages}{CSCW093}.
\newblock
\href{https://doi.org/10.1145/3710991}{doi:\nolinkurl{10.1145/3710991}}


\bibitem[Bäumler et~al\mbox{.}(2025b)]%
        {baumler_cyber_2025}
\bibfield{author}{\bibinfo{person}{Julian Bäumler}, \bibinfo{person}{Georg
  Voronin}, {and} \bibinfo{person}{Marc-André Kaufhold}.}
  \bibinfo{year}{2025}\natexlab{b}.
\newblock \showarticletitle{Cyber {Hate} {Awareness}: {Information} {Types} and
  {Technologies} {Relevant} to the {Law} {Enforcement} and {Reporting} {Center}
  {Domain}}.
\newblock \bibinfo{journal}{\emph{i-com - Journal of Interactive Media}}
  \bibinfo{volume}{24}, \bibinfo{number}{1} (\bibinfo{year}{2025}),
  \bibinfo{numpages}{20}~pages.
\newblock
\href{https://doi.org/10.1515/icom-2024-0062}{doi:\nolinkurl{10.1515/icom-2024-0062}}


\bibitem[Caselli et~al\mbox{.}(2021)]%
        {caselli_hatebert_2021}
\bibfield{author}{\bibinfo{person}{Tommaso Caselli}, \bibinfo{person}{Valerio
  Basile}, \bibinfo{person}{Jelena Mitrović}, {and} \bibinfo{person}{Michael
  Granitzer}.} \bibinfo{year}{2021}\natexlab{}.
\newblock \showarticletitle{{HateBERT}: {Retraining} {BERT} for {Abusive}
  {Language} {Detection} in {English}}. In
  \bibinfo{booktitle}{\emph{Proceedings of the 5th {Workshop} on {Online}
  {Abuse} and {Harms} ({WOAH} 2021)}}. \bibinfo{publisher}{Association for
  Computational Linguistics}, \bibinfo{address}{Online},
  \bibinfo{pages}{17--25}.
\newblock
\href{https://doi.org/10.18653/v1/2021.woah-1.3}{doi:\nolinkurl{10.18653/v1/2021.woah-1.3}}


\bibitem[Cheriyan et~al\mbox{.}(2021)]%
        {cheriyan_towards_2021}
\bibfield{author}{\bibinfo{person}{Jithin Cheriyan}, \bibinfo{person}{Bastin
  Tony~Roy Savarimuthu}, {and} \bibinfo{person}{Stephen Cranefield}.}
  \bibinfo{year}{2021}\natexlab{}.
\newblock \showarticletitle{Towards offensive language detection and reduction
  in four {Software} {Engineering} communities}. In
  \bibinfo{booktitle}{\emph{Evaluation and {Assessment} in {Software}
  {Engineering}}}. \bibinfo{publisher}{ACM}, \bibinfo{address}{Trondheim
  Norway}, \bibinfo{pages}{254--259}.
\newblock
\showISBNx{978-1-4503-9053-8}
\href{https://doi.org/10.1145/3463274.3463805}{doi:\nolinkurl{10.1145/3463274.3463805}}


\bibitem[Chetty and Alathur(2018)]%
        {chetty_hate_2018}
\bibfield{author}{\bibinfo{person}{Naganna Chetty} {and}
  \bibinfo{person}{Sreejith Alathur}.} \bibinfo{year}{2018}\natexlab{}.
\newblock \showarticletitle{Hate speech review in the context of online social
  networks}.
\newblock \bibinfo{journal}{\emph{Aggression and Violent Behavior}}
  \bibinfo{volume}{40} (\bibinfo{date}{May} \bibinfo{year}{2018}),
  \bibinfo{pages}{108--118}.
\newblock
\showISSN{13591789}
\href{https://doi.org/10.1016/j.avb.2018.05.003}{doi:\nolinkurl{10.1016/j.avb.2018.05.003}}


\bibitem[Chicco et~al\mbox{.}(2021)]%
        {chicco_matthews_2021}
\bibfield{author}{\bibinfo{person}{Davide Chicco}, \bibinfo{person}{Matthijs~J.
  Warrens}, {and} \bibinfo{person}{Giuseppe Jurman}.}
  \bibinfo{year}{2021}\natexlab{}.
\newblock \showarticletitle{The {Matthews} {Correlation} {Coefficient} ({MCC})
  is {More} {Informative} {Than} {Cohen}’s {Kappa} and {Brier} {Score} in
  {Binary} {Classification} {Assessment}}.
\newblock \bibinfo{journal}{\emph{IEEE Access}}  \bibinfo{volume}{9}
  (\bibinfo{year}{2021}), \bibinfo{pages}{78368--78381}.
\newblock
\showISSN{2169-3536}
\href{https://doi.org/10.1109/ACCESS.2021.3084050}{doi:\nolinkurl{10.1109/ACCESS.2021.3084050}}


\bibitem[Chung et~al\mbox{.}(2014)]%
        {chung_empirical_2014}
\bibfield{author}{\bibinfo{person}{Junyoung Chung}, \bibinfo{person}{Caglar
  Gulcehre}, \bibinfo{person}{KyungHyun Cho}, {and} \bibinfo{person}{Yoshua
  Bengio}.} \bibinfo{year}{2014}\natexlab{}.
\newblock \bibinfo{title}{Empirical {Evaluation} of {Gated} {Recurrent}
  {Neural} {Networks} on {Sequence} {Modeling}}.
\newblock
\href{https://doi.org/10.48550/ARXIV.1412.3555}{doi:\nolinkurl{10.48550/ARXIV.1412.3555}}
\newblock
\shownote{Version Number: 1}.


\bibitem[Chung et~al\mbox{.}(2019)]%
        {chung_conan_2019}
\bibfield{author}{\bibinfo{person}{Yi-Ling Chung}, \bibinfo{person}{Elizaveta
  Kuzmenko}, \bibinfo{person}{Serra~Sinem Tekiroglu}, {and}
  \bibinfo{person}{Marco Guerini}.} \bibinfo{year}{2019}\natexlab{}.
\newblock \showarticletitle{{CONAN} - {COunter} {NArratives} through
  {Nichesourcing}: a {Multilingual} {Dataset} of {Responses} to {Fight}
  {Online} {Hate} {Speech}}. In \bibinfo{booktitle}{\emph{Proceedings of the
  57th {Annual} {Meeting} of the {Association} for {Computational}
  {Linguistics}}}. \bibinfo{publisher}{Association for Computational
  Linguistics}, \bibinfo{address}{Florence, Italy},
  \bibinfo{pages}{2819--2829}.
\newblock
\href{https://doi.org/10.18653/v1/P19-1271}{doi:\nolinkurl{10.18653/v1/P19-1271}}


\bibitem[{cjadams} et~al\mbox{.}(2019)]%
        {cjadams_jigsaw_2019}
\bibfield{author}{\bibinfo{person}{{cjadams}}, \bibinfo{person}{Daniel Borkan},
  \bibinfo{person}{{inversion}}, \bibinfo{person}{Jeffrey Sorensen},
  \bibinfo{person}{Lucas Dixon}, \bibinfo{person}{Lucy Vasserman}, {and}
  \bibinfo{person}{{nithum}}.} \bibinfo{year}{2019}\natexlab{}.
\newblock \bibinfo{title}{Jigsaw {Unintended} {Bias} in {Toxicity}
  {Classification}}.
\newblock
\urldef\tempurl%
\url{https://kaggle.com/competitions/jigsaw-unintended-bias-in-toxicity-classification}
\showURL{%
\tempurl}


\bibitem[{cjadams} et~al\mbox{.}(2017)]%
        {cjadams_toxic_2017}
\bibfield{author}{\bibinfo{person}{{cjadams}}, \bibinfo{person}{Jeffrey
  Sorensen}, \bibinfo{person}{Julia Elliott}, \bibinfo{person}{Lucas Dixon},
  \bibinfo{person}{Mark McDonald}, \bibinfo{person}{{nithum}}, {and}
  \bibinfo{person}{Will Cukierski}.} \bibinfo{year}{2017}\natexlab{}.
\newblock \bibinfo{title}{Toxic {Comment} {Classification} {Challenge}}.
\newblock
\urldef\tempurl%
\url{https://kaggle.com/competitions/jigsaw-toxic-comment-classification-challenge}
\showURL{%
\tempurl}


\bibitem[Das et~al\mbox{.}(2022)]%
        {das_data_2022}
\bibfield{author}{\bibinfo{person}{Mithun Das}, \bibinfo{person}{Somnath
  Banerjee}, {and} \bibinfo{person}{Animesh Mukherjee}.}
  \bibinfo{year}{2022}\natexlab{}.
\newblock \showarticletitle{Data {Bootstrapping} {Approaches} to {Improve}
  {Low} {Resource} {Abusive} {Language} {Detection} for {Indic} {Languages}}.
  In \bibinfo{booktitle}{\emph{Proceedings of the 33rd {ACM} {Conference} on
  {Hypertext} and {Social} {Media}}}. \bibinfo{publisher}{ACM},
  \bibinfo{address}{Barcelona Spain}, \bibinfo{pages}{32--42}.
\newblock
\showISBNx{978-1-4503-9233-4}
\href{https://doi.org/10.1145/3511095.3531277}{doi:\nolinkurl{10.1145/3511095.3531277}}


\bibitem[Davidson et~al\mbox{.}(2017)]%
        {davidson_automated_2017}
\bibfield{author}{\bibinfo{person}{Thomas Davidson}, \bibinfo{person}{Dana
  Warmsley}, \bibinfo{person}{Michael Macy}, {and} \bibinfo{person}{Ingmar
  Weber}.} \bibinfo{year}{2017}\natexlab{}.
\newblock \showarticletitle{Automated {Hate} {Speech} {Detection} and the
  {Problem} of {Offensive} {Language}}.
\newblock \bibinfo{journal}{\emph{Proceedings of the International AAAI
  Conference on Web and Social Media}} \bibinfo{volume}{11},
  \bibinfo{number}{1} (\bibinfo{date}{May} \bibinfo{year}{2017}),
  \bibinfo{pages}{512--515}.
\newblock
\showISSN{2334-0770, 2162-3449}
\href{https://doi.org/10.1609/icwsm.v11i1.14955}{doi:\nolinkurl{10.1609/icwsm.v11i1.14955}}


\bibitem[Demus et~al\mbox{.}(2022)]%
        {demus_comprehensive_2022}
\bibfield{author}{\bibinfo{person}{Christoph Demus}, \bibinfo{person}{Jonas
  Pitz}, \bibinfo{person}{Mina Schütz}, \bibinfo{person}{Nadine Probol},
  \bibinfo{person}{Melanie Siegel}, {and} \bibinfo{person}{Dirk Labudde}.}
  \bibinfo{year}{2022}\natexlab{}.
\newblock \showarticletitle{A {Comprehensive} {Dataset} for {German}
  {Offensive} {Language} and {Conversation} {Analysis}}. In
  \bibinfo{booktitle}{\emph{Proceedings of the {Sixth} {Workshop} on {Online}
  {Abuse} and {Harms} ({WOAH})}}. \bibinfo{publisher}{Association for
  Computational Linguistics}, \bibinfo{address}{Seattle, Washington (Hybrid)},
  \bibinfo{pages}{143--153}.
\newblock
\href{https://doi.org/10.18653/v1/2022.woah-1.14}{doi:\nolinkurl{10.18653/v1/2022.woah-1.14}}


\bibitem[Devlin et~al\mbox{.}(2019)]%
        {devlin_bert_2019}
\bibfield{author}{\bibinfo{person}{Jacob Devlin}, \bibinfo{person}{Ming-Wei
  Chang}, \bibinfo{person}{Kenton Lee}, {and} \bibinfo{person}{Kristina
  Toutanova}.} \bibinfo{year}{2019}\natexlab{}.
\newblock \showarticletitle{{BERT}: {Pre}-training of {Deep} {Bidirectional}
  {Transformers} for {Language} {Understanding}}. In
  \bibinfo{booktitle}{\emph{Proceedings of the 2019 {Conference} of the {North}
  {American} {Chapter} of the {Association} for {Computational} {Linguistics}:
  {Human} {Language} {Technologies}}}. \bibinfo{publisher}{Association for
  Computational Linguistics}, \bibinfo{address}{Minneapolis, Minnesota, USA},
  \bibinfo{pages}{4171--4186}.
\newblock
\href{https://doi.org/10.18653/v1/N19-1423}{doi:\nolinkurl{10.18653/v1/N19-1423}}
\newblock
\shownote{\_eprint: 1810.04805}.


\bibitem[Dirting et~al\mbox{.}(2022)]%
        {dirting_multi-label_2022}
\bibfield{author}{\bibinfo{person}{Bakwa~Dunka Dirting},
  \bibinfo{person}{Gloria~A. Chukwudebe}, \bibinfo{person}{Euphemia~Chioma
  Nwokorie}, {and} \bibinfo{person}{Ikechukwu~Ignatius Ayogu}.}
  \bibinfo{year}{2022}\natexlab{}.
\newblock \showarticletitle{Multi-{Label} {Classification} of {Hate} {Speech}
  {Severity} on {Social} {Media} using {BERT} {Model}}. In
  \bibinfo{booktitle}{\emph{2022 {IEEE} {Nigeria} 4th {International}
  {Conference} on {Disruptive} {Technologies} for {Sustainable} {Development}
  ({NIGERCON})}}. \bibinfo{publisher}{IEEE}, \bibinfo{address}{Lagos, Nigeria},
  \bibinfo{pages}{1--5}.
\newblock
\showISBNx{978-1-66547-978-3}
\href{https://doi.org/10.1109/NIGERCON54645.2022.9803164}{doi:\nolinkurl{10.1109/NIGERCON54645.2022.9803164}}


\bibitem[Faal et~al\mbox{.}(2021)]%
        {faal_domain_2021}
\bibfield{author}{\bibinfo{person}{Farshid Faal}, \bibinfo{person}{Jia~Yuan
  Yu}, {and} \bibinfo{person}{Ketra~A Schmitt}.}
  \bibinfo{year}{2021}\natexlab{}.
\newblock \showarticletitle{Domain {Adaptation} {Multi}-task {Deep} {Neural}
  {Network} for {Mitigating} {Unintended} {Bias} in {Toxic} {Language}
  {Detection}}. In \bibinfo{booktitle}{\emph{Proceedings of the 13th
  {International} {Conference} on {Agents} and {Artificial} {Intelligence}
  ({ICAART} 2021)}}, Vol.~\bibinfo{volume}{2}. \bibinfo{publisher}{Scitepress},
  \bibinfo{address}{Virtual Event}, \bibinfo{pages}{932--940}.
\newblock
\urldef\tempurl%
\url{https://www.scitepress.org/Papers/2021/102661/102661.pdf}
\showURL{%
\tempurl}


\bibitem[Fanton et~al\mbox{.}(2021)]%
        {fanton_human---loop_2021}
\bibfield{author}{\bibinfo{person}{Margherita Fanton}, \bibinfo{person}{Helena
  Bonaldi}, \bibinfo{person}{Serra~Sinem Tekiroğlu}, {and}
  \bibinfo{person}{Marco Guerini}.} \bibinfo{year}{2021}\natexlab{}.
\newblock \showarticletitle{Human-in-the-{Loop} for {Data} {Collection}: a
  {Multi}-{Target} {Counter} {Narrative} {Dataset} to {Fight} {Online} {Hate}
  {Speech}}. In \bibinfo{booktitle}{\emph{Proceedings of the 59th {Annual}
  {Meeting} of the {Association} for {Computational} {Linguistics} and the 11th
  {International} {Joint} {Conference} on {Natural} {Language} {Processing}
  ({Volume} 1: {Long} {Papers})}}. \bibinfo{publisher}{Association for
  Computational Linguistics}, \bibinfo{address}{Online},
  \bibinfo{pages}{3226--3240}.
\newblock
\href{https://doi.org/10.18653/v1/2021.acl-long.250}{doi:\nolinkurl{10.18653/v1/2021.acl-long.250}}


\bibitem[Fersini et~al\mbox{.}(2018)]%
        {fersini_overview_2018}
\bibfield{author}{\bibinfo{person}{Elisabetta Fersini}, \bibinfo{person}{Debora
  Nozza}, \bibinfo{person}{Paolo Rosso}, {and} \bibinfo{person}{{others}}.}
  \bibinfo{year}{2018}\natexlab{}.
\newblock \showarticletitle{Overview of the evalita 2018 task on automatic
  misogyny identification (ami)}. In \bibinfo{booktitle}{\emph{{CEUR} workshop
  proceedings}}, Vol.~\bibinfo{volume}{2263}. \bibinfo{publisher}{CEUR-WS},
  \bibinfo{address}{Torino, Italy}, \bibinfo{pages}{1--9}.
\newblock
\href{https://doi.org/10.4000/books.aaccademia.4497}{doi:\nolinkurl{10.4000/books.aaccademia.4497}}


\bibitem[Fišer et~al\mbox{.}(2017)]%
        {fiser_legal_2017}
\bibfield{author}{\bibinfo{person}{Darja Fišer}, \bibinfo{person}{Tomaž
  Erjavec}, {and} \bibinfo{person}{Nikola Ljubešić}.}
  \bibinfo{year}{2017}\natexlab{}.
\newblock \showarticletitle{Legal {Framework}, {Dataset} and {Annotation}
  {Schema} for {Socially} {Unacceptable} {Online} {Discourse} {Practices} in
  {Slovene}}. In \bibinfo{booktitle}{\emph{Proceedings of the {First}
  {Workshop} on {Abusive} {Language} {Online}}}.
  \bibinfo{publisher}{Association for Computational Linguistics},
  \bibinfo{address}{Vancouver, BC, Canada}, \bibinfo{pages}{46--51}.
\newblock
\href{https://doi.org/10.18653/v1/W17-3007}{doi:\nolinkurl{10.18653/v1/W17-3007}}


\bibitem[Fortuna and Nunes(2019)]%
        {fortuna_survey_2019}
\bibfield{author}{\bibinfo{person}{Paula Fortuna} {and}
  \bibinfo{person}{Sérgio Nunes}.} \bibinfo{year}{2019}\natexlab{}.
\newblock \showarticletitle{A {Survey} on {Automatic} {Detection} of {Hate}
  {Speech} in {Text}}.
\newblock \bibinfo{journal}{\emph{Comput. Surveys}} \bibinfo{volume}{51},
  \bibinfo{number}{4} (\bibinfo{date}{July} \bibinfo{year}{2019}),
  \bibinfo{pages}{1--30}.
\newblock
\showISSN{0360-0300, 1557-7341}
\href{https://doi.org/10.1145/3232676}{doi:\nolinkurl{10.1145/3232676}}


\bibitem[Founta et~al\mbox{.}(2018)]%
        {founta_large_2018}
\bibfield{author}{\bibinfo{person}{Antigoni Founta},
  \bibinfo{person}{Constantinos Djouvas}, \bibinfo{person}{Despoina Chatzakou},
  \bibinfo{person}{Ilias Leontiadis}, \bibinfo{person}{Jeremy Blackburn},
  \bibinfo{person}{Gianluca Stringhini}, \bibinfo{person}{Athena Vakali},
  \bibinfo{person}{Michael Sirivianos}, {and} \bibinfo{person}{Nicolas
  Kourtellis}.} \bibinfo{year}{2018}\natexlab{}.
\newblock \showarticletitle{Large {Scale} {Crowdsourcing} and
  {Characterization} of {Twitter} {Abusive} {Behavior}}.
\newblock \bibinfo{journal}{\emph{Proceedings of the International AAAI
  Conference on Web and Social Media}} \bibinfo{volume}{12},
  \bibinfo{number}{1} (\bibinfo{date}{June} \bibinfo{year}{2018}),
  \bibinfo{pages}{491--500}.
\newblock
\showISSN{2334-0770, 2162-3449}
\href{https://doi.org/10.1609/icwsm.v12i1.14991}{doi:\nolinkurl{10.1609/icwsm.v12i1.14991}}


\bibitem[Gagliardone et~al\mbox{.}(2015)]%
        {gagliardone_countering_2015}
\bibfield{author}{\bibinfo{person}{Ignio Gagliardone}, \bibinfo{person}{Danit
  Gal}, \bibinfo{person}{Thiago Alves}, {and} \bibinfo{person}{Gabriela
  Martinez}.} \bibinfo{year}{2015}\natexlab{}.
\newblock \bibinfo{booktitle}{\emph{Countering {Online} {Hate} {Speech}}}.
\newblock \bibinfo{publisher}{UNESCO Publishing}, \bibinfo{address}{Paris}.
\newblock


\bibitem[Gunasekara and Nejadgholi(2018)]%
        {gunasekara_review_2018}
\bibfield{author}{\bibinfo{person}{Isuru Gunasekara} {and}
  \bibinfo{person}{Isar Nejadgholi}.} \bibinfo{year}{2018}\natexlab{}.
\newblock \showarticletitle{A {Review} of {Standard} {Text} {Classification}
  {Practices} for {Multi}-label {Toxicity} {Identification} of {Online}
  {Content}}. In \bibinfo{booktitle}{\emph{Proceedings of the 2nd {Workshop} on
  {Abusive} {Language} {Online} ({ALW2})}}. \bibinfo{publisher}{Association for
  Computational Linguistics}, \bibinfo{address}{Brussels, Belgium},
  \bibinfo{pages}{21--25}.
\newblock
\href{https://doi.org/10.18653/v1/W18-5103}{doi:\nolinkurl{10.18653/v1/W18-5103}}


\bibitem[Gupta et~al\mbox{.}(2023)]%
        {gupta_hateful_2023}
\bibfield{author}{\bibinfo{person}{Shrey Gupta}, \bibinfo{person}{Pratyush
  Priyadarshi}, {and} \bibinfo{person}{Manish Gupta}.}
  \bibinfo{year}{2023}\natexlab{}.
\newblock \showarticletitle{Hateful {Comment} {Detection} and {Hate} {Target}
  {Type} {Prediction} for {Video} {Comments}}. In
  \bibinfo{booktitle}{\emph{Proceedings of the 32nd {ACM} {International}
  {Conference} on {Information} and {Knowledge} {Management}}}.
  \bibinfo{publisher}{ACM}, \bibinfo{address}{Birmingham United Kingdom},
  \bibinfo{pages}{3923--3927}.
\newblock
\showISBNx{9798400701245}
\href{https://doi.org/10.1145/3583780.3615260}{doi:\nolinkurl{10.1145/3583780.3615260}}


\bibitem[Hartvigsen et~al\mbox{.}(2022)]%
        {hartvigsen_toxigen_2022}
\bibfield{author}{\bibinfo{person}{Thomas Hartvigsen}, \bibinfo{person}{Saadia
  Gabriel}, \bibinfo{person}{Hamid Palangi}, \bibinfo{person}{Maarten Sap},
  \bibinfo{person}{Dipankar Ray}, {and} \bibinfo{person}{Ece Kamar}.}
  \bibinfo{year}{2022}\natexlab{}.
\newblock \showarticletitle{{ToxiGen}: {A} {Large}-{Scale}
  {Machine}-{Generated} {Dataset} for {Adversarial} and {Implicit} {Hate}
  {Speech} {Detection}}. In \bibinfo{booktitle}{\emph{Proceedings of the 60th
  {Annual} {Meeting} of the {Association} for {Computational} {Linguistics}
  ({Volume} 1: {Long} {Papers})}}. \bibinfo{publisher}{Association for
  Computational Linguistics}, \bibinfo{address}{Dublin, Ireland},
  \bibinfo{pages}{3309--3326}.
\newblock
\href{https://doi.org/10.18653/v1/2022.acl-long.234}{doi:\nolinkurl{10.18653/v1/2022.acl-long.234}}


\bibitem[Hee et~al\mbox{.}(2024)]%
        {hee_recent_2024}
\bibfield{author}{\bibinfo{person}{Ming~Shan Hee}, \bibinfo{person}{Shivam
  Sharma}, \bibinfo{person}{Rui Cao}, \bibinfo{person}{Palash Nandi},
  \bibinfo{person}{Preslav Nakov}, \bibinfo{person}{Tanmoy Chakraborty}, {and}
  \bibinfo{person}{Roy Ka-Wei Lee}.} \bibinfo{year}{2024}\natexlab{}.
\newblock \showarticletitle{Recent {Advances} in {Online} {Hate} {Speech}
  {Moderation}: {Multimodality} and the {Role} of {Large} {Models}}. In
  \bibinfo{booktitle}{\emph{Findings of the {Association} for {Computational}
  {Linguistics}: {EMNLP} 2024}}. \bibinfo{publisher}{Association for
  Computational Linguistics}, \bibinfo{address}{Miami, Florida, USA},
  \bibinfo{pages}{4407--4419}.
\newblock
\href{https://doi.org/10.18653/v1/2024.findings-emnlp.254}{doi:\nolinkurl{10.18653/v1/2024.findings-emnlp.254}}


\bibitem[Hermida and Santos(2023)]%
        {hermida_detecting_2023}
\bibfield{author}{\bibinfo{person}{Paulo Cezar De~Q. Hermida} {and}
  \bibinfo{person}{Eulanda M.~Dos Santos}.} \bibinfo{year}{2023}\natexlab{}.
\newblock \showarticletitle{Detecting hate speech in memes: a review}.
\newblock \bibinfo{journal}{\emph{Artificial Intelligence Review}}
  \bibinfo{volume}{56}, \bibinfo{number}{11} (\bibinfo{date}{Nov.}
  \bibinfo{year}{2023}), \bibinfo{pages}{12833--12851}.
\newblock
\showISSN{0269-2821, 1573-7462}
\href{https://doi.org/10.1007/s10462-023-10459-7}{doi:\nolinkurl{10.1007/s10462-023-10459-7}}


\bibitem[Hochreiter(1998)]%
        {hochreiter_vanishing_1998}
\bibfield{author}{\bibinfo{person}{Sepp Hochreiter}.}
  \bibinfo{year}{1998}\natexlab{}.
\newblock \showarticletitle{The {Vanishing} {Gradient} {Problem} {During}
  {Learning} {Recurrent} {Neural} {Nets} and {Problem} {Solutions}}.
\newblock \bibinfo{journal}{\emph{International Journal of Uncertainty,
  Fuzziness and Knowledge-Based Systems}} \bibinfo{volume}{06},
  \bibinfo{number}{02} (\bibinfo{date}{April} \bibinfo{year}{1998}),
  \bibinfo{pages}{107--116}.
\newblock
\showISSN{0218-4885, 1793-6411}
\href{https://doi.org/10.1142/S0218488598000094}{doi:\nolinkurl{10.1142/S0218488598000094}}


\bibitem[Husnain et~al\mbox{.}(2021)]%
        {husnain_novel_2021}
\bibfield{author}{\bibinfo{person}{Muhammad Husnain}, \bibinfo{person}{Adnan
  Khalid}, {and} \bibinfo{person}{Numan Shafi}.}
  \bibinfo{year}{2021}\natexlab{}.
\newblock \showarticletitle{A {Novel} {Preprocessing} {Technique} for {Toxic}
  {Comment} {Classification}}. In \bibinfo{booktitle}{\emph{2021
  {International} {Conference} on {Artificial} {Intelligence} ({ICAI})}}.
  \bibinfo{publisher}{IEEE}, \bibinfo{address}{Islamabad, Pakistan},
  \bibinfo{pages}{22--27}.
\newblock
\showISBNx{978-1-66543-293-1}
\href{https://doi.org/10.1109/ICAI52203.2021.9445252}{doi:\nolinkurl{10.1109/ICAI52203.2021.9445252}}


\bibitem[Ibrahim et~al\mbox{.}(2018)]%
        {ibrahim_imbalanced_2018}
\bibfield{author}{\bibinfo{person}{Mai Ibrahim}, \bibinfo{person}{Marwan
  Torki}, {and} \bibinfo{person}{Nagwa El-Makky}.}
  \bibinfo{year}{2018}\natexlab{}.
\newblock \showarticletitle{Imbalanced {Toxic} {Comments} {Classification}
  {Using} {Data} {Augmentation} and {Deep} {Learning}}. In
  \bibinfo{booktitle}{\emph{2018 17th {IEEE} {International} {Conference} on
  {Machine} {Learning} and {Applications} ({ICMLA})}}.
  \bibinfo{publisher}{IEEE}, \bibinfo{address}{Orlando, FL},
  \bibinfo{pages}{875--878}.
\newblock
\showISBNx{978-1-5386-6805-4}
\href{https://doi.org/10.1109/ICMLA.2018.00141}{doi:\nolinkurl{10.1109/ICMLA.2018.00141}}


\bibitem[Jahan and Oussalah(2023)]%
        {jahan_systematic_2023}
\bibfield{author}{\bibinfo{person}{Md~Saroar Jahan} {and}
  \bibinfo{person}{Mourad Oussalah}.} \bibinfo{year}{2023}\natexlab{}.
\newblock \showarticletitle{A systematic review of hate speech automatic
  detection using natural language processing}.
\newblock \bibinfo{journal}{\emph{Neurocomputing}}  \bibinfo{volume}{546}
  (\bibinfo{date}{Aug.} \bibinfo{year}{2023}), \bibinfo{pages}{126232}.
\newblock
\showISSN{09252312}
\href{https://doi.org/10.1016/j.neucom.2023.126232}{doi:\nolinkurl{10.1016/j.neucom.2023.126232}}


\bibitem[Jeni et~al\mbox{.}(2013)]%
        {jeni_facing_2013}
\bibfield{author}{\bibinfo{person}{Laszlo~A. Jeni}, \bibinfo{person}{Jeffrey~F.
  Cohn}, {and} \bibinfo{person}{Fernando De~La~Torre}.}
  \bibinfo{year}{2013}\natexlab{}.
\newblock \showarticletitle{Facing {Imbalanced} {Data}--{Recommendations} for
  the {Use} of {Performance} {Metrics}}. In \bibinfo{booktitle}{\emph{2013
  {Humaine} {Association} {Conference} on {Affective} {Computing} and
  {Intelligent} {Interaction}}}. \bibinfo{publisher}{IEEE},
  \bibinfo{address}{Geneva, Switzerland}, \bibinfo{pages}{245--251}.
\newblock
\showISBNx{978-0-7695-5048-0}
\href{https://doi.org/10.1109/ACII.2013.47}{doi:\nolinkurl{10.1109/ACII.2013.47}}


\bibitem[Kaufhold et~al\mbox{.}(2023)]%
        {kaufhold_cylence_2023}
\bibfield{author}{\bibinfo{person}{Marc-André Kaufhold},
  \bibinfo{person}{Markus Bayer}, \bibinfo{person}{Julian Bäumler},
  \bibinfo{person}{Christian Reuter}, \bibinfo{person}{Stefan Stieglitz},
  \bibinfo{person}{Ali~Sercan Basyurt}, \bibinfo{person}{Milad Mirbabaie},
  \bibinfo{person}{Christoph Fuchss}, {and} \bibinfo{person}{Kaan Eyilmez}.}
  \bibinfo{year}{2023}\natexlab{}.
\newblock \showarticletitle{{CYLENCE}: {Strategies} and {Tools} for
  {Cross}-{Media} {Reporting}, {Detection}, and {Treatment} of {Cyberbullying}
  and {Hatespeech} in {Law} {Enforcement} {Agencies}}. In
  \bibinfo{booktitle}{\emph{Mensch und {Computer} 2023 - {Workshopband}}}.
  \bibinfo{publisher}{Gesellschaft für Informatik e.V.},
  \bibinfo{address}{Rapperswil, Switzerland}, \bibinfo{pages}{1--8}.
\newblock
\href{https://doi.org/10.18420/MUC2023-MCI-WS01-211}{doi:\nolinkurl{10.18420/MUC2023-MCI-WS01-211}}


\bibitem[Ke et~al\mbox{.}(2017)]%
        {ke_lightgbm_2017}
\bibfield{author}{\bibinfo{person}{Guolin Ke}, \bibinfo{person}{Qi Meng},
  \bibinfo{person}{Thomas Finley}, \bibinfo{person}{Taifeng Wang},
  \bibinfo{person}{Wei Chen}, \bibinfo{person}{Weidong Ma},
  \bibinfo{person}{Qiwei Ye}, {and} \bibinfo{person}{Tie-Yan Liu}.}
  \bibinfo{year}{2017}\natexlab{}.
\newblock \showarticletitle{{LightGBM}: {A} {Highly} {Efficient} {Gradient}
  {Boosting} {Decision} {Tree}}. In \bibinfo{booktitle}{\emph{Advances in
  {Neural} {Information} {Processing} {Systems}}},
  \bibfield{editor}{\bibinfo{person}{I.~Guyon}, \bibinfo{person}{U.~Von
  Luxburg}, \bibinfo{person}{S.~Bengio}, \bibinfo{person}{H.~Wallach},
  \bibinfo{person}{R.~Fergus}, \bibinfo{person}{S.~Vishwanathan}, {and}
  \bibinfo{person}{R.~Garnett}} (Eds.), Vol.~\bibinfo{volume}{30}.
  \bibinfo{publisher}{Curran Associates, Inc.}, \bibinfo{address}{Long Beach,
  CA, USA}, \bibinfo{pages}{1--9}.
\newblock
\urldef\tempurl%
\url{https://proceedings.neurips.cc/paper_files/paper/2017/file/6449f44a102fde848669bdd9eb6b76fa-Paper.pdf}
\showURL{%
\tempurl}


\bibitem[Keipi et~al\mbox{.}(2016)]%
        {keipi_online_2016}
\bibfield{author}{\bibinfo{person}{Teo Keipi}, \bibinfo{person}{Matti Näsi},
  \bibinfo{person}{Atte Oksanen}, {and} \bibinfo{person}{Pekka Räsänen}.}
  \bibinfo{year}{2016}\natexlab{}.
\newblock \bibinfo{booktitle}{\emph{Online {Hate} and {Harmful} {Content}:
  {Cross}-national perspectives} (\bibinfo{edition}{1} ed.)}.
\newblock \bibinfo{publisher}{Routledge}, \bibinfo{address}{London}.
\newblock
\showISBNx{978-1-315-62837-0}
\href{https://doi.org/10.4324/9781315628370}{doi:\nolinkurl{10.4324/9781315628370}}


\bibitem[Kennedy et~al\mbox{.}(2022)]%
        {kennedy_introducing_2022}
\bibfield{author}{\bibinfo{person}{Brendan Kennedy}, \bibinfo{person}{Mohammad
  Atari}, \bibinfo{person}{Aida~Mostafazadeh Davani}, \bibinfo{person}{Leigh
  Yeh}, \bibinfo{person}{Ali Omrani}, \bibinfo{person}{Yehsong Kim},
  \bibinfo{person}{Kris Coombs}, \bibinfo{person}{Shreya Havaldar},
  \bibinfo{person}{Gwenyth Portillo-Wightman}, \bibinfo{person}{Elaine
  Gonzalez}, \bibinfo{person}{Joe Hoover}, \bibinfo{person}{Aida Azatian},
  \bibinfo{person}{Alyzeh Hussain}, \bibinfo{person}{Austin Lara},
  \bibinfo{person}{Gabriel Cardenas}, \bibinfo{person}{Adam Omary},
  \bibinfo{person}{Christina Park}, \bibinfo{person}{Xin Wang},
  \bibinfo{person}{Clarisa Wijaya}, \bibinfo{person}{Yong Zhang},
  \bibinfo{person}{Beth Meyerowitz}, {and} \bibinfo{person}{Morteza Dehghani}.}
  \bibinfo{year}{2022}\natexlab{}.
\newblock \showarticletitle{Introducing the {Gab} {Hate} {Corpus}: defining and
  applying hate-based rhetoric to social media posts at scale}.
\newblock \bibinfo{journal}{\emph{Language Resources and Evaluation}}
  \bibinfo{volume}{56}, \bibinfo{number}{1} (\bibinfo{date}{March}
  \bibinfo{year}{2022}), \bibinfo{pages}{79--108}.
\newblock
\showISSN{1574-020X, 1574-0218}
\href{https://doi.org/10.1007/s10579-021-09569-x}{doi:\nolinkurl{10.1007/s10579-021-09569-x}}


\bibitem[Kralj~Novak et~al\mbox{.}(2022)]%
        {ciucci_handling_2022}
\bibfield{author}{\bibinfo{person}{Petra Kralj~Novak}, \bibinfo{person}{Teresa
  Scantamburlo}, \bibinfo{person}{Andraž Pelicon}, \bibinfo{person}{Matteo
  Cinelli}, \bibinfo{person}{Igor Mozetič}, {and} \bibinfo{person}{Fabiana
  Zollo}.} \bibinfo{year}{2022}\natexlab{}.
\newblock \showarticletitle{Handling {Disagreement} in {Hate} {Speech}
  {Modelling}}. In \bibinfo{booktitle}{\emph{Information {Processing} and
  {Management} of {Uncertainty} in {Knowledge}-{Based} {Systems}}},
  \bibfield{editor}{\bibinfo{person}{Davide Ciucci}, \bibinfo{person}{Inés
  Couso}, \bibinfo{person}{Jesús Medina}, \bibinfo{person}{Dominik Ślęzak},
  \bibinfo{person}{Davide Petturiti}, \bibinfo{person}{Bernadette
  Bouchon-Meunier}, {and} \bibinfo{person}{Ronald~R. Yager}} (Eds.),
  Vol.~\bibinfo{volume}{1602}. \bibinfo{publisher}{Springer International
  Publishing}, \bibinfo{address}{Cham}, \bibinfo{pages}{681--695}.
\newblock
\showISBNx{978-3-031-08973-2 978-3-031-08974-9}
\href{https://doi.org/10.1007/978-3-031-08974-9_54}{doi:\nolinkurl{10.1007/978-3-031-08974-9_54}}
\newblock
\shownote{Series Title: Communications in Computer and Information Science}.


\bibitem[Kumar et~al\mbox{.}(2023)]%
        {kumar_multilingual_2023}
\bibfield{author}{\bibinfo{person}{Ritesh Kumar}, \bibinfo{person}{Shyam
  Ratan}, \bibinfo{person}{Siddharth Singh}, \bibinfo{person}{Enakshi Nandi},
  \bibinfo{person}{Laishram~Niranjana Devi}, \bibinfo{person}{Akash Bhagat},
  \bibinfo{person}{Yogesh Dawer}, \bibinfo{person}{Bornini Lahiri}, {and}
  \bibinfo{person}{Akanksha Bansal}.} \bibinfo{year}{2023}\natexlab{}.
\newblock \showarticletitle{A multilingual, multimodal dataset of aggression
  and bias: the {ComMA} dataset}.
\newblock \bibinfo{journal}{\emph{Language Resources and Evaluation}}
  \bibinfo{volume}{58} (\bibinfo{date}{Nov.} \bibinfo{year}{2023}),
  \bibinfo{pages}{757–837}.
\newblock
\showISSN{1574-020X, 1574-0218}
\href{https://doi.org/10.1007/s10579-023-09696-7}{doi:\nolinkurl{10.1007/s10579-023-09696-7}}


\bibitem[Laaksonen(2023)]%
        {laaksonen_datafication_2023}
\bibfield{author}{\bibinfo{person}{Salla-Maaria Laaksonen}.}
  \bibinfo{year}{2023}\natexlab{}.
\newblock \showarticletitle{The {Datafication} of {Hate} {Speech}}.
\newblock In \bibinfo{booktitle}{\emph{Challenges and perspectives of hate
  speech research}}, \bibfield{editor}{\bibinfo{person}{Christian Strippel},
  \bibinfo{person}{Sünje Paasch-Colberg}, \bibinfo{person}{Martin Emmer},
  {and} \bibinfo{person}{Joachim Trebbe}} (Eds.). \bibinfo{series}{Digital
  {Communication}}, Vol.~\bibinfo{volume}{12}. \bibinfo{publisher}{Böhland \&
  Schremmer Verlag}, \bibinfo{address}{Berlin}, \bibinfo{pages}{301--317}.
\newblock
\urldef\tempurl%
\url{https://www.ssoar.info/ssoar/handle/document/86422}
\showURL{%
\tempurl}
\newblock
\shownote{ISBN: 9783945681121 Publisher: [object Object]}.


\bibitem[Leets(2002)]%
        {leets_experiencing_2002}
\bibfield{author}{\bibinfo{person}{Laura Leets}.}
  \bibinfo{year}{2002}\natexlab{}.
\newblock \showarticletitle{Experiencing {Hate} {Speech}: {Perceptions} and
  {Responses} to {Anti}‐{Semitism} and {Antigay} {Speech}}.
\newblock \bibinfo{journal}{\emph{Journal of Social Issues}}
  \bibinfo{volume}{58}, \bibinfo{number}{2} (\bibinfo{date}{Jan.}
  \bibinfo{year}{2002}), \bibinfo{pages}{341--361}.
\newblock
\showISSN{0022-4537, 1540-4560}
\href{https://doi.org/10.1111/1540-4560.00264}{doi:\nolinkurl{10.1111/1540-4560.00264}}


\bibitem[Lenhart et~al\mbox{.}(2016)]%
        {lenhart_online_2016}
\bibfield{author}{\bibinfo{person}{Amanda Lenhart}, \bibinfo{person}{Michele
  Ybarra}, \bibinfo{person}{Kathryn Zickuhr}, {and} \bibinfo{person}{Myeshia
  Price-Feeney}.} \bibinfo{year}{2016}\natexlab{}.
\newblock \bibinfo{booktitle}{\emph{Online harassment, digital abuse, and
  cyberstalking in {America}}}.
\newblock \bibinfo{type}{Research {Report}}. \bibinfo{institution}{Data and
  Society Research Institute. Center for Innovative Public Research}.
\newblock
\urldef\tempurl%
\url{https://www.datasociety.net/pubs/oh/Online_Harassment_2016.pdf}
\showURL{%
\tempurl}


\bibitem[Liu et~al\mbox{.}(2019)]%
        {liu_fuzzy_2019}
\bibfield{author}{\bibinfo{person}{Han Liu}, \bibinfo{person}{Pete Burnap},
  \bibinfo{person}{Wafa Alorainy}, {and} \bibinfo{person}{Matthew~L.
  Williams}.} \bibinfo{year}{2019}\natexlab{}.
\newblock \showarticletitle{Fuzzy {Multi}-task {Learning} for {Hate} {Speech}
  {Type} {Identification}}. In \bibinfo{booktitle}{\emph{The {World} {Wide}
  {Web} {Conference}}}. \bibinfo{publisher}{ACM}, \bibinfo{address}{San
  Francisco CA USA}, \bibinfo{pages}{3006--3012}.
\newblock
\showISBNx{978-1-4503-6674-8}
\href{https://doi.org/10.1145/3308558.3313546}{doi:\nolinkurl{10.1145/3308558.3313546}}


\bibitem[MacAvaney et~al\mbox{.}(2019)]%
        {macavaney_hate_2019}
\bibfield{author}{\bibinfo{person}{Sean MacAvaney}, \bibinfo{person}{Hao-Ren
  Yao}, \bibinfo{person}{Eugene Yang}, \bibinfo{person}{Katina Russell},
  \bibinfo{person}{Nazli Goharian}, {and} \bibinfo{person}{Ophir Frieder}.}
  \bibinfo{year}{2019}\natexlab{}.
\newblock \showarticletitle{Hate speech detection: {Challenges} and solutions}.
\newblock \bibinfo{journal}{\emph{PLOS ONE}} \bibinfo{volume}{14},
  \bibinfo{number}{8} (\bibinfo{date}{Aug.} \bibinfo{year}{2019}),
  \bibinfo{pages}{e0221152}.
\newblock
\showISSN{1932-6203}
\href{https://doi.org/10.1371/journal.pone.0221152}{doi:\nolinkurl{10.1371/journal.pone.0221152}}


\bibitem[Mandl et~al\mbox{.}(2020)]%
        {mandl_overview_2020}
\bibfield{author}{\bibinfo{person}{Thomas Mandl}, \bibinfo{person}{Sandip
  Modha}, \bibinfo{person}{Anand Kumar~M}, {and} \bibinfo{person}{Bharathi~Raja
  Chakravarthi}.} \bibinfo{year}{2020}\natexlab{}.
\newblock \showarticletitle{Overview of the {HASOC} {Track} at {FIRE} 2020:
  {Hate} {Speech} and {Offensive} {Language} {Identification} in {Tamil},
  {Malayalam}, {Hindi}, {English} and {German}}. In
  \bibinfo{booktitle}{\emph{Forum for {Information} {Retrieval} {Evaluation}}}.
  \bibinfo{publisher}{ACM}, \bibinfo{address}{Hyderabad India},
  \bibinfo{pages}{29--32}.
\newblock
\showISBNx{978-1-4503-8978-5}
\href{https://doi.org/10.1145/3441501.3441517}{doi:\nolinkurl{10.1145/3441501.3441517}}


\bibitem[Mandl et~al\mbox{.}(2019)]%
        {mandl_overview_2019}
\bibfield{author}{\bibinfo{person}{Thomas Mandl}, \bibinfo{person}{Sandip
  Modha}, \bibinfo{person}{Prasenjit Majumder}, \bibinfo{person}{Daksh Patel},
  \bibinfo{person}{Mohana Dave}, \bibinfo{person}{Chintak Mandlia}, {and}
  \bibinfo{person}{Aditya Patel}.} \bibinfo{year}{2019}\natexlab{}.
\newblock \showarticletitle{Overview of the {HASOC} track at {FIRE} 2019:
  {Hate} {Speech} and {Offensive} {Content} {Identification} in
  {Indo}-{European} {Languages}}. In \bibinfo{booktitle}{\emph{Proceedings of
  the 11th {Forum} for {Information} {Retrieval} {Evaluation}}}.
  \bibinfo{publisher}{ACM}, \bibinfo{address}{Kolkata India},
  \bibinfo{pages}{14--17}.
\newblock
\showISBNx{978-1-4503-7750-8}
\href{https://doi.org/10.1145/3368567.3368584}{doi:\nolinkurl{10.1145/3368567.3368584}}


\bibitem[Mansur et~al\mbox{.}(2023)]%
        {mansur_twitter_2023}
\bibfield{author}{\bibinfo{person}{Zainab Mansur}, \bibinfo{person}{Nazlia
  Omar}, {and} \bibinfo{person}{Sabrina Tiun}.}
  \bibinfo{year}{2023}\natexlab{}.
\newblock \showarticletitle{Twitter {Hate} {Speech} {Detection}: {A}
  {Systematic} {Review} of {Methods}, {Taxonomy} {Analysis}, {Challenges}, and
  {Opportunities}}.
\newblock \bibinfo{journal}{\emph{IEEE Access}}  \bibinfo{volume}{11}
  (\bibinfo{year}{2023}), \bibinfo{pages}{16226--16249}.
\newblock
\showISSN{2169-3536}
\href{https://doi.org/10.1109/ACCESS.2023.3239375}{doi:\nolinkurl{10.1109/ACCESS.2023.3239375}}


\bibitem[Mathew et~al\mbox{.}(2021)]%
        {mathew_hatexplain_2021}
\bibfield{author}{\bibinfo{person}{Binny Mathew}, \bibinfo{person}{Punyajoy
  Saha}, \bibinfo{person}{Seid~Muhie Yimam}, \bibinfo{person}{Chris Biemann},
  \bibinfo{person}{Pawan Goyal}, {and} \bibinfo{person}{Animesh Mukherjee}.}
  \bibinfo{year}{2021}\natexlab{}.
\newblock \showarticletitle{{HateXplain}: {A} {Benchmark} {Dataset} for
  {Explainable} {Hate} {Speech} {Detection}}.
\newblock \bibinfo{journal}{\emph{Proceedings of the AAAI Conference on
  Artificial Intelligence}} \bibinfo{volume}{35}, \bibinfo{number}{17}
  (\bibinfo{date}{May} \bibinfo{year}{2021}), \bibinfo{pages}{14867--14875}.
\newblock
\href{https://doi.org/10.1609/aaai.v35i17.17745}{doi:\nolinkurl{10.1609/aaai.v35i17.17745}}


\bibitem[Mazari et~al\mbox{.}(2024)]%
        {mazari_bert-based_2024}
\bibfield{author}{\bibinfo{person}{Ahmed~Cherif Mazari},
  \bibinfo{person}{Nesrine Boudoukhani}, {and} \bibinfo{person}{Abdelhamid
  Djeffal}.} \bibinfo{year}{2024}\natexlab{}.
\newblock \showarticletitle{{BERT}-based ensemble learning for multi-aspect
  hate speech detection}.
\newblock \bibinfo{journal}{\emph{Cluster Computing}} \bibinfo{volume}{27},
  \bibinfo{number}{1} (\bibinfo{date}{Feb.} \bibinfo{year}{2024}),
  \bibinfo{pages}{325--339}.
\newblock
\showISSN{1386-7857, 1573-7543}
\href{https://doi.org/10.1007/s10586-022-03956-x}{doi:\nolinkurl{10.1007/s10586-022-03956-x}}


\bibitem[Mishra et~al\mbox{.}(2021)]%
        {mishra_exploring_2021}
\bibfield{author}{\bibinfo{person}{Sudhanshu Mishra}, \bibinfo{person}{Shivangi
  Prasad}, {and} \bibinfo{person}{Shubhanshu Mishra}.}
  \bibinfo{year}{2021}\natexlab{}.
\newblock \showarticletitle{Exploring {Multi}-{Task} {Multi}-{Lingual}
  {Learning} of {Transformer} {Models} for {Hate} {Speech} and {Offensive}
  {Speech} {Identification} in {Social} {Media}}.
\newblock \bibinfo{journal}{\emph{SN Computer Science}} \bibinfo{volume}{2},
  \bibinfo{number}{2} (\bibinfo{date}{April} \bibinfo{year}{2021}),
  \bibinfo{pages}{72}.
\newblock
\showISSN{2662-995X, 2661-8907}
\href{https://doi.org/10.1007/s42979-021-00455-5}{doi:\nolinkurl{10.1007/s42979-021-00455-5}}


\bibitem[Mollas et~al\mbox{.}(2022)]%
        {mollas_ethos_2022}
\bibfield{author}{\bibinfo{person}{Ioannis Mollas}, \bibinfo{person}{Zoe
  Chrysopoulou}, \bibinfo{person}{Stamatis Karlos}, {and}
  \bibinfo{person}{Grigorios Tsoumakas}.} \bibinfo{year}{2022}\natexlab{}.
\newblock \showarticletitle{{ETHOS}: a multi-label hate speech detection
  dataset}.
\newblock \bibinfo{journal}{\emph{Complex \& Intelligent Systems}}
  \bibinfo{volume}{8}, \bibinfo{number}{6} (\bibinfo{date}{Dec.}
  \bibinfo{year}{2022}), \bibinfo{pages}{4663--4678}.
\newblock
\showISSN{2199-4536, 2198-6053}
\href{https://doi.org/10.1007/s40747-021-00608-2}{doi:\nolinkurl{10.1007/s40747-021-00608-2}}


\bibitem[Narula and Chaudhary(2025)]%
        {narula_comprehensive_2025}
\bibfield{author}{\bibinfo{person}{Rachna Narula} {and} \bibinfo{person}{Poonam
  Chaudhary}.} \bibinfo{year}{2025}\natexlab{}.
\newblock \showarticletitle{A comprehensive review on detection of hate speech
  for multi-lingual data}.
\newblock \bibinfo{journal}{\emph{Social Network Analysis and Mining}}
  \bibinfo{volume}{14}, \bibinfo{number}{1} (\bibinfo{date}{Jan.}
  \bibinfo{year}{2025}), \bibinfo{pages}{244}.
\newblock
\showISSN{1869-5469}
\href{https://doi.org/10.1007/s13278-024-01401-y}{doi:\nolinkurl{10.1007/s13278-024-01401-y}}


\bibitem[Nielsen(2002)]%
        {nielsen_subtle_2002}
\bibfield{author}{\bibinfo{person}{Laura~Beth Nielsen}.}
  \bibinfo{year}{2002}\natexlab{}.
\newblock \showarticletitle{Subtle, {Pervasive}, {Harmful}: {Racist} and
  {Sexist} {Remarks} in {Public} as {Hate} {Speech}}.
\newblock \bibinfo{journal}{\emph{Journal of Social Issues}}
  \bibinfo{volume}{58}, \bibinfo{number}{2} (\bibinfo{date}{Jan.}
  \bibinfo{year}{2002}), \bibinfo{pages}{265--280}.
\newblock
\showISSN{0022-4537, 1540-4560}
\href{https://doi.org/10.1111/1540-4560.00260}{doi:\nolinkurl{10.1111/1540-4560.00260}}


\bibitem[Ousidhoum et~al\mbox{.}(2019)]%
        {ousidhoum_multilingual_2019}
\bibfield{author}{\bibinfo{person}{Nedjma Ousidhoum}, \bibinfo{person}{Zizheng
  Lin}, \bibinfo{person}{Hongming Zhang}, \bibinfo{person}{Yangqiu Song}, {and}
  \bibinfo{person}{Dit-Yan Yeung}.} \bibinfo{year}{2019}\natexlab{}.
\newblock \showarticletitle{Multilingual and {Multi}-{Aspect} {Hate} {Speech}
  {Analysis}}. In \bibinfo{booktitle}{\emph{Proceedings of the 2019
  {Conference} on {Empirical} {Methods} in {Natural} {Language} {Processing}
  and the 9th {International} {Joint} {Conference} on {Natural} {Language}
  {Processing} ({EMNLP}-{IJCNLP})}}. \bibinfo{publisher}{Association for
  Computational Linguistics}, \bibinfo{address}{Hong Kong, China},
  \bibinfo{pages}{4674--4683}.
\newblock
\href{https://doi.org/10.18653/v1/D19-1474}{doi:\nolinkurl{10.18653/v1/D19-1474}}


\bibitem[Ozler et~al\mbox{.}(2020)]%
        {ozler_fine-tuning_2020}
\bibfield{author}{\bibinfo{person}{Kadir~Bulut Ozler}, \bibinfo{person}{Kate
  Kenski}, \bibinfo{person}{Steve Rains}, \bibinfo{person}{Yotam Shmargad},
  \bibinfo{person}{Kevin Coe}, {and} \bibinfo{person}{Steven Bethard}.}
  \bibinfo{year}{2020}\natexlab{}.
\newblock \showarticletitle{Fine-tuning for multi-domain and multi-label
  uncivil language detection}. In \bibinfo{booktitle}{\emph{Proceedings of the
  {Fourth} {Workshop} on {Online} {Abuse} and {Harms}}}.
  \bibinfo{publisher}{Association for Computational Linguistics},
  \bibinfo{address}{Online}, \bibinfo{pages}{28--33}.
\newblock
\href{https://doi.org/10.18653/v1/2020.alw-1.4}{doi:\nolinkurl{10.18653/v1/2020.alw-1.4}}


\bibitem[Paasch-Colberg et~al\mbox{.}(2021)]%
        {paasch-colberg_insult_2021}
\bibfield{author}{\bibinfo{person}{Sünje Paasch-Colberg},
  \bibinfo{person}{Christian Strippel}, \bibinfo{person}{Joachim Trebbe}, {and}
  \bibinfo{person}{Martin Emmer}.} \bibinfo{year}{2021}\natexlab{}.
\newblock \showarticletitle{From {Insult} to {Hate} {Speech}: {Mapping}
  {Offensive} {Language} in {German} {User} {Comments} on {Immigration}}.
\newblock \bibinfo{journal}{\emph{Media and Communication}}
  \bibinfo{volume}{9}, \bibinfo{number}{1} (\bibinfo{date}{Feb.}
  \bibinfo{year}{2021}), \bibinfo{pages}{171--180}.
\newblock
\showISSN{2183-2439}
\href{https://doi.org/10.17645/mac.v9i1.3399}{doi:\nolinkurl{10.17645/mac.v9i1.3399}}


\bibitem[Pan and Yang(2010)]%
        {pan_survey_2010}
\bibfield{author}{\bibinfo{person}{Sinno~Jialin Pan} {and}
  \bibinfo{person}{Qiang Yang}.} \bibinfo{year}{2010}\natexlab{}.
\newblock \showarticletitle{A Survey on Transfer Learning}.
\newblock \bibinfo{journal}{\emph{IEEE Transactions on Knowledge and Data
  Engineering}} \bibinfo{volume}{22}, \bibinfo{number}{10}
  (\bibinfo{year}{2010}), \bibinfo{pages}{1345--1359}.
\newblock
\href{https://doi.org/10.1109/TKDE.2009.191}{doi:\nolinkurl{10.1109/TKDE.2009.191}}


\bibitem[Parikh et~al\mbox{.}(2019)]%
        {parikh_multi-label_2019}
\bibfield{author}{\bibinfo{person}{Pulkit Parikh}, \bibinfo{person}{Harika
  Abburi}, \bibinfo{person}{Pinkesh Badjatiya}, \bibinfo{person}{Radhika
  Krishnan}, \bibinfo{person}{Niyati Chhaya}, \bibinfo{person}{Manish Gupta},
  {and} \bibinfo{person}{Vasudeva Varma}.} \bibinfo{year}{2019}\natexlab{}.
\newblock \showarticletitle{Multi-label {Categorization} of {Accounts} of
  {Sexism} using a {Neural} {Framework}}. In
  \bibinfo{booktitle}{\emph{Proceedings of the 2019 {Conference} on {Empirical}
  {Methods} in {Natural} {Language} {Processing} and the 9th {International}
  {Joint} {Conference} on {Natural} {Language} {Processing}
  ({EMNLP}-{IJCNLP})}}. \bibinfo{publisher}{Association for Computational
  Linguistics}, \bibinfo{address}{Hong Kong, China},
  \bibinfo{pages}{1642--1652}.
\newblock
\href{https://doi.org/10.18653/v1/D19-1174}{doi:\nolinkurl{10.18653/v1/D19-1174}}


\bibitem[Parikh et~al\mbox{.}(2021)]%
        {parikh_categorizing_2021}
\bibfield{author}{\bibinfo{person}{Pulkit Parikh}, \bibinfo{person}{Harika
  Abburi}, \bibinfo{person}{Niyati Chhaya}, \bibinfo{person}{Manish Gupta},
  {and} \bibinfo{person}{Vasudeva Varma}.} \bibinfo{year}{2021}\natexlab{}.
\newblock \showarticletitle{Categorizing {Sexism} and {Misogyny} through
  {Neural} {Approaches}}.
\newblock \bibinfo{journal}{\emph{ACM Transactions on the Web}}
  \bibinfo{volume}{15}, \bibinfo{number}{4} (\bibinfo{date}{Nov.}
  \bibinfo{year}{2021}), \bibinfo{pages}{1--31}.
\newblock
\showISSN{1559-1131, 1559-114X}
\href{https://doi.org/10.1145/3457189}{doi:\nolinkurl{10.1145/3457189}}


\bibitem[Paré et~al\mbox{.}(2015)]%
        {pare_synthesizing_2015}
\bibfield{author}{\bibinfo{person}{Guy Paré}, \bibinfo{person}{Marie-Claude
  Trudel}, \bibinfo{person}{Mirou Jaana}, {and} \bibinfo{person}{Spyros
  Kitsiou}.} \bibinfo{year}{2015}\natexlab{}.
\newblock \showarticletitle{Synthesizing information systems knowledge: {A}
  typology of literature reviews}.
\newblock \bibinfo{journal}{\emph{Information \& Management}}
  \bibinfo{volume}{52}, \bibinfo{number}{2} (\bibinfo{date}{March}
  \bibinfo{year}{2015}), \bibinfo{pages}{183--199}.
\newblock
\showISSN{03787206}
\href{https://doi.org/10.1016/j.im.2014.08.008}{doi:\nolinkurl{10.1016/j.im.2014.08.008}}


\bibitem[Perez et~al\mbox{.}(2023)]%
        {perez_pysentimiento_2023}
\bibfield{author}{\bibinfo{person}{Juan~Manuel Perez}, \bibinfo{person}{Mariela
  Rajngewerc}, \bibinfo{person}{Juan~Carlos Giudici},
  \bibinfo{person}{Damián~Ariel Furman}, \bibinfo{person}{Franco Luque},
  \bibinfo{person}{Laura~Alonso Alemany}, {and} \bibinfo{person}{María~Vanina
  Martínez}.} \bibinfo{year}{2023}\natexlab{}.
\newblock \bibinfo{title}{pysentimiento: {A} {Python} {Toolkit} for {Opinion}
  {Mining} and {Social} {NLP} tasks}.
\newblock
\href{https://doi.org/10.21203/rs.3.rs-3570648/v1}{doi:\nolinkurl{10.21203/rs.3.rs-3570648/v1}}


\bibitem[Peskine et~al\mbox{.}(2023)]%
        {peskine_definitions_2023}
\bibfield{author}{\bibinfo{person}{Youri Peskine}, \bibinfo{person}{Damir
  Korenčić}, \bibinfo{person}{Ivan Grubisic}, \bibinfo{person}{Paolo
  Papotti}, \bibinfo{person}{Raphael Troncy}, {and} \bibinfo{person}{Paolo
  Rosso}.} \bibinfo{year}{2023}\natexlab{}.
\newblock \showarticletitle{Definitions {Matter}: {Guiding} {GPT} for
  {Multi}-label {Classification}}. In \bibinfo{booktitle}{\emph{Findings of the
  {Association} for {Computational} {Linguistics}: {EMNLP} 2023}}.
  \bibinfo{publisher}{Association for Computational Linguistics},
  \bibinfo{address}{Singapore}, \bibinfo{pages}{4054--4063}.
\newblock
\href{https://doi.org/10.18653/v1/2023.findings-emnlp.267}{doi:\nolinkurl{10.18653/v1/2023.findings-emnlp.267}}


\bibitem[Pfeffer et~al\mbox{.}(2023)]%
        {pfeffer_this_2023}
\bibfield{author}{\bibinfo{person}{Jürgen Pfeffer}, \bibinfo{person}{Angelina
  Mooseder}, \bibinfo{person}{Jana Lasser}, \bibinfo{person}{Luca Hammer},
  \bibinfo{person}{Oliver Stritzel}, {and} \bibinfo{person}{David Garcia}.}
  \bibinfo{year}{2023}\natexlab{}.
\newblock \showarticletitle{This {Sample} {Seems} to {Be} {Good} {Enough}!
  {Assessing} {Coverage} and {Temporal} {Reliability} of {Twitter}’s
  {Academic} {API}}.
\newblock \bibinfo{journal}{\emph{Proceedings of the International AAAI
  Conference on Web and Social Media}}  \bibinfo{volume}{17}
  (\bibinfo{date}{June} \bibinfo{year}{2023}), \bibinfo{pages}{720--729}.
\newblock
\showISSN{2334-0770, 2162-3449}
\href{https://doi.org/10.1609/icwsm.v17i1.22182}{doi:\nolinkurl{10.1609/icwsm.v17i1.22182}}


\bibitem[Plaza et~al\mbox{.}(2023)]%
        {arampatzis_overview_2023}
\bibfield{author}{\bibinfo{person}{Laura Plaza}, \bibinfo{person}{Jorge
  Carrillo-de Albornoz}, \bibinfo{person}{Roser Morante},
  \bibinfo{person}{Enrique Amigó}, \bibinfo{person}{Julio Gonzalo},
  \bibinfo{person}{Damiano Spina}, {and} \bibinfo{person}{Paolo Rosso}.}
  \bibinfo{year}{2023}\natexlab{}.
\newblock \showarticletitle{Overview of {EXIST} 2023 – {Learning} with
  {Disagreement} for {Sexism} {Identification} and {Characterization}}.
\newblock In \bibinfo{booktitle}{\emph{Experimental {IR} {Meets}
  {Multilinguality}, {Multimodality}, and {Interaction}}},
  \bibfield{editor}{\bibinfo{person}{Avi Arampatzis},
  \bibinfo{person}{Evangelos Kanoulas}, \bibinfo{person}{Theodora Tsikrika},
  \bibinfo{person}{Stefanos Vrochidis}, \bibinfo{person}{Anastasia Giachanou},
  \bibinfo{person}{Dan Li}, \bibinfo{person}{Mohammad Aliannejadi},
  \bibinfo{person}{Michalis Vlachos}, \bibinfo{person}{Guglielmo Faggioli},
  {and} \bibinfo{person}{Nicola Ferro}} (Eds.). Vol.~\bibinfo{volume}{14163}.
  \bibinfo{publisher}{Springer Nature Switzerland}, \bibinfo{address}{Cham},
  \bibinfo{pages}{316--342}.
\newblock
\showISBNx{978-3-031-42447-2 978-3-031-42448-9}
\href{https://doi.org/10.1007/978-3-031-42448-9_23}{doi:\nolinkurl{10.1007/978-3-031-42448-9_23}}
\newblock
\shownote{Series Title: Lecture Notes in Computer Science}.


\bibitem[Poletto et~al\mbox{.}(2021)]%
        {poletto_resources_2021}
\bibfield{author}{\bibinfo{person}{Fabio Poletto}, \bibinfo{person}{Valerio
  Basile}, \bibinfo{person}{Manuela Sanguinetti}, \bibinfo{person}{Cristina
  Bosco}, {and} \bibinfo{person}{Viviana Patti}.}
  \bibinfo{year}{2021}\natexlab{}.
\newblock \showarticletitle{Resources and benchmark corpora for hate speech
  detection: a systematic review}.
\newblock \bibinfo{journal}{\emph{Language Resources and Evaluation}}
  \bibinfo{volume}{55}, \bibinfo{number}{2} (\bibinfo{date}{June}
  \bibinfo{year}{2021}), \bibinfo{pages}{477--523}.
\newblock
\showISSN{1574-020X, 1574-0218}
\href{https://doi.org/10.1007/s10579-020-09502-8}{doi:\nolinkurl{10.1007/s10579-020-09502-8}}


\bibitem[Price et~al\mbox{.}(2020)]%
        {price_six_2020}
\bibfield{author}{\bibinfo{person}{Ilan Price}, \bibinfo{person}{Jordan
  Gifford-Moore}, \bibinfo{person}{Jory Flemming}, \bibinfo{person}{Saul
  Musker}, \bibinfo{person}{Maayan Roichman}, \bibinfo{person}{Guillaume
  Sylvain}, \bibinfo{person}{Nithum Thain}, \bibinfo{person}{Lucas Dixon},
  {and} \bibinfo{person}{Jeffrey Sorensen}.} \bibinfo{year}{2020}\natexlab{}.
\newblock \showarticletitle{Six {Attributes} of {Unhealthy} {Conversations}}.
  In \bibinfo{booktitle}{\emph{Proceedings of the {Fourth} {Workshop} on
  {Online} {Abuse} and {Harms}}}. \bibinfo{publisher}{Association for
  Computational Linguistics}, \bibinfo{address}{Online},
  \bibinfo{pages}{114--124}.
\newblock
\href{https://doi.org/10.18653/v1/2020.alw-1.15}{doi:\nolinkurl{10.18653/v1/2020.alw-1.15}}


\bibitem[Ramos et~al\mbox{.}(2024)]%
        {ramos_comprehensive_2024}
\bibfield{author}{\bibinfo{person}{Gil Ramos}, \bibinfo{person}{Fernando
  Batista}, \bibinfo{person}{Ricardo Ribeiro}, \bibinfo{person}{Pedro Fialho},
  \bibinfo{person}{Sérgio Moro}, \bibinfo{person}{António Fonseca},
  \bibinfo{person}{Rita Guerra}, \bibinfo{person}{Paula Carvalho},
  \bibinfo{person}{Catarina Marques}, {and} \bibinfo{person}{Cláudia Silva}.}
  \bibinfo{year}{2024}\natexlab{}.
\newblock \showarticletitle{A comprehensive review on automatic hate speech
  detection in the age of the transformer}.
\newblock \bibinfo{journal}{\emph{Social Network Analysis and Mining}}
  \bibinfo{volume}{14}, \bibinfo{number}{1} (\bibinfo{date}{Oct.}
  \bibinfo{year}{2024}), \bibinfo{pages}{204}.
\newblock
\showISSN{1869-5469}
\href{https://doi.org/10.1007/s13278-024-01361-3}{doi:\nolinkurl{10.1007/s13278-024-01361-3}}


\bibitem[Ratadiya and Mishra(2019)]%
        {ratadiya_attention_2019}
\bibfield{author}{\bibinfo{person}{Pratik Ratadiya} {and}
  \bibinfo{person}{Deepak Mishra}.} \bibinfo{year}{2019}\natexlab{}.
\newblock \showarticletitle{An {Attention} {Ensemble} {Based} {Approach} for
  {Multilabel} {Profanity} {Detection}}. In \bibinfo{booktitle}{\emph{2019
  {International} {Conference} on {Data} {Mining} {Workshops} ({ICDMW})}}.
  \bibinfo{publisher}{IEEE}, \bibinfo{address}{Beijing, China},
  \bibinfo{pages}{544--550}.
\newblock
\showISBNx{978-1-72814-896-0}
\href{https://doi.org/10.1109/ICDMW.2019.00083}{doi:\nolinkurl{10.1109/ICDMW.2019.00083}}


\bibitem[Rawat et~al\mbox{.}(2024)]%
        {rawat_hate_2024}
\bibfield{author}{\bibinfo{person}{Anchal Rawat}, \bibinfo{person}{Santosh
  Kumar}, {and} \bibinfo{person}{Surender~Singh Samant}.}
  \bibinfo{year}{2024}\natexlab{}.
\newblock \showarticletitle{Hate speech detection in social media:
  {Techniques}, recent trends, and future challenges}.
\newblock \bibinfo{journal}{\emph{WIREs Computational Statistics}}
  \bibinfo{volume}{16}, \bibinfo{number}{2} (\bibinfo{date}{March}
  \bibinfo{year}{2024}), \bibinfo{pages}{e1648}.
\newblock
\showISSN{1939-5108, 1939-0068}
\href{https://doi.org/10.1002/wics.1648}{doi:\nolinkurl{10.1002/wics.1648}}


\bibitem[Risch and Krestel(2020)]%
        {agarwal_toxic_2020}
\bibfield{author}{\bibinfo{person}{Julian Risch} {and} \bibinfo{person}{Ralf
  Krestel}.} \bibinfo{year}{2020}\natexlab{}.
\newblock \showarticletitle{Toxic {Comment} {Detection} in {Online}
  {Discussions}}.
\newblock In \bibinfo{booktitle}{\emph{Deep {Learning}-{Based} {Approaches} for
  {Sentiment} {Analysis}}}, \bibfield{editor}{\bibinfo{person}{Basant Agarwal},
  \bibinfo{person}{Richi Nayak}, \bibinfo{person}{Namita Mittal}, {and}
  \bibinfo{person}{Srikanta Patnaik}} (Eds.). \bibinfo{publisher}{Springer
  Singapore}, \bibinfo{address}{Singapore}, \bibinfo{pages}{85--109}.
\newblock
\showISBNx{9789811512155 9789811512162}
\href{https://doi.org/10.1007/978-981-15-1216-2_4}{doi:\nolinkurl{10.1007/978-981-15-1216-2_4}}
\newblock
\shownote{Series Title: Algorithms for Intelligent Systems}.


\bibitem[Ruder et~al\mbox{.}(2019)]%
        {ruder_latent_2019}
\bibfield{author}{\bibinfo{person}{Sebastian Ruder}, \bibinfo{person}{Joachim
  Bingel}, \bibinfo{person}{Isabelle Augenstein}, {and} \bibinfo{person}{Anders
  Søgaard}.} \bibinfo{year}{2019}\natexlab{}.
\newblock \showarticletitle{Latent {Multi}-{Task} {Architecture} {Learning}}.
\newblock \bibinfo{journal}{\emph{Proceedings of the AAAI Conference on
  Artificial Intelligence}} \bibinfo{volume}{33}, \bibinfo{number}{01}
  (\bibinfo{date}{July} \bibinfo{year}{2019}), \bibinfo{pages}{4822--4829}.
\newblock
\showISSN{2374-3468, 2159-5399}
\href{https://doi.org/10.1609/aaai.v33i01.33014822}{doi:\nolinkurl{10.1609/aaai.v33i01.33014822}}


\bibitem[Rybinski et~al\mbox{.}(2018)]%
        {del_ser_design_2018}
\bibfield{author}{\bibinfo{person}{Maciej Rybinski}, \bibinfo{person}{William
  Miller}, \bibinfo{person}{Javier Del~Ser}, \bibinfo{person}{Miren~Nekane
  Bilbao}, {and} \bibinfo{person}{José~F. Aldana-Montes}.}
  \bibinfo{year}{2018}\natexlab{}.
\newblock \showarticletitle{On the {Design} and {Tuning} of {Machine}
  {Learning} {Models} for {Language} {Toxicity} {Classification} in {Online}
  {Platforms}}.
\newblock In \bibinfo{booktitle}{\emph{Intelligent {Distributed} {Computing}
  {XII}}}, \bibfield{editor}{\bibinfo{person}{Javier Del~Ser},
  \bibinfo{person}{Eneko Osaba}, \bibinfo{person}{Miren~Nekane Bilbao},
  \bibinfo{person}{Javier~J. Sanchez-Medina}, \bibinfo{person}{Massimo
  Vecchio}, {and} \bibinfo{person}{Xin-She Yang}} (Eds.).
  Vol.~\bibinfo{volume}{798}. \bibinfo{publisher}{Springer International
  Publishing}, \bibinfo{address}{Cham}, \bibinfo{pages}{329--343}.
\newblock
\showISBNx{978-3-319-99625-7 978-3-319-99626-4}
\href{https://doi.org/10.1007/978-3-319-99626-4_29}{doi:\nolinkurl{10.1007/978-3-319-99626-4_29}}
\newblock
\shownote{Series Title: Studies in Computational Intelligence}.


\bibitem[Röttger et~al\mbox{.}(2021)]%
        {rottger_hatecheck_2021}
\bibfield{author}{\bibinfo{person}{Paul Röttger}, \bibinfo{person}{Bertie
  Vidgen}, \bibinfo{person}{Dong Nguyen}, \bibinfo{person}{Zeerak Waseem},
  \bibinfo{person}{Helen Margetts}, {and} \bibinfo{person}{Janet
  Pierrehumbert}.} \bibinfo{year}{2021}\natexlab{}.
\newblock \showarticletitle{{HateCheck}: {Functional} {Tests} for {Hate}
  {Speech} {Detection} {Models}}. In \bibinfo{booktitle}{\emph{Proceedings of
  the 59th {Annual} {Meeting} of the {Association} for {Computational}
  {Linguistics} and the 11th {International} {Joint} {Conference} on {Natural}
  {Language} {Processing} ({Volume} 1: {Long} {Papers})}}.
  \bibinfo{publisher}{Association for Computational Linguistics},
  \bibinfo{address}{Online}, \bibinfo{pages}{41--58}.
\newblock
\href{https://doi.org/10.18653/v1/2021.acl-long.4}{doi:\nolinkurl{10.18653/v1/2021.acl-long.4}}


\bibitem[Saeed et~al\mbox{.}(2018)]%
        {saeed_overlapping_2018}
\bibfield{author}{\bibinfo{person}{Hafiz~Hassaan Saeed},
  \bibinfo{person}{Khurram Shahzad}, {and} \bibinfo{person}{Faisal Kamiran}.}
  \bibinfo{year}{2018}\natexlab{}.
\newblock \showarticletitle{Overlapping {Toxic} {Sentiment} {Classification}
  {Using} {Deep} {Neural} {Architectures}}. In \bibinfo{booktitle}{\emph{2018
  {IEEE} {International} {Conference} on {Data} {Mining} {Workshops}
  ({ICDMW})}}. \bibinfo{publisher}{IEEE}, \bibinfo{address}{Singapore,
  Singapore}, \bibinfo{pages}{1361--1366}.
\newblock
\showISBNx{978-1-5386-9288-2}
\href{https://doi.org/10.1109/ICDMW.2018.00193}{doi:\nolinkurl{10.1109/ICDMW.2018.00193}}


\bibitem[Salminen et~al\mbox{.}(2018)]%
        {salminen_anatomy_2018}
\bibfield{author}{\bibinfo{person}{Joni Salminen}, \bibinfo{person}{Hind
  Almerekhi}, \bibinfo{person}{Milica Milenković}, \bibinfo{person}{Soon-gyo
  Jung}, \bibinfo{person}{Jisun An}, \bibinfo{person}{Haewoon Kwak}, {and}
  \bibinfo{person}{Bernard Jansen}.} \bibinfo{year}{2018}\natexlab{}.
\newblock \showarticletitle{Anatomy of {Online} {Hate}: {Developing} a
  {Taxonomy} and {Machine} {Learning} {Models} for {Identifying} and
  {Classifying} {Hate} in {Online} {News} {Media}}.
\newblock \bibinfo{journal}{\emph{Proceedings of the International AAAI
  Conference on Web and Social Media}} \bibinfo{volume}{12},
  \bibinfo{number}{1} (\bibinfo{date}{June} \bibinfo{year}{2018}),
  \bibinfo{pages}{330--339}.
\newblock
\showISSN{2334-0770, 2162-3449}
\href{https://doi.org/10.1609/icwsm.v12i1.15028}{doi:\nolinkurl{10.1609/icwsm.v12i1.15028}}


\bibitem[Schaffner et~al\mbox{.}(2024)]%
        {schaffner_community_2024}
\bibfield{author}{\bibinfo{person}{Brennan Schaffner},
  \bibinfo{person}{Arjun~Nitin Bhagoji}, \bibinfo{person}{Siyuan Cheng},
  \bibinfo{person}{Jacqueline Mei}, \bibinfo{person}{Jay~L Shen},
  \bibinfo{person}{Grace Wang}, \bibinfo{person}{Marshini Chetty},
  \bibinfo{person}{Nick Feamster}, \bibinfo{person}{Genevieve Lakier}, {and}
  \bibinfo{person}{Chenhao Tan}.} \bibinfo{year}{2024}\natexlab{}.
\newblock \showarticletitle{"{Community} {Guidelines} {Make} this the {Best}
  {Party} on the {Internet}": {An} {In}-{Depth} {Study} of {Online}
  {Platforms}' {Content} {Moderation} {Policies}}. In
  \bibinfo{booktitle}{\emph{Proceedings of the {CHI} {Conference} on {Human}
  {Factors} in {Computing} {Systems}}}. \bibinfo{publisher}{ACM},
  \bibinfo{address}{Honolulu HI USA}, \bibinfo{pages}{1--16}.
\newblock
\showISBNx{9798400703300}
\href{https://doi.org/10.1145/3613904.3642333}{doi:\nolinkurl{10.1145/3613904.3642333}}


\bibitem[Scheuerman et~al\mbox{.}(2021)]%
        {scheuerman_framework_2021}
\bibfield{author}{\bibinfo{person}{Morgan~Klaus Scheuerman},
  \bibinfo{person}{Jialun~Aaron Jiang}, \bibinfo{person}{Casey Fiesler}, {and}
  \bibinfo{person}{Jed~R. Brubaker}.} \bibinfo{year}{2021}\natexlab{}.
\newblock \showarticletitle{A {Framework} of {Severity} for {Harmful} {Content}
  {Online}}.
\newblock \bibinfo{journal}{\emph{Proceedings of the ACM on Human-Computer
  Interaction}} \bibinfo{volume}{5}, \bibinfo{number}{CSCW2}
  (\bibinfo{date}{Oct.} \bibinfo{year}{2021}), \bibinfo{pages}{1--33}.
\newblock
\showISSN{2573-0142}
\href{https://doi.org/10.1145/3479512}{doi:\nolinkurl{10.1145/3479512}}


\bibitem[Schmidt and Wiegand(2017)]%
        {schmidt_survey_2017}
\bibfield{author}{\bibinfo{person}{Anna Schmidt} {and} \bibinfo{person}{Michael
  Wiegand}.} \bibinfo{year}{2017}\natexlab{}.
\newblock \showarticletitle{A {Survey} on {Hate} {Speech} {Detection} using
  {Natural} {Language} {Processing}}. In \bibinfo{booktitle}{\emph{Proceedings
  of the {Fifth} {International} {Workshop} on {Natural} {Language}
  {Processing} for {Social} {Media}}}. \bibinfo{publisher}{Association for
  Computational Linguistics}, \bibinfo{address}{Valencia, Spain},
  \bibinfo{pages}{1--10}.
\newblock
\href{https://doi.org/10.18653/v1/W17-1101}{doi:\nolinkurl{10.18653/v1/W17-1101}}


\bibitem[Schäfer(2023)]%
        {schafer_bias_2023}
\bibfield{author}{\bibinfo{person}{Johannes Schäfer}.}
  \bibinfo{year}{2023}\natexlab{}.
\newblock \showarticletitle{Bias {Mitigation} for {Capturing} {Potentially}
  {Illegal} {Hate} {Speech}}.
\newblock \bibinfo{journal}{\emph{Datenbank-Spektrum}} \bibinfo{volume}{23},
  \bibinfo{number}{1} (\bibinfo{date}{March} \bibinfo{year}{2023}),
  \bibinfo{pages}{41--51}.
\newblock
\showISSN{1618-2162, 1610-1995}
\href{https://doi.org/10.1007/s13222-023-00439-0}{doi:\nolinkurl{10.1007/s13222-023-00439-0}}


\bibitem[Sellars(2016)]%
        {sellars_defining_2016}
\bibfield{author}{\bibinfo{person}{Andrew Sellars}.}
  \bibinfo{year}{2016}\natexlab{}.
\newblock \bibinfo{booktitle}{\emph{Defining {Hate} {Speech}}}.
\newblock \bibinfo{type}{Research {Publication}} 2016-20.
  \bibinfo{institution}{Berkman Klein Center for Internet and Society},
  \bibinfo{address}{Cambridge, MA, USA}.
\newblock
\urldef\tempurl%
\url{https://www.ssrn.com/abstract=2882244}
\showURL{%
\tempurl}


\bibitem[Siegel(2020)]%
        {siegel_online_2020}
\bibfield{author}{\bibinfo{person}{Alexandra~A. Siegel}.}
  \bibinfo{year}{2020}\natexlab{}.
\newblock \showarticletitle{Online {Hate} {Speech}}.
\newblock In \bibinfo{booktitle}{\emph{Social {Media} and {Democracy}: {The}
  {State} of the {Field}, {Prospects} for {Reform}}},
  \bibfield{editor}{\bibinfo{person}{Nathaniel Persily} {and}
  \bibinfo{person}{Joshua~A. Tucker}} (Eds.). \bibinfo{publisher}{Cambridge
  University Press}, \bibinfo{address}{Cambridge, MA, USA},
  \bibinfo{pages}{56--88}.
\newblock


\bibitem[Singhal et~al\mbox{.}(2023)]%
        {singhal_sok_2023}
\bibfield{author}{\bibinfo{person}{Mohit Singhal}, \bibinfo{person}{Chen Ling},
  \bibinfo{person}{Pujan Paudel}, \bibinfo{person}{Poojitha Thota},
  \bibinfo{person}{Nihal Kumarswamy}, \bibinfo{person}{Gianluca Stringhini},
  {and} \bibinfo{person}{Shirin Nilizadeh}.} \bibinfo{year}{2023}\natexlab{}.
\newblock \showarticletitle{{SoK}: {Content} {Moderation} in {Social} {Media},
  from {Guidelines} to {Enforcement}, and {Research} to {Practice}}. In
  \bibinfo{booktitle}{\emph{2023 {IEEE} 8th {European} {Symposium} on
  {Security} and {Privacy} ({EuroS}\&{P})}}. \bibinfo{publisher}{IEEE},
  \bibinfo{address}{Delft, Netherlands}, \bibinfo{pages}{868--895}.
\newblock
\showISBNx{978-1-66546-512-0}
\href{https://doi.org/10.1109/EuroSP57164.2023.00056}{doi:\nolinkurl{10.1109/EuroSP57164.2023.00056}}


\bibitem[Subramanian et~al\mbox{.}(2023)]%
        {subramanian_survey_2023}
\bibfield{author}{\bibinfo{person}{Malliga Subramanian},
  \bibinfo{person}{Veerappampalayam Easwaramoorthy~Sathiskumar},
  \bibinfo{person}{G. Deepalakshmi}, \bibinfo{person}{Jaehyuk Cho}, {and}
  \bibinfo{person}{G. Manikandan}.} \bibinfo{year}{2023}\natexlab{}.
\newblock \showarticletitle{A survey on hate speech detection and sentiment
  analysis using machine learning and deep learning models}.
\newblock \bibinfo{journal}{\emph{Alexandria Engineering Journal}}
  \bibinfo{volume}{80} (\bibinfo{date}{Oct.} \bibinfo{year}{2023}),
  \bibinfo{pages}{110--121}.
\newblock
\showISSN{11100168}
\href{https://doi.org/10.1016/j.aej.2023.08.038}{doi:\nolinkurl{10.1016/j.aej.2023.08.038}}


\bibitem[Tsoumakas and Katakis(2007)]%
        {tsoumakas_multi-label_2007}
\bibfield{author}{\bibinfo{person}{Grigorios Tsoumakas} {and}
  \bibinfo{person}{Ioannis Katakis}.} \bibinfo{year}{2007}\natexlab{}.
\newblock \showarticletitle{Multi-{Label} {Classification}: {An} {Overview}}.
\newblock \bibinfo{journal}{\emph{International Journal of Data Warehousing and
  Mining}} \bibinfo{volume}{3}, \bibinfo{number}{3} (\bibinfo{date}{July}
  \bibinfo{year}{2007}), \bibinfo{pages}{1--13}.
\newblock
\showISSN{1548-3924, 1548-3932}
\href{https://doi.org/10.4018/jdwm.2007070101}{doi:\nolinkurl{10.4018/jdwm.2007070101}}


\bibitem[Van~Aken et~al\mbox{.}(2018)]%
        {van_aken_challenges_2018}
\bibfield{author}{\bibinfo{person}{Betty Van~Aken}, \bibinfo{person}{Julian
  Risch}, \bibinfo{person}{Ralf Krestel}, {and} \bibinfo{person}{Alexander
  Löser}.} \bibinfo{year}{2018}\natexlab{}.
\newblock \showarticletitle{Challenges for {Toxic} {Comment} {Classification}:
  {An} {In}-{Depth} {Error} {Analysis}}. In
  \bibinfo{booktitle}{\emph{Proceedings of the 2nd {Workshop} on {Abusive}
  {Language} {Online} ({ALW2})}}. \bibinfo{publisher}{Association for
  Computational Linguistics}, \bibinfo{address}{Brussels, Belgium},
  \bibinfo{pages}{33--42}.
\newblock
\href{https://doi.org/10.18653/v1/W18-5105}{doi:\nolinkurl{10.18653/v1/W18-5105}}


\bibitem[Vaswani et~al\mbox{.}(2017)]%
        {vaswani_attention_2017}
\bibfield{author}{\bibinfo{person}{Ashish Vaswani}, \bibinfo{person}{Noam
  Shazeer}, \bibinfo{person}{Niki Parmar}, \bibinfo{person}{Jakob Uszkoreit},
  \bibinfo{person}{Llion Jones}, \bibinfo{person}{Aidan~N. Gomez},
  \bibinfo{person}{Lukasz Kaiser}, {and} \bibinfo{person}{Illia Polosukhin}.}
  \bibinfo{year}{2017}\natexlab{}.
\newblock \showarticletitle{Attention is {All} you {Need}}. In
  \bibinfo{booktitle}{\emph{Advances in {Neural} {Information} {Processing}
  {Systems} 30: {Annual} {Conference} on {Neural} {Information} {Processing}
  {Systems}}}, \bibfield{editor}{\bibinfo{person}{Isabelle Guyon},
  \bibinfo{person}{Ulrike von Luxenburg}, \bibinfo{person}{Samy Bengio},
  \bibinfo{person}{Hannah~M. Wallach}, \bibinfo{person}{Rob Fergus},
  \bibinfo{person}{S.~V.~N. Vishwanathan}, {and} \bibinfo{person}{Roman
  Garnett}} (Eds.). \bibinfo{publisher}{Curran Associates, Inc.},
  \bibinfo{address}{Long Beach, CA, USA}, \bibinfo{pages}{5998--6008}.
\newblock
\urldef\tempurl%
\url{https://proceedings.neurips.cc/paper/2017/hash/3f5ee243547dee91fbd053c1c4a845aa-Abstract.html}
\showURL{%
\tempurl}


\bibitem[Vidgen et~al\mbox{.}(2019)]%
        {vidgen_challenges_2019}
\bibfield{author}{\bibinfo{person}{Bertie Vidgen}, \bibinfo{person}{Alex
  Harris}, \bibinfo{person}{Dong Nguyen}, \bibinfo{person}{Rebekah Tromble},
  \bibinfo{person}{Scott Hale}, {and} \bibinfo{person}{Helen Margetts}.}
  \bibinfo{year}{2019}\natexlab{}.
\newblock \showarticletitle{Challenges and frontiers in abusive content
  detection}. In \bibinfo{booktitle}{\emph{Proceedings of the {Third}
  {Workshop} on {Abusive} {Language} {Online}}}.
  \bibinfo{publisher}{Association for Computational Linguistics},
  \bibinfo{address}{Florence, Italy}, \bibinfo{pages}{80--93}.
\newblock
\href{https://doi.org/10.18653/v1/W19-3509}{doi:\nolinkurl{10.18653/v1/W19-3509}}


\bibitem[Vidgen et~al\mbox{.}(2021)]%
        {vidgen_learning_2021}
\bibfield{author}{\bibinfo{person}{Bertie Vidgen}, \bibinfo{person}{Tristan
  Thrush}, \bibinfo{person}{Zeerak Waseem}, {and} \bibinfo{person}{Douwe
  Kiela}.} \bibinfo{year}{2021}\natexlab{}.
\newblock \showarticletitle{Learning from the {Worst}: {Dynamically}
  {Generated} {Datasets} to {Improve} {Online} {Hate} {Detection}}. In
  \bibinfo{booktitle}{\emph{Proceedings of the 59th {Annual} {Meeting} of the
  {Association} for {Computational} {Linguistics} and the 11th {International}
  {Joint} {Conference} on {Natural} {Language} {Processing} ({Volume} 1: {Long}
  {Papers})}}. \bibinfo{publisher}{Association for Computational Linguistics},
  \bibinfo{address}{Online}, \bibinfo{pages}{1667--1682}.
\newblock
\href{https://doi.org/10.18653/v1/2021.acl-long.132}{doi:\nolinkurl{10.18653/v1/2021.acl-long.132}}


\bibitem[Vom~Brocke et~al\mbox{.}(2015)]%
        {vom_brocke_standing_2015}
\bibfield{author}{\bibinfo{person}{Jan Vom~Brocke}, \bibinfo{person}{Alexander
  Simons}, \bibinfo{person}{Kai Riemer}, \bibinfo{person}{Björn Niehaves},
  \bibinfo{person}{Ralf Plattfaut}, {and} \bibinfo{person}{Anne Cleven}.}
  \bibinfo{year}{2015}\natexlab{}.
\newblock \showarticletitle{Standing on the {Shoulders} of {Giants}:
  {Challenges} and {Recommendations} of {Literature} {Search} in {Information}
  {Systems} {Research}}.
\newblock \bibinfo{journal}{\emph{Communications of the Association for
  Information Systems}}  \bibinfo{volume}{37} (\bibinfo{year}{2015}),
  \bibinfo{pages}{205–224}.
\newblock
\showISSN{15293181}
\href{https://doi.org/10.17705/1CAIS.03709}{doi:\nolinkurl{10.17705/1CAIS.03709}}


\bibitem[Waseem(2016)]%
        {waseem_are_2016}
\bibfield{author}{\bibinfo{person}{Zeerak Waseem}.}
  \bibinfo{year}{2016}\natexlab{}.
\newblock \showarticletitle{Are {You} a {Racist} or {Am} {I} {Seeing} {Things}?
  {Annotator} {Influence} on {Hate} {Speech} {Detection} on {Twitter}}. In
  \bibinfo{booktitle}{\emph{Proceedings of the {First} {Workshop} on {NLP} and
  {Computational} {Social} {Science}}}. \bibinfo{publisher}{Association for
  Computational Linguistics}, \bibinfo{address}{Austin, Texas},
  \bibinfo{pages}{138--142}.
\newblock
\href{https://doi.org/10.18653/v1/W16-5618}{doi:\nolinkurl{10.18653/v1/W16-5618}}


\bibitem[Waseem and Hovy(2016)]%
        {waseem_hateful_2016}
\bibfield{author}{\bibinfo{person}{Zeerak Waseem} {and} \bibinfo{person}{Dirk
  Hovy}.} \bibinfo{year}{2016}\natexlab{}.
\newblock \showarticletitle{Hateful {Symbols} or {Hateful} {People}?
  {Predictive} {Features} for {Hate} {Speech} {Detection} on {Twitter}}. In
  \bibinfo{booktitle}{\emph{Proceedings of the {NAACL} {Student} {Research}
  {Workshop}}}. \bibinfo{publisher}{Association for Computational Linguistics},
  \bibinfo{address}{San Diego, California}, \bibinfo{pages}{88--93}.
\newblock
\href{https://doi.org/10.18653/v1/N16-2013}{doi:\nolinkurl{10.18653/v1/N16-2013}}


\bibitem[Wiegand et~al\mbox{.}(2018)]%
        {wiegand_overview_2018}
\bibfield{author}{\bibinfo{person}{Michael Wiegand}, \bibinfo{person}{Melanie
  Siegel}, {and} \bibinfo{person}{Josef Ruppenhofer}.}
  \bibinfo{year}{2018}\natexlab{}.
\newblock \showarticletitle{Overview of the germeval 2018 shared task on the
  identification of offensive language}. In
  \bibinfo{booktitle}{\emph{Proceedings of the {GermEval} 2018 {Workshop}}}.
  \bibinfo{publisher}{Austrian Academy of Sciences}, \bibinfo{address}{Vienna,
  Austria}, \bibinfo{pages}{1--10}.
\newblock
\newblock
\shownote{Publisher: Verlag der Österreichischen Akademie der Wissenschaften}.


\bibitem[Wulczyn et~al\mbox{.}(2017)]%
        {wulczyn_ex_2017}
\bibfield{author}{\bibinfo{person}{Ellery Wulczyn}, \bibinfo{person}{Nithum
  Thain}, {and} \bibinfo{person}{Lucas Dixon}.}
  \bibinfo{year}{2017}\natexlab{}.
\newblock \showarticletitle{Ex {Machina}: {Personal} {Attacks} {Seen} at
  {Scale}}. In \bibinfo{booktitle}{\emph{Proceedings of the 26th
  {International} {Conference} on {World} {Wide} {Web}}}.
  \bibinfo{publisher}{International World Wide Web Conferences Steering
  Committee}, \bibinfo{address}{Perth Australia}, \bibinfo{pages}{1391--1399}.
\newblock
\showISBNx{978-1-4503-4913-0}
\href{https://doi.org/10.1145/3038912.3052591}{doi:\nolinkurl{10.1145/3038912.3052591}}


\bibitem[Yin and Zubiaga(2021)]%
        {yin_towards_2021}
\bibfield{author}{\bibinfo{person}{Wenjie Yin} {and} \bibinfo{person}{Arkaitz
  Zubiaga}.} \bibinfo{year}{2021}\natexlab{}.
\newblock \showarticletitle{Towards generalisable hate speech detection: a
  review on obstacles and solutions}.
\newblock \bibinfo{journal}{\emph{PeerJ Computer Science}}  \bibinfo{volume}{7}
  (\bibinfo{date}{June} \bibinfo{year}{2021}), \bibinfo{pages}{e598}.
\newblock
\showISSN{2376-5992}
\href{https://doi.org/10.7717/peerj-cs.598}{doi:\nolinkurl{10.7717/peerj-cs.598}}


\bibitem[Zampieri et~al\mbox{.}(2019)]%
        {zampieri_predicting_2019}
\bibfield{author}{\bibinfo{person}{Marcos Zampieri}, \bibinfo{person}{Shervin
  Malmasi}, \bibinfo{person}{Preslav Nakov}, \bibinfo{person}{Sara Rosenthal},
  \bibinfo{person}{Noura Farra}, {and} \bibinfo{person}{Ritesh Kumar}.}
  \bibinfo{year}{2019}\natexlab{}.
\newblock \showarticletitle{Predicting the {Type} and {Target} of {Offensive}
  {Posts} in {Social} {Media}}. In \bibinfo{booktitle}{\emph{Proceedings of the
  2019 {Conference} of the {North}}}. \bibinfo{publisher}{Association for
  Computational Linguistics}, \bibinfo{address}{Minneapolis, Minnesota},
  \bibinfo{pages}{1415--1420}.
\newblock
\href{https://doi.org/10.18653/v1/N19-1144}{doi:\nolinkurl{10.18653/v1/N19-1144}}


\bibitem[Zhang and Zhou(2014)]%
        {zhang_review_2014}
\bibfield{author}{\bibinfo{person}{Min-Ling Zhang} {and}
  \bibinfo{person}{Zhi-Hua Zhou}.} \bibinfo{year}{2014}\natexlab{}.
\newblock \showarticletitle{A {Review} on {Multi}-{Label} {Learning}
  {Algorithms}}.
\newblock \bibinfo{journal}{\emph{IEEE Transactions on Knowledge and Data
  Engineering}} \bibinfo{volume}{26}, \bibinfo{number}{8} (\bibinfo{date}{Aug.}
  \bibinfo{year}{2014}), \bibinfo{pages}{1819--1837}.
\newblock
\showISSN{1041-4347}
\href{https://doi.org/10.1109/TKDE.2013.39}{doi:\nolinkurl{10.1109/TKDE.2013.39}}


\bibitem[Zhang et~al\mbox{.}(2015)]%
        {zhang_bidirectional_2015}
\bibfield{author}{\bibinfo{person}{Shu Zhang}, \bibinfo{person}{Dequan Zheng},
  \bibinfo{person}{Xinchen Hu}, {and} \bibinfo{person}{Ming Yang}.}
  \bibinfo{year}{2015}\natexlab{}.
\newblock \showarticletitle{Bidirectional {Long} {Short}-{Term} {Memory}
  {Networks} for {Relation} {Classification}}. In
  \bibinfo{booktitle}{\emph{Proceedings of the 29th {Pacific} {Asia}
  {Conference} on {Language}, {Information} and {Computation}}},
  \bibfield{editor}{\bibinfo{person}{Hai Zhao}} (Ed.).
  \bibinfo{publisher}{Association for Computational Linguistics},
  \bibinfo{address}{Shanghai, China}, \bibinfo{pages}{73--78}.
\newblock
\urldef\tempurl%
\url{https://aclanthology.org/Y15-1009/}
\showURL{%
\tempurl}


\bibitem[Zhang and Yang(2018)]%
        {zhang_overview_2018}
\bibfield{author}{\bibinfo{person}{Yu Zhang} {and} \bibinfo{person}{Qiang
  Yang}.} \bibinfo{year}{2018}\natexlab{}.
\newblock \showarticletitle{An overview of multi-task learning}.
\newblock \bibinfo{journal}{\emph{National Science Review}}
  \bibinfo{volume}{5}, \bibinfo{number}{1} (\bibinfo{date}{Jan.}
  \bibinfo{year}{2018}), \bibinfo{pages}{30--43}.
\newblock
\showISSN{2095-5138, 2053-714X}
\href{https://doi.org/10.1093/nsr/nwx105}{doi:\nolinkurl{10.1093/nsr/nwx105}}


\bibitem[Zhao et~al\mbox{.}(2021)]%
        {zhao_comparative_2021}
\bibfield{author}{\bibinfo{person}{Zhixue Zhao}, \bibinfo{person}{Ziqi Zhang},
  {and} \bibinfo{person}{Frank Hopfgartner}.} \bibinfo{year}{2021}\natexlab{}.
\newblock \showarticletitle{A {Comparative} {Study} of {Using} {Pre}-trained
  {Language} {Models} for {Toxic} {Comment} {Classification}}. In
  \bibinfo{booktitle}{\emph{Companion {Proceedings} of the {Web} {Conference}
  2021}}. \bibinfo{publisher}{ACM}, \bibinfo{address}{Ljubljana Slovenia},
  \bibinfo{pages}{500--507}.
\newblock
\showISBNx{978-1-4503-8313-4}
\href{https://doi.org/10.1145/3442442.3452313}{doi:\nolinkurl{10.1145/3442442.3452313}}


\bibitem[Zufall et~al\mbox{.}(2022)]%
        {zufall_legal_2022}
\bibfield{author}{\bibinfo{person}{Frederike Zufall}, \bibinfo{person}{Marius
  Hamacher}, \bibinfo{person}{Katharina Kloppenborg}, {and}
  \bibinfo{person}{Torsten Zesch}.} \bibinfo{year}{2022}\natexlab{}.
\newblock \showarticletitle{A {Legal} {Approach} to {Hate} {Speech} –
  {Operationalizing} the {EU}’s {Legal} {Framework} against the {Expression}
  of {Hatred} as an {NLP} {Task}}. In \bibinfo{booktitle}{\emph{Proceedings of
  the {Natural} {Legal} {Language} {Processing} {Workshop} 2022}}.
  \bibinfo{publisher}{Association for Computational Linguistics},
  \bibinfo{address}{Abu Dhabi, United Arab Emirates (Hybrid)},
  \bibinfo{pages}{53--64}.
\newblock
\href{https://doi.org/10.18653/v1/2022.nllp-1.5}{doi:\nolinkurl{10.18653/v1/2022.nllp-1.5}}


\end{thebibliography}

\appendix
\section{Appendix}
\label{sec:review:appendix}

\subsection{Performance of State-of-the-Art Binary Hate Speech Classification Models}
\label{sec:review:appendix:SotA}
To evaluate the performance of the most prominent models for \textbf{binary} hate speech classification, we have used the HateCheck dataset \cite{rottger_hatecheck_2021}, which has specifically been designed to evaluate the performance of (multi-label) hate speech classification models. We evaluated the following models: the hate speech detection module from the Pysentimiento toolkit\footnote{https://github.com/pysentimiento/pysentimiento}, Google’s Perspective \ac{api}\footnote{https://perspectiveapi.com/} as well as the seven most popular (more than 1000 downloads in the last 30 days) Hugging Face\footnote{https://huggingface.co/} models when filtering for the keyword ``hate'', while filters for English Language and text classification are applied. The ToxiGen model \cite{hartvigsen_toxigen_2022} was not investigated because it fails with an error on some inputs (\eg, ``Nice''). The performance evaluation results are presented in \tabref{tab:review:sota}.

\begin{table}[ht!]
\caption{Performances of binary models on HateCheck \cite{rottger_hatecheck_2021}. The best values are highlighted in \textbf{bold}.}
\label{tab:review:sota}
\centering
\begin{tabular}{lrrrrr}
\toprule
Model & Acc & F1 & P & R & ROC-AUC \\
\midrule
Perspective API \href{https://perspectiveapi.com/}{[link]} & 0.578 & 0.559 & \textbf{0.993} & 0.389 & 0.894 \\
pysentimiento \cite{perez_pysentimiento_2023} & 0.521 & 0.527 & 0.820 & 0.388 & 0.688 \\
Facebook's RoBERTa Hate Speech Model \cite{vidgen_learning_2021} & \textbf{0.956} & \textbf{0.968} & 0.963 & \textbf{0.973} & \textbf{0.979} \\
English Abusive MuRIL \cite{das_data_2022} & 0.491 & 0.558 & 0.694 & 0.466 & 0.505 \\
BERT HateXplain \cite{mathew_hatexplain_2021} & 0.384 & 0.270 & 0.730 & 0.165 & N/A\\
DehateBERT Mono English \cite{aluru_deep_2020} & 0.425 & 0.351 & 0.784 & 0.226 & 0.580\\
IMSyPP Hate Speech \cite{ciucci_handling_2022} & 0.750 & 0.826 & 0.790 & 0.866 & N/A\\
Twitter RoBERTa Large Hate \cite{antypas_supertweeteval_2023} & 0.615 & 0.640 & 0.898 & 0.497 & N/A \\
DistilRoBERTa Hateful Speech \href{https://huggingface.co/badmatr11x/distilroberta-base-offensive-hateful-speech-text-multiclassification}{[link]} & 0.568 & 0.652 & 0.730 & 0.590 & N/A\\
\bottomrule
\end{tabular}
\raggedright
\footnotesize Abbreviations: \acs{acc} = \acl{acc}; \acs{f1} = \acl{f1}; \acs{p} = \acl{p}; \acs{r} = \acl{r}; \acs{roc-auc} = \acl{roc-auc}.
\end{table}

\end{document}